\documentclass{article}

\usepackage[preprint,nonatbib]{neurips_2025}

\usepackage{inputenc} % allow utf-8 input
\usepackage[T1]{fontenc}    % use 8-bit T1 fonts
\usepackage{hyperref}       % hyperlinks
\usepackage{url}            % simple URL typesetting
\usepackage{booktabs}       % professional-quality tables
\usepackage{amsfonts}       % blackboard math symbols
\usepackage{nicefrac}       % compact symbols for 1/2, etc.
\usepackage{microtype}      % microtypography     % colors
\usepackage{amsmath}
\usepackage{framed}
\usepackage{amsthm}
\usepackage{mathrsfs}
\usepackage{amssymb}
\usepackage{enumitem}
\usepackage{graphicx}
\usepackage{caption}
\usepackage{float}

\usepackage{tabularx}
\usepackage{booktabs}
\usepackage{makecell}
\usepackage{geometry}
\usepackage{footnote}
\usepackage[perpage,para]{footmisc}
\usepackage[svgnames, dvipsnames, table]{xcolor}
\usepackage{tikz} 
\usepackage{verbatim}
\usepackage{esint}
\usepackage{xspace}
\usepackage{CJK}
\usepackage[numbers,sort&compress]{natbib}
\usepackage{indentfirst}
\usepackage[commandnameprefix=always,highlightmarkup=uuline]{changes}
\usepackage{xurl} % make URLs breakable in references
\usepackage[most]{tcolorbox}
\usepackage{xparse} % 定义带可选参数的命令

\usepackage{tabularray}
\UseTblrLibrary{booktabs}
\definecolor{DeepPurple}{HTML}{3C1361}
\definecolor{LightPurpleA}{HTML}{C8B3E6}
\definecolor{LightPurpleB}{HTML}{E3DAF7}

\setdeletedmarkup{\color{gray}{\sout{#1}}}

\NewTColorBox{PromptBlock}{O{} m}{%
  enhanced, breakable,
  title={#2},               % 标题文字
  fonttitle=\bfseries,      % 标题加粗
  coltitle=white,           % 标题文字颜色（白字）
  colbacktitle=blue!60!black!80,     % 标题带背景（蓝）
  title filled,             % 开启“整条标题带”样式
  colback=gray!4,           % 正文背景
  colframe=blue!60!black,   % 边框颜色
  boxrule=0.5pt, arc=2pt,   % 边框粗细和圆角
  left=2mm, right=2mm, top=1.2mm, bottom=1.2mm,
  #1                        % 可选项透传，方便微调
}

\NewTColorBox{ResponseBlock}{O{} m}{%
  enhanced, breakable,
  title={#2},
  fonttitle=\bfseries,
  coltitle=white,
  colback=gray!5,                      % 背景更接近灰色
  colframe=green!25!black,             % 边框低饱和
  colbacktitle=green!35!black!75,      % 标题背景偏灰绿、亮度降低
  title filled,
  boxrule=0.5pt, arc=2pt,
  left=2mm, right=2mm, top=1.2mm, bottom=1.2mm,
  #1
}

\NewTColorBox{AnnotationGreen}{O{} O{Note}}{%
    enhanced, breakable,
    title={#2},
    fonttitle=\normalfont\itshape\bfseries\footnotesize,
    colback=gray!4,
    colframe=orange!50!black,
    colbacktitle=orange!50!black!80,   % 标题带背景（黄/橙）
    coltitle=white,
    title filled,
    boxrule=0.5pt, arc=2pt,
    left=1.2mm, right=1.2mm, top=1mm, bottom=1mm,
    before skip=2mm, after skip=2mm,
    fontupper=\normalfont\itshape,     % 正文为斜体，常规字号
    fontlower=\normalfont\itshape,     % 同样适用于底部
    #1
}

\NewTColorBox{AnnotationRed}{O{} O{Note}}{%
    enhanced, breakable,
    title={#2},
    fonttitle=\normalfont\itshape\bfseries\footnotesize,
    colback=gray!5,                       % 浅灰背景，降低整体亮度
    colframe=red!80!black,                % 深红灰边框，低饱和
    colbacktitle=red!80!black!90,         % 标题栏为偏暗红，白字可读性好
    coltitle=white,
    title filled,
    boxrule=0.5pt, arc=2pt,
    left=1.2mm, right=1.2mm, top=1mm, bottom=1mm,
    before skip=2mm, after skip=2mm,
    fontupper=\normalfont\itshape,     % 正文为斜体，常规字号
    fontlower=\normalfont\itshape,     % 同样适用于底部
#1
}

\definecolor{HeaderBlue}{RGB}{31,74,125}
\definecolor{RowGreen}{RGB}{217,240,211}
\definecolor{RowYellow}{RGB}{254,242,204}
\definecolor{RowOrange}{RGB}{253,225,152}
\definecolor{Lgreen}{rgb}{0,0.6,0}

\newcolumntype{C}{>{\centering\arraybackslash}X}

\newcommand{\R}{\mathbb{R}} % 实数集符号
\newcommand{\abs}[1]{\left|#1\right|} % 自动调整大小的绝对值
\newcommand{\norm}[1]{\left\Vert#1\right\Vert} % 自动调整大小的范数
\newcommand{\indicator}[1]{\mathbf{1}_{#1}} % 指示函数（规范符号）
\theoremstyle{plain}
\newtheorem*{lemma*}{Lemma} % 无自动编号，完全遵循原始编号（含重复Lemma 4）
\newtheorem*{definition*}{Definition} 

\newcommand{\diff}{\mathop{}\!\mathrm{d}}
\newcommand{\Div}{\operatorname{div}}
\newtheorem{lemma}{Lemma}
\newtheorem{maincontentlemma}{Lemma}
\newtheorem{theorem}{Theorem}
\newtheorem{definition}{Definition}
\newtheorem{conjecture}{Conjecture}

\setlist[itemize]{leftmargin=2em}

\title{AI Mathematician as a Partner \\in Advancing Mathematical Discovery \\ {\it\large--- A Case Study in Homogenization Theory}}
\author{
    Yuanhang Liu$^{1}$,
    Beichen Wang$^{1}$,
    Peng Li$^{2}$\thanks{Correspondence: Peng Li (lipeng@air.tsinghua.edu.cn) and Yang Liu (liuyang2011@tsinghua.edu.cn).}$^{\phantom{*}}$,
    Yang Liu$^{2,3}$\footnotemark[1]\\
    $^{1}$ Qiuzhen College, Tsinghua University, Beijing, China\\
    $^{2}$ Institute for AI Industry Research (AIR), Tsinghua University, Beijing, China\\
    $^{3}$ Dept. of Comp. Sci. \& Tech., Institute for AI, Tsinghua University, Beijing, China
}
\date{}

\begin{document}
%\begin{CJK*}{UTF8}{gbsn}

\maketitle
\begin{abstract}
    Artificial intelligence (AI) has demonstrated impressive progress in mathematical reasoning, yet its integration into the practice of mathematical research remains limited. In this study, we investigate how the AI Mathematician (AIM) system can operate as a research partner rather than a mere problem solver. Focusing on a challenging problem in homogenization theory, we analyze the autonomous reasoning trajectories of AIM and incorporate targeted human interventions to structure the discovery process. Through iterative decomposition of the problem into tractable subgoals, selection of appropriate analytical methods, and validation of intermediate results, we reveal how human intuition and machine computation can complement one another. This collaborative paradigm enhances the reliability, transparency, and interpretability of the resulting proofs, while retaining human oversight for formal rigor and correctness. The approach leads to a complete and verifiable proof, and more broadly, demonstrates how systematic human-AI co-reasoning can advance the frontier of mathematical discovery. 
\end{abstract}

\begin{figure}[H]
    \centering
    \includegraphics[width=0.82\textwidth]{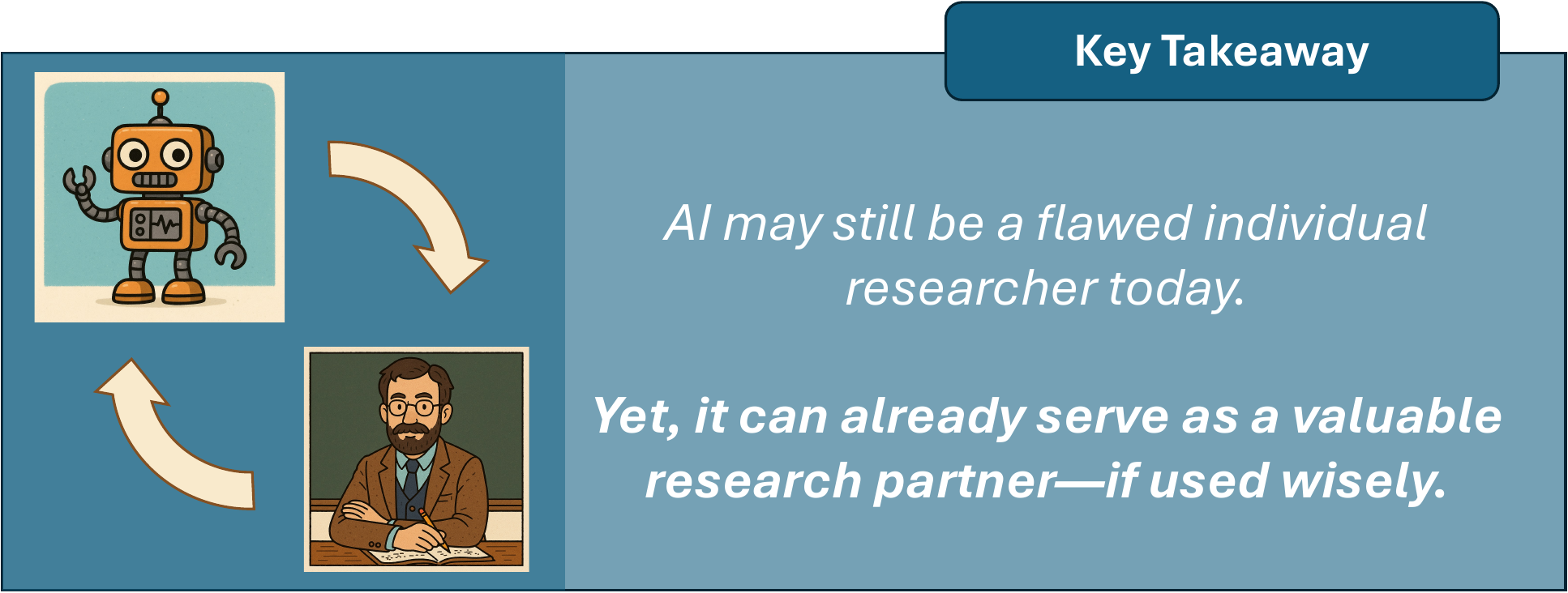}
\end{figure}

\section{Introduction}

In recent years, artificial intelligence (AI) has made remarkable progress in mathematical reasoning, achieving milestones once thought to be exclusive to human intelligence. In mathematical competitions, large language models (LLMs) have demonstrated outstanding performance. For example, several LLMs have achieved scores exceeding 90 on the AIME benchmarks~\citep{GPT-5,o4-mini-medium,Qwen3-Max-Thinking,Grok4}, which are constructed from real American Invitational Mathematics Examination (AIME) problems~\citep{AIME-MAA,AIME-Problems-and-Solutions}, and some have even reached perfect 100-point scores~\citep{GPT-5,Qwen3-Max-Thinking,Grok4}. Furthermore, Gemini with Deep Think has officially attained a gold-medal standard at the 66th International Mathematical Olympiad (IMO 2025)~\citep{GeminiIMO2025}, marking a symbolic moment in the competitive mathematical performance of AI.
As competition-based benchmarks begin to approach saturation, \citet{FrontierMath} introduced FrontierMath, a challenging new evaluation suite composed of problems crafted by expert mathematicians that typically require hours of deliberate reasoning for experts to solve. Notably, the o4-mini-medium model~\citep{o4-mini-medium} has been reported to outperform the average human team on this benchmark~\citep{FrontierMath-superhuman}, underscoring the growing ability of AI to engage with complex, research-level mathematics.
Beyond competition-style problem solving, progress has also emerged in AI-assisted mathematical discovery. \citet{Romera-Paredes2024} and \citet{AlphaEvolve} demonstrate that LLMs can facilitate genuine mathematical discovery through guided program search. Similarly, GPT-5-Thinking~\citep{GPT-5} has been credited with helping renowned researchers resolve a challenging quantum computing problem~\citep{aaronson2025QMA}. Taken together, these developments suggest that AI is beginning to move beyond solving predefined problems toward a more engaged role in mathematical exploration.

Despite substantial progress, a considerable gap remains between current AI capabilities and the requirements of genuine mathematical research. In competitive settings, such as mathematical contests, problems are typically solved within minutes or hours. Even demanding benchmarks like FrontierMath challenge expert mathematicians for only several hours. By contrast, authentic mathematical research often unfolds over months or even years of sustained reasoning, conceptual innovation, and proof development. For instance, it took more than seven years for Andrew Wiles to complete the proof of Fermat's Last Theorem~\citep{Fermat1997}, illustrating the temporal and intellectual depth of true mathematical inquiry.
In the pursuit of mathematical discovery, existing representative AI systems are grounded in search-based paradigms that face inherent scalability and generalization constraints. Approaches such as FunSearch~\citep{Romera-Paredes2024} and AlphaEvolve~\citep{AlphaEvolve} depend on problems being formalizable in programmatic form, thereby limiting their applicability to only a subset of mathematical domains. Similarly, AlphaGeometry~\citep{AlphaGeometry} and its successor AlphaGeometry2~\citep{chervonyi2025goldmedalistperformancesolvingolympiad} focus exclusively on geometric reasoning, leaving vast areas of mathematics unexplored. While there are cases in which AI systems have contributed valuable insights---such as the work by \citet{aaronson2025QMA}---the completion and validation of proofs ultimately remain the domain of human researchers. This highlights the current limitations of AI in conducting autonomous, creative, and deeply integrative mathematical research.

To mitigate these limitations, this study explores the potential of more collaborative and interactive AI systems capable of engaging in sustained reasoning and iterative refinement. We investigate the application of AIM~\citep{AIM}, a multi-agent framework developed for autonomous mathematical exploration and proof generation, as a research partner in advancing mathematical inquiry. AIM harnesses the capabilities of LLMs to iteratively formulate conjectures, verify proofs, optimize reasoning pathways and refine proof details. By incorporating human expertise through a through several representative interaction paradigm, the framework seeks to integrate the complementary strengths of human intuition and machine intelligence in addressing complex, research-level mathematical problems.

In particular, we examine the use of AIM in tackling a challenge in homogenization theory, a discipline concerned with characterizing the macroscopic behavior of heterogeneous materials by averaging their microscopic properties across multiple scales. Through systematic analysis of AIM's intermediate reasoning trajectories and targeted human interventions, we decompose the overarching problem into tractable subgoals, guide the selection of appropriate analytical methodologies, and rigorously validate the correctness of intermediate results. This human-AI co-reasoning paradigm improves the reliability, transparency, and interpretability of candidate proofs and produces auditable reasoning traces, while recognizing that formal rigor and final correctness require human oversight. Through this collaborative approach, we successfully derive a complete proof (Appendix~\ref{sec:proof}) for the aforementioned problem and, in the process, systematize representative modes of human-AI interaction while distilling key empirical insights. We believe this study provides meaningful guidance for AI-assisted mathematical research and establishes a foundation for deeper human-AI collaboration in advancing mathematical discovery.

In summary, our main contributions are as follows:
\begin{itemize}
\item We advocate a human-AI collaborative paradigm for mathematical research that integrates the computational capabilities of AI systems with the intuition and domain expertise of human mathematicians.
\item We conduct a case study on a challenging problem in homogenization theory, resulting in a rigorous proof spanning nearly seventeen pages (Appendix~\ref{sec:proof}). A substantial portion of this proof is generated by AI, which makes nontrivial contributions throughout the process, demonstrating the potential of the paradigm for tackling complex, research-level mathematical problems.
\item We systematize modes of human-AI interaction and extract empirical insights that can inform the design of future AI-assisted mathematical research frameworks, which may also serve as a practical guideline for mathematicians seeking to leverage AI in their own research.
\end{itemize}

\section{Preliminaries}

\subsection{The Homogenization Problem}

The mathematical research problem we investigate in this work is an instance of a Stokes--Lam\'{e} transmission system with a vanishing fluid inclusion, analyzed in the homogenization regime $\varepsilon \rightarrow 0$. This problem will be referred to as the \textit{\textbf{Homogenization Problem}} in the rest of this work.

Consider $D \subset \Omega \subset R^{d} \,(d\ge 2)$, where $\Omega$ is elastic material and $D$ is the high contrast inclusion part.

\begin{itemize}
    \item  $\Omega$ is open bounded with connected \(C^{\infty}\) boundary $\partial\Omega$.
    \item $D$ is open, has a finite number of components and has a Lipschitz boundary $\partial D$.
    \item $\Omega \setminus D$ is connected with Lipschitz boundary $\partial \Omega \cup \partial D$. The connected components of $D$ are enumerated as $D_i$, $i = 1, \ldots, N$, $N$ is finite.
\end{itemize}

 And given \(\varepsilon \in (0, 1)\), \(D = D_\varepsilon\) is part of an \(\varepsilon\)-periodic array of small inclusions constructed as follows, in several steps.

\(Y = (-\frac{1}{2}, \frac{1}{2})^d\) is the unit cell. 
\(\omega \subset Y\) is a simple connected open subset with connected Lipschitz boundary such that \({dist}(\omega, \partial Y) > 0\).
\(Y_f = Y \setminus \overline{\omega}\) is the model environment in the unit scale.

Given \(\varepsilon > 0\) and \(\mathbf{n} \in \mathbb{Z}^d\), we denote \(\varepsilon(\mathbf{n} + Y)\) and \(\varepsilon(\mathbf{n} + \omega)\) by \(Y^\mathbf{n}_\varepsilon\) and \(\omega^\mathbf{n}_\varepsilon\), respectively.
Let \(\Pi_\varepsilon\) be the set of lattice points \(\mathbf{n}\) such that \(\overline{Y}^\mathbf{n}_\varepsilon\) be contained in \(\Omega\), i.e.,
\begin{equation}
    \Pi_\varepsilon := \left\{ \mathbf{n} \in \mathbb{Z}^d : \overline{Y}^\mathbf{n}_\varepsilon \subset \Omega \right\},
\end{equation}
then the inclusions set \(D = D_\varepsilon\) and the background part \(\Omega_\varepsilon\) are defined by
\begin{equation}
    D_\varepsilon := \bigcup_{\mathbf{n} \in \Pi_\varepsilon} \omega^\mathbf{n}_\varepsilon \quad \Omega_\varepsilon := \Omega \setminus \overline{D}_\varepsilon.
\end{equation}

A pair of real numbers \((\lambda, \mu)\) is called admissible and referred to as a $ \text{Lam\'e} $ pair, if they satisfy
    $\mu > 0$ and $ d\lambda + 2\mu > 0$. 
    For a $ \text{Lam\'e} $ pair \((\lambda, \mu)\), the elastostatic system ($ \text{Lam\'e} $ system) reads
    \begin{equation}
    \mathcal{L}_{\lambda,\mu}u := \mu\Delta u + (\lambda + \mu)\nabla\text{div}\,u,
   \end{equation}
where \( u = (u^1, \ldots, u^d) \) represents the displacement field and the divergence of \(u\) is given by \(\text{div}\,u = \sum_{i=1}^{d} \frac{\partial u^i}{\partial x_i}\). \textit{\textbf{It is worth noting that $u$, rather than $\mathbf{u}$, is used here to ensure consistency with the notation adopted by AIM.}}
The $ \text{Lam\'e} $ operator can be written as \(\nabla \cdot \sigma(u)\) where
     \begin{equation}
     \sigma(u) := \lambda(\nabla \cdot u)\mathbb{I}_d + 2\mu\mathcal{D}(u),
     \end{equation}
     \begin{equation}
     \mathcal{D}(u) = \frac{1}{2}(\nabla + \nabla^T)u = \frac{1}{2}(\partial_i u^j + \partial_j u^i)_{ij}.
     \end{equation}

The corresponding conormal derivative (boundary traction) at the boundary of a domain \(E\) is
\begin{equation}
    \left.\frac{\partial u}{\partial \nu_{(\lambda,\mu)}}\right|_{\partial E} := \sigma(u)N = \lambda(\text{div } u)N + 2\mu\mathcal{D}(u)N \quad \text{on } \partial E.
\end{equation}

We consider the space \(\mathbb{R}\) of rigid motions in \(\mathbb{R}^{d}\), defined by
\[
\mathbb{R} := \left\{ \mathbf{r} = (r_{1}, \ldots, r_{d})^{T} : \mathcal{D}(\mathbf{r}) = 0 \text{ in } \mathbb{R}^{d} \right\}.
\]We define \( H_{\mathbb{R}}^{-\frac{1}{2}}(\partial D_{\varepsilon}) \) as the subspace of \( H^{-\frac{1}{2}}(\partial D_{\varepsilon}) \) that is orthogonal to \(\mathbb{R}\), i.e.,
\begin{equation}
    H_{\mathbb{R}}^{-\frac{1}{2}}(\partial D_{\varepsilon}) := \left\{ \phi \in H^{-\frac{1}{2}}(\partial D_{\varepsilon}): \left(\phi,\mathbf{r}\right)_{(H^{\frac{1}{2}}(\partial {D_{\varepsilon_i}}),H^{-\frac{1}{2}}(\partial {D_{\varepsilon_i}}))} = 0 ,\forall \mathbf{r} \in \mathbb{R} \text{ and } 1 \leq i \leq N \right\}.
\end{equation}

Consider the displacement field $u_{\varepsilon}$ satisfying the following transmission system:
    \begin{equation}
    \begin{cases} 
        \mathcal{L}_{\lambda, \mu} u_{\varepsilon} = 0 & \text{in } \Omega \setminus \overline{D}_{\varepsilon}, \\ 
        \mathcal{L}_{\widetilde{\lambda}, \widetilde{\mu}} u_{\varepsilon} = 0 & \text{in } D_{\varepsilon}, \\
        u_{\varepsilon}|_{-} = u_{\varepsilon}|_{+} \text{ and } \left.\frac{\partial u_{\varepsilon}}{\partial \nu_{(\widetilde{\lambda}, \widetilde{\mu})}}\right|_{-} = \left.\frac{\partial u_{\varepsilon}}{\partial \nu_{(\lambda, \mu)}}\right|_{+} & \text{on } \partial D_{\varepsilon},  \\
        \left.\frac{\partial u_{\varepsilon}}{\partial \nu_{(\lambda, \mu)}}\right|_{\partial \Omega} = g \in H_{\mathbb{R}}^{-\frac{1}{2}}(\partial \Omega) \quad \text{and} \quad u_{\varepsilon}|_{\partial \Omega} \in H_{\mathbb{R}}^{\frac{1}{2}}(\partial \Omega).
    \end{cases}
    \end{equation}
    
Suppose \(\widetilde{{\mu}}\) fixed, then we arrive at the equations about the \textbf{incompressible inclusion limit}. In this case, the transmission problem is a coupled \textbf{$\text{Lam\'e}$-Stokes system}:  
    \begin{equation}
    \begin{cases} 
    \mathcal{L}_{\lambda, \mu} u_{\varepsilon} = 0 & \text{in } \Omega \setminus \overline{D}_{\varepsilon}, \\ 
    \mathcal{L}_{\widetilde{\mu}}(u_{\varepsilon}, p_{\varepsilon}) = 0 \:\text{and} \operatorname{div} \:u_{\varepsilon}=0  & \text{in } D_{\varepsilon}, \\ 
    
    u_{\varepsilon}|_- = u_{\varepsilon}|_+ \: \text{and} \: \left. \frac{\partial (u_{\varepsilon}, p_{\varepsilon})}{\partial \nu_{(\infty, \widetilde{\mu})}} \right|_- = \left. \frac{\partial u_{\varepsilon}}{\partial \nu_{(\lambda, \mu)}} \right|_+ & \text{on } \partial D_{\varepsilon}, \\ 
    
    \left. \frac{\partial u_{\varepsilon}}{\partial \nu_{(\lambda, \mu)}} \right|_{\partial \Omega} = g \in H_{\mathbb{R}}^{-\frac{1}{2}} (\partial \Omega) \quad\text{and} \quad u_{\varepsilon}|_{\partial \Omega} \in H_{\mathbb{R}}^{\frac{1}{2}} (\partial \Omega).
    \end{cases}
    \end{equation}
    Here, $\mathcal{L}_{\widetilde{\mu}}(u_{\varepsilon}, p_{\varepsilon}) = \widetilde{\mu} \Delta u_{\varepsilon} + \nabla p_{\varepsilon}$ denotes the Stokes operator with viscosity constant $\widetilde{\mu}$, and $p_{\varepsilon}$ is the pressure field.  \(N\) is the outward unit normal vector to the boundary of the domain. Its exterior derivative is defined as $
\frac{\partial (u, p)}{\partial \nu_{(\infty, \mu)}} \bigg|_{-} := pN + 2\mu D(u)N$.

We need to conclude the limit homogenization equation as the scale of the cell tends to be zero $\varepsilon \rightarrow 0 $. At the same time, we wonder the estimate between the original solution $u_{\varepsilon}$ and the limited solution $u_{\lim}$, i.e., 
\begin{equation}
    \Vert u_{\varepsilon}-u_{\lim} \Vert_{H^1(\Omega)}\lesssim \varepsilon^{\alpha}
    \label{eq:general_results}
\end{equation}
for some $\alpha \in (0,1)$. It is necessary to analyze and obtain determined value of $\alpha$, and strictly prove this conclusion.

\subsection{AIM: An AI Mathematician System}

AIM is a multi-agent framework built upon large language models (LLMs) for conducting mathematical research~\citep{AIM}. Its design addresses two fundamental challenges: the intrinsic complexity of mathematical theory and the rigor of reasoning processes.
On one hand, AIM incorporates an {\it exploration and memory mechanism} that decomposes complex problems into multi-step explorations, generates intermediate conjectures, and iteratively reuses verified lemmas to refine reasoning. This enables the framework to tackle problems that are beyond the direct problem-solving capabilities of LLMs.
On the other hand, AIM employs {\it Pessimistic Rational Verification (PRV)}, in which multiple independent verifiers evaluate each intermediate proof, and a proof is judged as incorrect if any verifier deems it incorrect. Experimental results demonstrate that PRV is effective in detecting erroneous proofs generated by LLMs, though it is not entirely reliable.

AIM consists of three core agents---the explorer, verifier, and optimizer---along with a memory module. The explorer is responsible for generating conjectures and constructing candidate proofs. The verifier independently examines the logical correctness of each proof step, while the optimizer refines and corrects errors identified during verification. The memory module archives automatically verified lemmas and other intermediate results for future reference.
The agents operate in an iterative manner: given a problem, the explorer first proposes a sequence of intermediate conjectures accompanied by detailed proofs, potentially leveraging information stored in the memory module. The verifier evaluates their validity, and if errors are detected, the optimizer revises the proofs based on feedback from the verifier. Verified proofs are then stored in the memory module as reusable lemmas, enabling the explorer to leverage accumulated knowledge in subsequent reasoning. This iterative process continues until a valid proof is obtained or a predefined iteration limit is reached.

AIM has been evaluated on four mathematical research problems: the quantum algorithm problem, the absorbing boundary condition problem, the high-contrast limit problem, and the homogenization problem examined in this work\footnote{For further details on the first three problems, please refer to the AIM technical report~\citep{AIM}.}. Experimental results show that AIM autonomously solved the first two problems, completed the main proof for the third, but failed to fully resolve the homogenization problem, leaving a substantial gap. Nevertheless, AIM made notable progress on this problem, and its intermediate results and proof sketches were found by human researchers to be highly insightful. This naturally leads to a key question: \textit{\textbf{Can human expertise be harnessed to guide AIM toward completing the proof for the homogenization problem?}} This question forms the basis of our investigation into human-AI collaboration in mathematical research.

\section{Overview}

Based on an AI-human collaborative paradigm, we successfully completed the proof of the homogenization problem. The full proof is presented in Appendix~\ref{sec:proof}. Throughout this process, \textit{\textbf{we deliberately minimized human intervention}}, allowing AIM to autonomously explore, reason, and construct mathematical arguments as much as possible. This design enables us to better understand both the strengths and limitations of AIM in mathematical research. The main conclusion for the homogenization problem is summarized as follows:
We derived the homogenization equation in limit case and denoted its solution as $u_{\lim}$ (Eq.~\ref{eq:homogenization} in Appendix~\ref{sec:proof}). The error estimation between the limit solution $u_{\lim}$ and the original solution $u_{\varepsilon}$ under the scale $\varepsilon$ is further explored. We have analyzed that the $\alpha=\frac{1}{2}$ and strictly proven the following conclusion:
\begin{equation}
\Vert u_{\varepsilon}-u_{\lim} \Vert_{H^1(\Omega)}\lesssim {\varepsilon}^{\frac{1}{2}}.
\label{eq:conclusion}
\end{equation}

Briefly, the proof was achieved through a staged process. Building on the autonomous reasoning results of AIM, we observed that the original problem exceeds the current capability of AIM; however, AIM could successfully tackle reduced or simplified subproblems and showed strong ability to recognize and apply relevant mathematical theories even without explicit guidance.
This observation brings three key advantages:
(1) it reduces the need for detailed human supervision in solving subproblems;
(2) it promotes exploration of diverse solution paths; and
(3) it accelerates the overall proof process.
Building on these insights, we divided the original homogenization problem into six subproblems and guided AIM through each. By weaving together the resulting proofs, we ultimately arrived at a complete and coherent solution to the problem.

The six subproblems include: (1) {\it Two-Scale Expansion}, (2) {\it Cell Problem and Homogenization Equation}, (3) {\it Existence and Uniqueness}, (4) {\it Ellipticity of Operator}, (5) {\it Error Estimation and Control}, and (6) {\it Regularity of Cell Problem}. The objectives of each subproblem and the respective roles of AIM and human experts are summarized in Table~\ref{tab:subproblems} and detailed below.
\begin{itemize}
    \item {\it Two-Scale Expansion}: By performing a multidimensional expansion on the equation based on this formula \(\nabla:=\nabla_x+\frac{1}{\varepsilon}\nabla_y\), we obtain equations corresponding to different dimensions. AIM made various errors in complex symbolic reasoning tasks, though in reality, this process is a relatively straightforward derivation for human workers. Here, we have manually derived the content of this work.
    \item {\it Cell Problem and Homogenization Equation}: Based on the equations at different scales, we derive the homogenization equation and construct the cell problem. To construct a cell problem, it is necessary to combine the inherent characteristics of the equation itself with the boundary conditions of the specific geometric structure. Experiments have shown that AIM has insufficient understanding of this task, making it difficult to obtain correct derivation results. Moreover, the thinking and methods for this task itself are straightforward and clear. Therefore, this task has been completed manually.
    \item {\it Existence and Uniqueness}:  After obtaining the homogenization equation, the first step is to analyze the existence and uniqueness of the solution $u_{\lim}$. AIM explored and applied reasonable theorems to analyze the operator properties of the equation, and the proof process was obtained after we manually adjusted and refined the details.
    \item {\it Ellipticity of Operator}: This subproblem concerns the proof of the ellipticity of the operator, which is a fundamental property in our analysis. Here, the experimental results of AIM provide a proof with a relatively high degree of completion. We have also obtained the derivation results manually through analytical techniques.
    \item {\it Error Estimation and Control}: This is the most complex subproblem, requiring a rigorous analytical process and detailed derivation process. Based on the output results of AIM, we manually adjusted and decomposed the subproblem, and finally obtained the complete proof process by manually refining results of AIM. Furthermore, in the analysis of the results, we discovered an important property and identified it as the next subproblem.
    \item {\it Regularity of Cell Problem}: This is the conclusion put forward by AIM during its autonomous exploration. However, AIM failed to provide a valid proof for it. In subsequent manual analysis, we determined that this property is a crucial conclusion for proving error estimation. Therefore, we attempted to apply AIM to prove this property. After multiple human-AI interaction experiments and with appropriate theoretical guidance, AIM finally produced a complete and correct proof.
\end{itemize}

The most representative AI-human interaction occurred during the resolution of the \textit{Error Estimation and Control} subproblem, which is the most complex one. Initially, AIM produced a seemingly convincing proof. However, upon closer examination, we found that it relied on a particular property of the cell problem equations without providing a rigorous justification. Drawing on mathematical intuition, the human expert conjectured that the property should hold and asked AIM to prove it---but AIM failed. This prompted deeper analysis, revealing that establishing this property required a more profound understanding of the underlying theory than initially assumed. In other words, \textit{this process helped the human expert better grasp the intrinsic difficulty of the problem.}
The human expert then suggested that AIM consider mathematical tools such as the Difference Quotient~\citep{leveque2007finite}, Galerkin Method~\citep{galerkin1915method}, and Schauder Theory~\citep{ladyzhenskaya1968linear,boccia2013schauder,ZHUGE2021213}, providing only their definitions without procedural hints or proof outlines. Notably, even the expert was uncertain whether these tools would apply. Eventually, AIM succeeded in proving the property using Schauder Theory.
This experience vividly illustrates that \textit{\textbf{although AIM may still be a flawed individual researcher today, it can already serve as a valuable research partner---if used wisely.}}

\begin{table}[ht]
\centering
\begin{tblr}{
width=\linewidth,
colspec = {|X[2,c] | Q[c,wd=2.2cm] | X[5,c] |}, % 三列都水平居中
hlines = {white, 1.4pt},
vlines = {white, 1.4pt},
colsep = 4pt,
rowsep = 7pt,
rows = {halign=c, valign=m}, % 全局：水平居中 + 垂直居中
row{1} = {bg=DeepPurple, fg=white, font=\bfseries, halign=c}, % 表头也居中
row{2,4,6} = {bg=LightPurpleA, fg=black},
row{3,5,7} = {bg=LightPurpleB, fg=black}
}
Steps & Hardness & Current Status \\
\textbf{Two-Scale Expansion} & \textbf{Easy} &  Correct expansion and derivation were completed manually.
 \\
\textbf{Cell Problem and Homogenization Equation} & \textbf{Medium} & The cell problem and homogenization equation were manually constructed and derived. \\
\textbf{Existence and Uniqueness} & \textbf{Hard} & AIM applied the correct theorem to get this conclusion. \\
\textbf{Ellipticity of Operator} & \textbf{Medium} & AIM provided a proof with a high level of completeness. \\
\textbf{Error Estimation and Control} & \textbf{Hard} & AIM presented the correct proof approach, which after human adjustments, led to a complete proof process. \\
\textbf{Regularity of Cell Problem} & \textbf{Hard} & By AIM, complete proof process was provided. \\
\end{tblr}
\vspace{0.5em}
\caption{Problem Decomposition and Summary}
\label{tab:subproblems}
\end{table}

\section{Modes of Human-AI Interaction}

In pursuing the complete proof of the homogenization problem, we found that effective human-AI collaboration plays a crucial role. Based on extensive experimentation, we summarize five representative modes of interaction that proved particularly effective:

\begin{itemize}
    \item \textbf{Direct Prompting} (Sec.~\ref{sec:direct_prompting}).
    This mode guides the agent toward promising proof directions and optimizes its reasoning path through targeted yet concise instructions. It can be further divided into three subtypes: \textit{Theorem Prompts} (Sec.~\ref{sec:theorem_prompt}), \textit{Conceptual Guidance} (Sec.~\ref{sec:conceptual_guidance}), and \textit{Detail Refinement} (Sec.~\ref{sec:detail_refinement}).

    \item \textbf{Theory-Coordinated Application} (Sec.~\ref{sec:theory_coordinated_application}).  
    In this mode, the agent is provided with a coherent body of mathematical theory, enabling it to derive related results within the theoretical framework. Unlike Theorem Prompts, which focus on specific goals, this mode emphasizes the integration and application of an entire theoretical system.

    \item \textbf{Interactive Iterative Refinement} (Sec.~\ref{sec:interactive_iteration}).  
    This mode follows a ``Feedback -- Revision -- Re-reasoning'' cycle, through which human experts and AIM collaboratively refine and complete proofs, leading to a more coherent and rigorous reasoning process.

    \item \textbf{Applicability Boundary and Exclusion Domain} (Sec.~\ref{sec:applicability_boundary_exclusion_domain}).  
    Certain tasks---such as decomposing proof strategies or constructing problem formulations---remain challenging for AIM. We therefore recommend assigning these tasks to human experts, while reserving AIM's involvement for domains where it demonstrates reliability and insight.

    \item \textbf{Auxiliary Optimization Strategies} (Sec.~\ref{sec:auxiliary_optimization_methods}). 
    These strategies enhance the correctness and robustness of proofs by iteratively providing additional contextual information and optimizing the selection or combination of mathematical tools.
\end{itemize}

These modes are not mutually exclusive and can be flexibly combined depending on the context and requirements of a given mathematical problem. In the following sections, we provide detailed explanations and representative examples for each mode.

\subsection{Direct Prompting}
\label{sec:direct_prompting}

Direct Prompting is an approach that provides clear human guidance, directing AIM's attention to critical elements in the proof process, thereby minimizing irrelevant or off-topic reasoning to the greatest extent. This approach is particularly effective during the initial construction phase of the proof framework, or when handling critical steps that require conceptual breakthroughs. The human's role is to pre-provide clear theorems, key formulas or theorem citations along with their applicable conditions, and to supply additional information such as required parameters, units, and boundary constraints. The role of agents is to verify the applicability of the input, then complete the reasoning process according to the indicated content without conducting independent strategy searches or method replacements. The main goal is to obtain fast, controllable, stable, and verifiable consistent results with minimal uncertainty and the shortest decision chain.
\begin{itemize}
    \item \textbf{Applicable Conditions}:  The solution approach is foreseeable by the human, and the required theoretical components are known, but the derivation details are tedious. When the agents repeatedly reasons without converging on a reasonable approach, constrain it to use only the given knowledge, thereby reducing autonomous conjecture and limiting unstated assumptions. This approach is also suitable for verification tasks, where the aim is to check whether a claim follows from the provided premises.
    \item \textbf{Typical Mathematical Scenarios}: For relatively simple subproblems, we can quickly identify the applicable theorem or solution method and directly prompt AIM to use the corresponding tool. For computational cases, we can specify common methods and techniques and require the model to follow the prescribed steps.
    When validating conclusions or conjectures, we can provide guidance (e.g., boundary-case and counterexample checks, numerical experiments, derivation paths based on known theorems) to let AIM explore the process and record intermediate findings and failed attempts.
    \item \textbf{Expected Outcomes}: AIM correctly applies the provided theorems and completes the proof within the given knowledge pack, with preconditions checked and steps properly cited.
    Note that the agents may misapplies the theorems, e.g., errors in derivation details or failure to satisfy applicability conditions. In this case, record the issues and consider a new interaction mode.
\end{itemize}

The prompts can be categorized into three common and distinct types:
\begin{itemize}
    \item \textbf{Theorem Prompts}: Directly supplying formal statements, assumptions, or target theorems to anchor AIM's reasoning process.
    \item \textbf{Conceptual Guidance}: Offering high-level descriptions of strategies, intuitions, or mathematical insights that contextualize the problem.
    \item \textbf{Detail Refinement}: Providing step-specific instructions or local corrections to ensure logical coherence and technical precision at finer granularity.
\end{itemize}

This structured prompting framework enables more controlled and efficient interaction with AIM, particularly in complex mathematical scenarios.

\subsubsection{Theorem Prompts}
\label{sec:theorem_prompt}

During experiments, we directly supplied AIM with theorems and lemmas relevant to the current stage of the proof, thus anchoring its reasoning process within a well-defined mathematical foundation. This form of guidance proves to be highly effective in narrowing AIM's focus and structuring the proof pathway.

For instance, in the proof of the Cell Problem subproblem, we explicitly introduced auxiliary lemmas from Schauder Theory and provided them as prompts to AIM. By doing so, we instructed AIM to utilize these lemmas in deriving subsequent conclusions, effectively structuring and constraining its reasoning process toward a valid and complete argument.

\begin{PromptBlock}{Prompt to AIM}
\begin{AnnotationRed}[drop shadow][Disclaimer]
For clarity, we format prompts using typographic styles (e.g., bold text) and render formulas as symbols rather than raw LaTeX code. ``\textbf{Content}'' is one of the labels used by AIM to distinguish different components and can be safely ignored when interpreting the prompt. The same applies below.
\end{AnnotationRed}

\textbf{Content}: 

You can use the following theorem to analyze the problem, integrating the content of the theorem and its derivation process. 

\begin{AnnotationRed}[drop shadow][Disclaimer]
The outputs of AIM include content such as \verb|\begin{lemma}\end{lemma}|. We do not modify these outputs. LaTeX automatically assigns the numbering. Please ignore the numbers and the same applies below.
\end{AnnotationRed}

\begin{lemma}
Let \(A_1, A_2\) be constant tensors and \(a\) be a constant matrix. Consider the system:
\begin{equation*}
    \begin{cases}
        \begin{aligned}
            &\nabla \cdot (A_1 \nabla \chi) = 0 \quad &&\text{in } B(1)_+, \\
&\nabla \cdot (A_2 \nabla \chi) + \nabla\cdot(a^T r) = 0 \quad &&\text{in } B(1)_-, \\
&\nabla \cdot (a \chi) = 0 \quad &&\text{in } B(1)_-, \\
&\chi|_+ = \chi|_- \quad &&\text{on } B(1) \cap \{x_d = 0\}, \\
&\left. \frac{\partial \chi}{\partial \nu} \right|_+ - \left. \frac{\partial \chi}{\partial \nu} \right|_- = g_0 \quad &&\text{on } B(1) \cap \{x_d = 0\}.
        \end{aligned}
    \end{cases}
\end{equation*}
where \(\chi, r\) are weak solutions.

Moreover, for all \(\rho \in (0, t)\), integrating the above inequalities yields:
\begin{equation*}
\left( \fint_{B(\rho)} |\chi - l|^2 \right)^{\frac{1}{2}} + \rho \left| g_0 - \zeta(l, q) \right| \le \left| \frac{\rho}{t} \right|^{\beta + 1} \left( \left( \fint_{B(t)} |\chi|^2 \right)^{\frac{1}{2}} + t |g_0- \zeta(l, q)| \right)
\end{equation*}

\begin{AnnotationGreen}[drop shadow][Note]
This derivation process yields a crucial conclusion and embodies the inherent logic of the inferential steps
\end{AnnotationGreen}

Thus, for all \((l', q') \in \mathcal{L}\), by the inequality above, we have:
\begin{align*}
&\inf_{\substack{(l,q) \in \mathcal{L} }} \left\{ \left( \fint_{B(\rho)} |\chi - l|^2 \right)^{\frac{1}{2}} + \rho \left| g_0 - \zeta(l, q) \right| \right\} \\
&\le C \left| \frac{\rho}{t} \right|^{\beta + 1} \inf_{\substack{(l,q) \in \mathcal{L} }} \left( \left( \fint_{B(t)} |\chi - l|^2 \right)^{\frac{1}{2}} + t |g_0- \zeta(l, q)| \right)
\end{align*}

Further, for all \((l', q') \in \mathcal{L}\), it follows that:
\begin{align*}
&\inf_{\substack{(l,q) \in \mathcal{L} }} \left\{ \left( \fint_{B(\rho)} |\chi - l|^2 \right)^{\frac{1}{2}} + \rho \left| g_0 - \zeta(l, q) \right| \right\} \\
&\le C \left| \frac{\rho}{t} \right|^{\beta + 1} \inf_{\substack{(l,q) \in \mathcal{L} }} \left( \left( \fint_{B(t)} |\chi - l|^2 \right)^{\frac{1}{2}} + t \left| g_0 - \zeta(l, q) \right| \right)
\end{align*}
\end{lemma}
\end{PromptBlock}

 Under this prompt, the output of AIM indicates that it proactively and appropriately incorporates the prompt information and executes the correct derivational procedure.
\begin{ResponseBlock}{Response of AIM}
\begin{AnnotationRed}[drop shadow][Disclaimer]
For clarity, we render formulas as symbols rather than raw LaTeX code. The same applies below.
\end{AnnotationRed}

\small
    **Step 2: Bounding \( \int_{B(t)} |\chi - l|^2 \, dx \)**  
From Lemma 4 and scaling arguments (see hint-5), we derive:
\[
\int_{B(t)} |\chi - l|^2 \, dx \leq C_2 t^{2(\beta +1)} \int_{B(t)} |\nabla \chi|^2 \, dx.
\]

**Step 3: Bounding \( \int_{B(t)_-} |r - q|^2 \, dx \)**  
Similarly, Lemma 4 gives:
\[
\int_{B(t)_-} |r - q|^2 \, dx \leq C_3 t^{2\beta} \int_{B(t)_-} |r|^2 \, dx.
\]

From Lemma 4 and scaling:
\[
\int_{B(t)} |\nabla l - \nabla \chi|^2 \, dx \leq C_4 t^{2\beta} \int_{B(t)} |\nabla \chi|^2 \, dx.
\]    
\end{ResponseBlock}

\subsubsection{Conceptual Guidance}
\label{sec:conceptual_guidance}
When analyzing the proof strategy for a given problem, AIM often exhibits vague or circular reasoning. In such cases, we---as human experts---can provide high-level framework prompts drawn from relevant mathematical fields. This allows the agents to conduct proof experiments within a conceptually sound structure, which can then be refined through iterative analysis of its outputs.

In our preliminary experiments, we prompted AIM to reason along established conceptual pathways---first performing a two-scale expansion, then deriving the corresponding cell problem, and finally carrying out the error estimation---while also specifying the expected form of the final conclusion (see the prompts bellow). Under this guidance, AIM generated several plausible intermediate results and proof strategies~\citep{AIM}.

\begin{PromptBlock}{Prompt to AIM}

    \noindent\textbf{Content:}

You can use the two-scale expansion method to derive the cell problem and then use the cell problem to define or study the homogenized equation.
\begin{AnnotationGreen}[drop shadow][Note]
    Here we provide guidance on the commonly used treatment methods for homogenization problems and the subsequent processing directions.
\end{AnnotationGreen}

    {\it \color{gray} \vspace{1em}Some prompts are omitted here for clarity.\vspace{1em} }
    
\noindent\textbf{Content:}

The error estimate between the original solution and the homogenized solution has the form
\[
  C\,\varepsilon^{\alpha}\,\|g\|_{H^{-1/2}(\partial\Omega)}
  \quad
  \text{for some }\alpha\in(0,1).
\]
\begin{AnnotationGreen}[drop shadow][Note]  
    Herein, we outline the form of conclusions to be demonstrated.
\end{AnnotationGreen}
\end{PromptBlock}

Similarly, we observed that in AIM’s initial experiments, it proposed an estimate form of the error:
\begin{equation}
    u_\varepsilon^{\rm app}(x)
  = u_0(x) + \varepsilon\,\chi^{D_xu_0(x)}\bigl(\tfrac{x}{\varepsilon}\bigr)
  \quad\text{in }\Omega_\varepsilon.
\end{equation}

Upon careful analysis of the experimental results, it became evident that the proposed form was inaccurate, and AIM's derivation process lacked sufficient detail. 
\begin{ResponseBlock}{Response of AIM without guidance}
    By the coercivity and the conjecture:
\[
c \|\mathbf{e}\|_{H^1(\Omega)}^2 \leq a(\mathbf{e}, \mathbf{e}) = |\text{Res}(\mathbf{e})| \leq C \varepsilon^{\frac{1}{2}} \|\mathbf{u}_0\|_{H^1(\Omega)} \|\mathbf{e}\|_{H^1(\Omega)}.
\]
Dividing by \(\|\mathbf{e}\|_{H^1(\Omega)}\):
\[
\|\mathbf{e}\|_{H^1(\Omega)} \leq \frac{C}{c} \varepsilon^{\frac{1}{2}} \|\mathbf{u}_0\|_{H^1(\Omega)}.
\]
For the homogenized problem with Neumann condition \(\left.\frac{\partial \mathbf{u}_0}{\partial \nu^*}\right|_{\partial \Omega} = g\), the solution satisfies:
\[
\inf_{\mathbf{r} \in \mathbb{R}} \|\mathbf{u}_0 - \mathbf{r}\|_{H^1(\Omega)} \leq C' \|g\|_{H^{-\frac{1}{2}}(\partial \Omega)}.
\]
Since \(\mathbf{u}_0\) is already orthogonal to \(\mathbb{R}\):
\[
\|\mathbf{u}_0\|_{H^1(\Omega)} \leq C' \|g\|_{H^{-\frac{1}{2}}(\partial \Omega)}.
\]
Combining:
\[
\|\mathbf{u}_0 - \mathbf{u}_\varepsilon\|_{H^1(\Omega)} \leq \left( \frac{C}{c} C' \right) \varepsilon^{\frac{1}{2}} \|g\|_{H^{-\frac{1}{2}}(\partial \Omega)}.
\]
Thus, the estimate holds with \(C_{\text{final}} = \frac{C}{c} C'\).
\end{ResponseBlock}

Therefore, we revisited the problem from the perspective of homogenization error and, in subsequent proof attempts, provided a mathematically sound formulation to serve as the estimated error form and use it as guiding information to prompt AIM.

\pagebreak
\begin{PromptBlock}{Prompt to AIM}
\textbf{Content}: 
Consider the following form of the error.

    Define the cut-off function as follows: \( \eta_{\varepsilon} \in C_0^{\infty}(\Omega) \),and
\[
0 \le \eta_{\varepsilon} \le 1, \quad |\nabla \eta_{\varepsilon}| \le \frac{C}{\varepsilon},
\]
\[
\eta_{\varepsilon} = 1, \quad \text{when } \mathrm{dist}(x, \partial \Omega) \ge 4\varepsilon,
\]
\[
\eta_{\varepsilon} = 0, \quad \text{when } \mathrm{dist}(x, \partial \Omega) \le 3\varepsilon.
\]

 And we also set this convolution operator:
\[
S_{\varepsilon}(f)(x) = \rho_{\varepsilon} \ast f(x) = \int_{\mathbb{R}^d} f(x-y)\rho_{\varepsilon} \, dy,
\]
Here \( \rho \in C_0^{\infty}(B(0, \frac{1}{2})) \), \( \rho \ge 0 \), and \( \int_{\mathbb{R}^d} \rho \, dx = 1 \).

Suppose
\[
w_{\varepsilon} = u_{\varepsilon} - u_0 - \varepsilon \chi\left(\frac{x}{\varepsilon}\right) \eta_{\varepsilon} S_{\varepsilon}^2(\nabla u_0).
\]

\begin{AnnotationGreen}[drop shadow][Note]
    Here, we present the error form of the homogenization problem, constructed via convolution.
\end{AnnotationGreen}
\end{PromptBlock}

After we provided this clearer and more accurate form of error estimation, AIM also offered a more reasonable proof process.
\begin{ResponseBlock}{Response of AIM with guidance}
    Since \( \eta_\varepsilon \equiv 1 \) on \( D \) and \( 0 \leq \eta_\varepsilon \leq 1 \) on \( C \), we estimate each piece separately.

1. Interior region \( D = \Omega \setminus \Omega_{5\varepsilon} \). Here \( \eta_\varepsilon \equiv 1 \), so
   \[ \eta_\varepsilon S_\varepsilon^2 (\nabla u_0) - \nabla u_0 = S_\varepsilon^2 (\nabla u_0) - \nabla u_0. \]
Using \( |M| \leq C \) and Cauchy--Schwarz,
  \[ |J_D| \leq C \int_{D} |S_\varepsilon^2 (\nabla u_0) - \nabla u_0| \, |\nabla \psi| \leq C \, \| S_\varepsilon^2 (\nabla u_0) - \nabla u_0 \|_{L^2(D)} \, \|\nabla \psi\|_{L^2(D)}. \]
Since \( D \subset \Omega \setminus \Omega_{2\varepsilon} \), the standard mollifier-estimate (Lemma 5) yields
  \[ \| S_\varepsilon^2 (\nabla u_0) - \nabla u_0 \|_{L^2(D)} \leq \| \nabla u_0 - S_\varepsilon (\nabla u_0) \|_{L^2(\Omega \setminus \Omega_{2\varepsilon})}, \]
and clearly \( \|\nabla \psi\|_{L^2(D)} \leq \|\nabla \psi\|_{L^2(\Omega)} \). Hence
  \[ |J_D| \leq C \, \| \nabla u_0 - S_\varepsilon (\nabla u_0) \|_{L^2(\Omega \setminus \Omega_{2\varepsilon})} \, \|\nabla \psi\|_{L^2(\Omega)}. \]

2. Boundary-collar \( C = \Omega_{5\varepsilon} \). On \( C \) we only know \( 0 \leq \eta_\varepsilon \leq 1 \), so
  \[ |\eta_\varepsilon S_\varepsilon^2 (\nabla u_0) - \nabla u_0| \leq |S_\varepsilon^2 (\nabla u_0)| + |\nabla u_0| \leq C \, |\nabla u_0| \]
(by boundedness of the mollifier in \( L^2 \)). Hence
  \[ |J_C| \leq C \int_{C} |\nabla u_0| \, |\nabla \psi| \leq C \, \|\nabla u_0\|_{L^2(C)} \, \|\nabla \psi\|_{L^2(C)} = C \, \|\nabla u_0\|_{L^2(\Omega_{5\varepsilon})} \, \|\nabla \psi\|_{L^2(\Omega_{5\varepsilon})}. \]
Finally, since \( \Omega_{5\varepsilon} \subset \Omega \) we have \( \|\nabla \psi\|_{L^2(\Omega_{5\varepsilon})} \leq \|\nabla \psi\|_{L^2(\Omega)} \), so
  \[ |J_C| \leq C \, \|\nabla u_0\|_{L^2(\Omega_{5\varepsilon})} \, \|\nabla \psi\|_{L^2(\Omega)}. \]
\end{ResponseBlock}

\subsubsection{Detail Refinement}
\label{sec:detail_refinement}
For reasoning details generated by AIM---such as formula derivations, symbol definitions, and computational steps---directly identifying deviations and providing corrective guidance is an effective strategy for local error calibration. Similarly, proof accuracy can be enhanced by prompting targeted revisions within specific segments of the argument. Once an error is detected, it can be incorporated into the prompt for subsequent experiments, thereby progressively improving the validity of the conclusions.

In our equation derivation experiments, we frequently observed that AIM misinterpreted symbolic notation. To address this, we explicitly included symbol definitions within the input prompts during each iteration, specifying their precise mathematical meaning. This adjustment significantly reduced such errors in subsequent reasoning steps.

\begin{PromptBlock}{Prompt to AIM}
    \textbf{Content}:   A pair of real numbers \((\lambda, \mu)\) are called admissible and referred to as a $\text{Lam\'e}$ pair, if they satisfy
    \( \mu > 0 \quad \text{and} \quad d\lambda + 2\mu > 0.\)
    For a $\text{Lam\'e}$ pair\((\lambda, \mu)\), the elastostatic system ($\text{Lam\'e}$ system) reads
    \[  \mathcal{L}_{\lambda,\mu}\mathbf{u} := \mu\Delta\mathbf{u} + (\lambda + \mu)\nabla\text{div}\,\mathbf{u},
   \]
        
    where \( \mathbf{u} = (u^1, \ldots, u^d) \) represents the displacement field and the divergence of \(\mathbf{u}\) is given by
    \[
    \text{div}\,\mathbf{u} = \sum_{i=1}^{d} \frac{\partial u^i}{\partial x_i}.
    \]
    \begin{AnnotationGreen}[drop shadow][Note]
        The operational rules are emphasized here through specific formulas.
    \end{AnnotationGreen}
\end{PromptBlock}

\subsection{Theory-Coordinated Application}
\label{sec:theory_coordinated_application}

The Theory-Coordinated Application paradigm is an interaction mode that constrains the agents to operate within a pre-established theoretical framework to conduct derivations or proofs. It is intended for relatively complex 
intermediate problems where the solution path is foreseeable but detailed reasoning is nontrivial. The division of responsibilities is as follows: the human assembles a knowledge package containing precise definitions, lemmas, usable theorems with their applicability conditions, and permitted inference rules, and also specifies the target proposition and scope constraints. The model performs the derivation within this scope, proceeding step by step by referencing the knowledge package as needed and reasoning in accordance with the inferential logic inherent in the given theory.
\begin{itemize}
    \item \textbf{Application Conditions}: This paradigm is appropriate when the problem is relatively complex and demands a multi-step chain of reasoning.
The paradigm should be used once a suitable theoretical framework has been identified for the problem.
The human can encode the theory's prior knowledge---definitions, lemmas, and the theory's inherent derivation logic or inference schemata---into a knowledge package that the model will follow.
\item \textbf{Typical Mathematical Scenarios}: 
 During the actual proof process, intermediate problems often arise that require resolution. These may take the form of certain properties or steps identified during the proof experiment that need to be addressed.
This paradigm targets scenarios where the final theorem or proposition can be obtained by progressively applying the selected theory's intrinsic logic, moving stepwise from definitions and lemmas to the desired conclusion. Representative contexts include any domain in which a well-chosen theory provides a structured path to the result.
\item \textbf{Expected Outcomes}: Final theorem proof derived strictly from the provided prior theoretical knowledge: AIM arrives at the target proposition by applying only the definitions, lemmas, and rules encoded in the knowledge package.
The reasoning process is layered and progressive, unfolding in well-structured stages where each step is justified by the intrinsic inferential logic of the theory and accompanied by explicit condition checks. Throughout the process, AIM demonstrates a correct grasp of the derivation logic, faithfully applying the theory’s proof patterns to advance from premises to the theorem without relying on heuristic leaps.
\end{itemize}

During the proof of the homogenization problem, AIM proposed a property pertaining to the Cell Problem: \(\chi \in W^{1,\infty} (\omega)\). Analytical judgment suggested this result is likely correct and essential for the error estimation component. Consequently, multiple theoretical approaches were evaluated to address this intermediate property. Ultimately, AIM successfully constructed a proof by applying Schauder theory, following some appropriate instruction. Specifically, the knowledge package on Schauder theory is provided to AIM as follows:

\begin{PromptBlock}{Prompt to AIM: Some prior knowledge of the entire Schauder theory}

\textbf{Content}: 
Lemma 1: Suppose \(\Omega_{\pm}=\mathbb{R}^d_{\pm}\), \(S=\{x_d=0\}\), \(B_{+}=\{x \in B(1):x_d>0\}\) and \(B_{-}=\{x \in B(1):x_d<0\}\). Here \(B(1)=\{\|x\| \le 1\}\). Consider this equation: for \(V \in H_0^1(B(1);\mathbb{R}^d)\)
\[
(\nabla V :A_1 \nabla \tilde{\chi})_{B_{+}}+(\nabla V :A_2 \nabla \tilde{\chi})_{B_{-}}+(\tilde{r},\nabla \cdot(aV))_{B_{-}}=0,
\]
\[
\nabla \cdot(a \tilde{\chi})=0,
\]
here \(\tilde{\chi}=D^{\alpha}\chi\), \(\tilde{r}=D^{\alpha} r\), \(|\alpha| \ge 1\) and \(A_1, A_2\) are constant tensors, \(a\) is a constant matrix. We have that \(\sum_{\pm} \| \chi\|_{H^k(B(\frac{1}{2},\pm))} \le C\| \chi\|_{L^2(B(1))}\) and \(\| r\|_{H^k(B(\frac{1}{2})_-)} \le C\| r\|_{L^2(B(\frac{1}{2})_-)}\) for \(\forall k \ge 1\).

\textbf{Content}: 
Lemma 2: Suppose that \(M_{\pm}\) is the constant matrix in \(\mathbb{R}^{d \times d}\), the following are equivalent:
\[\forall y \in \{y_d=0\}, M_{+}x = M_{-}x,\]
\[\exists c \in \mathbb{R}^d, \text{s.t. } M_{+} - M_{-} = c e_d^T,\]
\[(I - e_d^T e_d) M_{+} = (I - e_d^T e_d) M_{-}.\]

\textbf{Content}: 
Definition: \( A_1, A_2 \) are tensor constant, \( a \) is matrix constant. If \( M_{\pm} \) satisfy the above lemma, and \( \nabla \cdot (a M_{-} y) = 0 \quad \text{in} \quad B(t)_{-} \). We let \( l(y) = M_{+} y 1_{y \ge 0} + M_{-} y 1_{y \le 0} + C \), \( q(y) = r(0) \). We call \( l, q \) the piecewise linear solution of the following equation:
\[ \nabla \cdot (A_1 \nabla l) = 0 \quad \text{in} \quad B(t)_{+}; \]
\[ \nabla \cdot (A_2 \nabla l) + \nabla\cdot (a^T q) = 0, \quad \nabla \cdot (a l) = 0, \quad \text{in} \quad B(t)_{-}; \]
\[ l_{+} = l_{-}; \quad \frac{\partial l}{\partial \nu}\bigg|_{+} - \frac{\partial l}{\partial \nu}\bigg|_{-} = (A_1 M_{+}) e_d - (A_2 M_{-} +a^T  r(0)) e_d, \quad \text{on} \quad \partial B(t). \]
Suppose that \( \mathcal{L} \) is the space of all the piecewise linear solutions of the above equation. And \( \forall (l, q) \in \mathcal{L} \), we define \( \zeta(l, q) = \left( \frac{\partial l}{\partial \nu} \right)_{+} - \left( \frac{\partial l}{\partial \nu} \right)_{-} \).
\begin{AnnotationGreen}[drop shadow][Note]
    The prerequisite knowledge of the theory has a logical deductive relationship with one another.
\end{AnnotationGreen}

\textbf{content}: 
Lemma 3: \(A_1,A_2\) are the tensor constant, \(a\) is the matrix constant.
\[
\nabla \cdot (A_1\nabla \chi) = 0 \quad \text{in} \quad B(1)_{+};
\]
\[
\nabla \cdot (A_2\nabla \chi) +\nabla \cdot (a^T  r) = 0, \quad \nabla \cdot (a\chi) = 0, \quad \text{in} \quad B(1)_{-};
\]
\[
\chi_{+} = \chi_{-}; \quad \left.\frac{\partial \chi}{\partial \nu}\right|_{+} - \left.\frac{\partial \chi}{\partial \nu}\right|_{-} = g_0, \quad \text{on} \quad B(1) \cap \{x_d = 0\}.
\]
\(\chi, r\) are the weak solutions of the above equation. Then for \(\forall k \ge 0, \alpha \in [0,1]\), we have
\[
\sum_{\pm} \Vert \chi \Vert_{H^k(B(\frac{1}{2},\pm))} \le C \left( \Vert \chi \Vert_{L^2(B(1))} + \vert g_0 \vert \right).
\]
    
\end{PromptBlock}

Based on these established lemmas, we utilized AIM to verify the required theorem. As demonstrated by the experimental results below, AIM appropriately applied the lemmas and provided a correct derivation process throughout the proof.

\begin{ResponseBlock}{Response of AIM}
We follow the strategy already sketched in Lemma 5. Fix \(0 < \rho < t \leq 1/2\) and let \((l^*, q^*) \in \mathcal{L}\) be an almost-minimizer at scale \(t\), namely

\[
E(t) \leq \left( \int_{B(t)} |\chi - l^*|^2 \right)^{1/2} 
+ t \cdot |g_0 - \zeta(l^*, q^*)| 
+ \left( \int_{B(t)_-} |r - q^*|^2 \right)^{1/2} 
+ \varepsilon,
\]

where \(\varepsilon > 0\) can be made arbitrarily small. We will estimate \(E(\rho)\) using the same \((l^*, q^*)\). By Lemma 5 (the integrated \(C^{1,\beta}\)-approximation on \(B(t)\)) we have for this fixed \((l^*, q^*)\):

1)
\[
\begin{split}
&\left( \int_{B(\rho)} |\chi - l^*|^2 \right)^{1/2} + \rho \cdot |g_0 - \zeta(l^*, q^*)| \\
&\leq C \cdot \left( \frac{\rho}{t} \right)^{\beta + 1} \cdot \left[ 
\left( \int_{B(t)} |\chi - l^*|^2 \right)^{1/2} + t \cdot |g_0 - \zeta(l^*, q^*)| 
\right].
\end{split}
\]
2)
\[
\left( \int_{B(\rho)_-} |r - q^*|^2 \right)^{1/2} 
\leq C \cdot \left( \frac{\rho}{t} \right)^{\beta + 1} \cdot \left( \int_{B(t)_-} |r - q^*|^2 \right)^{1/2}.
\]
\end{ResponseBlock}

Furthermore, AIM ultimately completed the entire proof process through the appropriate application of this theoretical framework.
\begin{ResponseBlock}{Response of AIM}
    3. Campanato embedding $\rightarrow$ one-sided $C^{1,(\alpha+\delta)/2}$.  
   The estimate  
     $\int_{B(\rho)}|\chi-l_{\rho}|^{2}  \leq  C''\,\rho^{d+2+\alpha+\delta}  $   shows that on each half-ball $B(\rho)\cap\{\pm x_{d}\geq 0\}$ the mean oscillation of $\chi-l_{\rho}$ is $O(\rho^{1+(\alpha+\delta)/2})$. By the Campanato characterization of H\"older spaces, one deduces for each sign $\pm$  
     $\chi\in C^{1,(\alpha+\delta)/2}\left( \overline{B_{\frac{1}{2},\pm}} \right)$,  
   with norm bound $\leq C \left[\Phi(\tfrac{1}{2})+\Psi(\tfrac{1}{2})\right]^{\frac{1}{2}}$.  

4. Iterative bootstrap to arbitrary $\gamma<1$.  
   Having now $C^{1,\gamma_{0}}$--control with $\gamma_{0}=(\alpha+\delta)/2\gamma_{0}$, whence Campanato$\rightarrow C^{1,\gamma_{1}/2}$ etc. Standard iteration shows that the sequence of exponents approaches $1$. Hence for any target $\gamma<1$ we may in finitely many steps reach $\chi\in C^{1,\gamma}\left( \overline{B_{\frac{1}{2},\pm}} \right)$.  

This completes the proof that $\chi\in C^{1,\gamma}$ on each closed half--ball up to the flat interface, for every $\gamma<1$. 
\end{ResponseBlock}

This case illustrates that when a well-defined theoretical framework supports a conclusion, its lemmas can be directly utilized to formulate the proof. Under these conditions, AIM demonstrates coherent understanding and systematic application of both the provided lemmas and the ensuing proof strategy, thereby enhancing the probability of obtaining a complete and valid argument. The supplied definitions and lemmas aid the agents in clarifying the logical structure of the proof. During the final verification of the theorem, this logical framework can also be applied to streamline the proof process and refine deductive details. The result is a controlled, progressive derivation in which every step is justified by explicit knowledge package citations and condition checks, and any missing prerequisites are recorded as gaps rather than bridged by heuristic leaps. This approach shifts effort from unguided search to deliberate theory selection and packaging, yielding proofs that are more reliable, reproducible, and amenable to formal verification.

\subsection{Interactive Iteration}
\label{sec:interactive_iteration}
In practical applications, obtaining a complete and correct mathematical proof solely through agents often remains challenging when relying exclusively on the methods described above. These limitations stem from the inherent reasoning constraints of LLMs, which necessitate the development of more advanced interactive paradigms.

Interactive Iteration is a paradigm for uncertain and complex problem settings built around a cyclic process: output $\rightarrow$ human diagnosis and correction $\rightarrow$ model optimization $\rightarrow$ output  $\rightarrow$ $\cdots$. By supplying only the minimal necessary incremental information, performing condition checks, and validating with small examples, it avoids overcommitting to incorrect lines of reasoning and enables the reuse of validated fragments as stable modules. The approach requires careful human inspection of AIM's outputs and evaluation of the derivation process, with concrete revision suggestions and directions for iterative adjustments, to better guide subsequent proof work.

Each experiment conducted with AIM must be rigorously analyzed, with emphasis placed on distinguishing between plausible intermediate results and unsound derivations. Incorrect applications of theorems or invalid inference steps should be corrected and reformulated into updated prompts for the next iteration. Conversely, valid proof segments should be retained and reused as established building blocks in subsequent interactions. This iterative refinement helps decompose complex proofs into tractable components, thereby improving both efficiency and rigor. Through repeated cycles of adjustment, the proof is progressively elaborated and verified, ultimately yielding a complete and logically sound derivation.

\begin{itemize}
    \item \textbf{Application Conditions}: We adopt an iterative interactive refinement mode when problem complexity precludes a one-shot solution, such as in multi-stage derivations, cross-domain dependencies, or tightly coupled constraints that necessitate phased exploration. This regime is also indicated when the core reasoning pathway or the final conclusion is uncertain—e.g., missing intermediate lemmas, competing hypotheses, or underspecified objectives—or when there is no clear initial entry point and multiple candidate strategies must be generated, compared, and pruned.
    \item \textbf{Typical Mathematical Scenarios}: In a typical mathematical scenario involving complex proofs, we often lack a clear and complete proof strategy and possess only preliminary derivation directions. In such cases, agent-generated proofs may contain conspicuous error such as gaps in reasoning, improper theorem invocation and ignored applicability conditions, which compromise correctness. To remedy this, we can adopt this approach to adjust and repair the proof route, jointly decomposing the objective, auditing preconditions for theorems, introducing necessary intermediate lemmas, localizing and isolating faulty steps, and validating proposed fixes on small or simplified instances before generalization. Through this iterative co-working loop, an ambiguous starting point is transformed into a disciplined proof search, with progressively refined hypotheses, judicious tool selection, and explicit checkpoints and acceptance criteria, ultimately converging to a sound, transparent, and reproducible derivation.
    \item \textbf{Expected Outcomes}: The expected outcomes of each iterative cycle are structured around error correction and direction guidance, forming a mechanism that drives the gradual refinement of proofs. The specific content is as follows: First, AIM shall output a precisely targeted and revised derivation process. This output must respond specifically to the errors identified through manual diagnosis in the previous iteration, and attach verifiable bases to each step of the derivation to ensure rigor. Second, based on AIM's revised results, we need to extract guidance content for the next iteration. This guidance must accomplish three core tasks: (1) Identify verified stable modules (e.g., intermediate lemmas) and clarify their position in the overall proof as well as scenarios for reuse. (2) Analyze the unfinished segments and proof directions in the current derivation to determine the breakthrough direction for the next iteration. (3) Transform the guidance into specific executable instructions for AIM, ensuring that the next round of derivation proceeds along the correct path and gradually achieves the completeness and logical consistency of the mathematical proof.

\end{itemize}

When addressing core aspects of the homogenization problem—such as the cell problem and error estimation—significant challenges arose. Employing the interactive iterative approach, we systematically broke down and incrementally verified the proof structures for these key results.

In initial experiments related to error estimation, AIM was prompted with the error formulation. Analysis of its output revealed the following intermediate conclusions and derivations proposed by AIM:
\begin{ResponseBlock}{Response of AIM}
    \begin{lemma}
Let $\Omega \subset \mathbb{R}^d$ be a bounded Lipschitz domain and let $u_0 \in H^2(\Omega)$. 
Let $\eta_\varepsilon$ and $S_\varepsilon$ be as in Lemmas 5--6. Assume furthermore that the 
coefficient $A(y)$ is 1--periodic, uniformly elliptic, bounded, and lies in $H^\mu(Y;\mathbb{R}^{d \times d})$ 
for some $\mu > d/2$. Denote its homogenized matrix by  
\[
\hat{A} = \int_Y A(y)\diff y,
\]  
and for $t > 0$ set the boundary collars  
\[
\Omega_t = \{x \in \Omega : \mathrm{dist}(x, \partial \Omega) < t\}.
\]  

Then there exist $\varepsilon_0 > 0$ and a constant $C$, depending only on $d$, $\Omega$, 
$\|A\|_{H^\mu(Y)}$, and the constructions of $\eta_\varepsilon$ and $S_\varepsilon$, such that 
for every $\psi \in H_0^1(\Omega;\mathbb{R}^d)$ and every $\varepsilon \in (0, \varepsilon_0)$ 
one has the estimate  
\[
\begin{split}
&\left| \int_\Omega (\hat{A} - A(x/\varepsilon)) 
  (\nabla u_0 - \eta_\varepsilon S_\varepsilon^2(\nabla u_0)) : \nabla \psi \, dx \right|  
\leq C \left[ \varepsilon^{1/2} \|u_0\|_{H^2(\Omega)} \right. \\
&\quad \left. + \|\nabla u_0 - S_\varepsilon(\nabla u_0)\|_{L^2(\Omega \setminus \Omega_{2\varepsilon})} \right] 
  \|\nabla \psi\|_{L^2(\Omega)}.
\end{split}
\]
\end{lemma}
\begin{AnnotationGreen}[drop shadow][Note]
    The conclusion here is what we care about, and it also provides inspiration for our subsequent analysis and decomposition of the problem.
\end{AnnotationGreen}
\end{ResponseBlock}

We observed that the experimental results yielded a control conclusion regarding \(\nabla w_\varepsilon = \nabla u_0 - \eta_\varepsilon S_\varepsilon^2 (\nabla u_0)\). After examining the derivation process provided by AIM, we determined that this constitutes a valid intermediate conclusion and aligns with the expected direction of the proof.

Meanwhile, AIM also proposed the following lemma:
\begin{ResponseBlock}{Response of AIM}
\begin{proof}
Since $A$ and $\chi$ are bounded, 
\[
\left|\int_{\varepsilon} A \chi \nabla(\eta_{\varepsilon} S_{\varepsilon}^2 \nabla u_0) : \nabla \psi \right|
\leq C \varepsilon \|\nabla(\eta_{\varepsilon} S_{\varepsilon}^2 \nabla u_0)\|_{L^2(\Omega)} \|\nabla \psi\|_{L^2(\Omega)}.
\]
By Lemma 6,
\[
\|\nabla(\eta_{\varepsilon} S_{\varepsilon}^2 \nabla u_0)\|_{L^2(\Omega)}
\leq C \left\{ \|S_{\varepsilon}(\nabla^2 u_0)\|_{L^2(\Omega \setminus \Omega_{2\varepsilon})} + \varepsilon^{-1} \|\nabla u_0\|_{L^2(\Omega_{5\varepsilon})} \right\}.
\]
Hence
\[
\left| \dots \right|
\leq C \varepsilon \left\{ \|S_{\varepsilon}(\nabla^2 u_0)\|_{L^2(\Omega \setminus \Omega_{2\varepsilon})} + \varepsilon^{-1} \|\nabla u_0\|_{L^2(\Omega_{5\varepsilon})} \right\} \|\nabla \psi\|_{L^2(\Omega)},
\]
as claimed.
\end{proof}
\end{ResponseBlock}

However, the derivation process here is noticeably lacking in detail and substance. While analyzing this derivation and the lemmas provided above, we summarized and formulated the following conclusion:
\begin{maincontentlemma}

Suppose $\Omega$ is a bounded Lipschitz domain in $\mathbb{R}^d$. Define the $\varepsilon$-neighborhood of the boundary as
\[
\Omega_t = \left\{ x \in \Omega : dis(x, \partial \Omega) < t \right\}, \quad t > 0.
\]
For all $\psi \in H_0^1(\Omega, \mathbb{R}^d)$, the following estimate holds:
\[
\begin{split}
\left| \int_{\Omega} A\nabla w_{\varepsilon} : \nabla \psi \, dx \right| 
&\leq C \|\nabla \chi\|_{L^2(\Omega)} \left\{ 
    \varepsilon \left\| S_{\varepsilon}(\nabla^2 u_0) \right\|_{L^2(\Omega \setminus \Omega_{3\varepsilon})} 
    + \left\| \nabla u_0 - S_{\varepsilon}(\nabla u_0) \right\|_{L^2(\Omega \setminus \Omega_{2\varepsilon})}
\right\} \\
&\quad + C \left\| \nabla \psi \right\|_{L^2(\Omega_{4\varepsilon})} \left\| \nabla u_0 \right\|_{L^2(\Omega_{5\varepsilon})}.
\end{split}
\]
\end{maincontentlemma}
We performed a certain amount of manual, straightforward derivation and determined that the following lemma is correct. Consequently, we provided this lemma to AIM, which then produced a correct derivation process.

Through this interactive process that combines experimental exploration with manual derivation, we arrived at the correct conclusion. The lemma served as a crucial intermediate step in our error estimation argument, simplifying the subsequent proof and guiding AIM to reason about the specific form of the error estimate. With this lemma as context, AIM’s ensuing derivation became more direct and clearly structured, leading to a coherent completion of the overall argument.

\subsection{Applicability Boundary and Exclusion Domain}
\label{sec:applicability_boundary_exclusion_domain}

For LLMs, some certain mathematical processes can be challenging to comprehend and execute accurately. However, these processes may not necessarily be complex or intuitively understandable for human mathematicians. Examples include handling intricate geometric configurations, deconstructing intermediate proof strategies, or addressing difficulties in mathematical derivations.  
When confronted with such issues, attempting to forcefully resolve them solely through automated agents may lead to a case of more effort for less result, consuming substantial resources and time while still failing to yield reasonable conclusions. Therefore, it is advisable to address these processes through human intervention, thereby enhancing both the efficiency and effectiveness of the overall theoretical proof.
The following are two examples we encountered for illustration.

\subsubsection{Example 1: Complex Symbolic Reasoning}
\label{sec:complex_symbolic_reasoning}
First, we show the computational bottlenecks faced by AIM in performing multi-scale expansions. This process requires applying the two-scale expansion to the original equation, which involves a second scale \(y = \frac{x}{\varepsilon}\). The solution \(u_{\varepsilon}(x) = u_{\varepsilon}(x, y)\) is then expanded multidimensionally, followed by the derivation of equations at different scales. For mathematicians, this task is essentially a process of symbolic reasoning, which requires only time to perform the calculations.  

However, when the corresponding operational information and requirements are provided to AIM, it encounters the following bottlenecks and errors: failure to comprehend the multi-scale expansion process leading to computational mistakes, and after performing certain derivations, the agent begins considering some content unrelated to the intended proof direction.
\begin{ResponseBlock}{Response of AIM}
    \begin{lemma}

In the perforated (fluid) inclusion $D_{\varepsilon}$ the next-order pressure corrector $p_{0}^{-}(x,y)$ is not in general $y$--independent, but rather admits the representation
\[
p_{0}^{-}(x,y) = \sum_{k=1}^{d} \pi^{k}(y) \, \partial_{x_k} p_{-1}(x),
\]
where for each coordinate direction $k$ the pair $(w^{k}, \pi^{k}) \in [H^{1}_{\text{per}}(\omega)]^{d} \times L^{2}_{\text{per}}(\omega)/\mathbb{R}$ is the unique solution of the ``cell--Stokes'' system
\[
-\tilde{\mu} \, \Delta_{y} w^{k} + \nabla_{y} \pi^{k} = e_{k} \quad \text{in } \omega,
\]
\[
\operatorname{div}_{y} w^{k} = 0 \quad \text{in } \omega,
\]
\[
w^{k} = 0 \quad \text{on } \partial\omega,
\]
with $Y$--periodicity on $\partial Y$ and $\int_{\omega} \pi^{k} \, dy = 0$.
\end{lemma}
\begin{AnnotationGreen}[drop shadow][Note]
Incorrect cell system and expansion expression!
\end{AnnotationGreen}
\end{ResponseBlock}

\begin{ResponseBlock}{Response of AIM}
3. (Representation via fundamental strains.)  
By linearity of the cell-problem in the macroscopic gradient $\nabla_{x} u_{0}^{+}$, one constructs for each unit strain $e_{p} \otimes e_{q}$ the unique periodic corrector $\chi^{pq}(y)$ solving  
\[
-\nabla_{y} \cdot \sigma_{y}(\chi^{pq}) = 0 \quad \text{in } Y_{f},
\]  
\[
\sigma_{y}(\chi^{pq}) \cdot n = -(C : e_{p} \otimes e_{q}) \, n \quad \text{on } \partial \omega,
\]  
\[
\chi^{pq} \text{ periodic}, \quad \int_{Y_{f}} \chi^{pq} \, dy = 0.
\]  
Then by superposition the general first-order corrector is  
\[
u_{1}^{+}(x,y) = -\sum_{p,q} \chi^{pq}(y) \, \partial_{x_{q}} u_{0}^{+p}(x),
\]  
\begin{AnnotationGreen}[drop shadow][Note]
Incorrect and expansion expression!
\end{AnnotationGreen}
which is the asserted result.
\end{ResponseBlock}

\subsubsection{Example 2: Constructive Problems Related to Geometric Setup}
\label{sec:constructive_problems_related_to_geometric_setup}
In the proof process of nearly every homogenization problem, there exists a component related to constructing the cell problem. This problem serves as a bridge connecting the microscopic and macroscopic scales. By focusing on each unit cell within the microstructure, we formulate and solve local cell problems to extract parameter information, which in turn allows us to derive the macroscopic homogenized equations.  

This requires an understanding of how the equations behave at the geometric boundaries of the cell and the construction of corresponding intra-cell equations based on the expanded forms. The primary difficulty for agents in such tasks lies in their insufficient ability to comprehend geometric structures, which often leads to errors in both the formulation and derivation of the whole problem.
\begin{ResponseBlock}{Response of AIM}
 
     For each fixed symmetric matrix \(E\in\mathbb{R}^{d\times d}_{\rm sym}\), find
\[
  (\chi_e^E,\;\chi_i^E,\;\pi^E)
  \;\in\;
  H^1_{\rm per}(Y_e;\mathbb{R}^d)
  \;\times\;
  H^1_{\rm per}(Y_i;\mathbb{R}^d)
  \;\times\;
  L^2_{\rm per}(Y_i)/\mathbb{R}
\]
solving
\begin{align*}
  &\text{(1) in the elastic cell }Y_e:&&
    -\Div_y\bigl[\mathbb{C}\bigl(E + D_y\chi_e^E\bigr)\bigr]=0,
    \quad
    \Div_y\bigl(Ey + \chi_e^E\bigr)=0,\\
  &\text{(2) in the fluid cell }Y_i:&&
    -\mu\,\Delta_y\chi_i^E + \nabla_y\pi^E=0,
    \quad
    \Div_y\bigl(Ey + \chi_i^E\bigr)=0,\\
  &\text{(3) on the interface }\Gamma=\partial Y_i:&&
    \chi_e^E = \chi_i^E,
    \quad
    \bigl[\mathbb{C}\bigl(E+D_y\chi_e^E\bigr)\,n\bigr]
    = \bigl[2\mu\,D_y\chi_i^E - \pi^E I\bigr]\,n,\\
  &\text{(4) periodicity and zero‐mean:}&&
    (Ey+\chi^E)\;\text{is }Y\text{‐periodic,}\quad
    \int_{Y_e}\chi_e^E\,dy = 0,\;
    \int_{Y_i}\chi_i^E\,dy = 0.
\end{align*}
\end{ResponseBlock}
Here we can observe a clear mathematical error:  
\[
\text{in the elastic cell } Y_e: \quad \Div_y \bigl( E y + \chi_e^E \bigr) = 0.
\]  
This error further leads to significant challenges for AIM in deriving a reasonable and correct form of the homogenization equation.  For such geometrically dependent constructive problems, direct human-driven derivation and analysis may represent a more efficient and reliable approach.

\subsection{Auxiliary Optimization Methods }
\label{sec:auxiliary_optimization_methods}
Beyond core strategies such as \textbf{Direct Prompting} and \textbf{Interactive Iteration Refinement}, we employed several supplementary techniques to enhance the reliability of AIM's 
 mathematical output.

\begin{itemize}
    \item \textbf{Repeated Attempts and Proof Screening.}  
    Owing to the inherent stochasticity of LLMs, the same query often yields varying responses. For mathematical problems requiring extended reasoning, multiple trials can produce divergent proof attempts. By repeatedly testing identical problems, we filtered the generated proofs and selected the most complete and coherent versions for further refinement.

    \item \textbf{Providing Target Conclusions to Improve Correctness.}  
   We also found that providing the target conclusion, rather than posing fully open-ended proof tasks, materially improves correctness by constraining the search space and guiding the reasoning path. For instance, in error estimation experiments, prior mathematical intuition suggested a likely order for the scaling parameter alpha. AIM initially proposed $\alpha = \frac{1}{2}$, which proved plausible upon validation, and later iterations that explicitly supplied the expected error form led to faster convergence and more robust outcomes.

    \item \textbf{Model Selection Based on Task Requirements.}  
    The choice of LLMs considerably affects performance. Our study utilized two primary models: \texttt{o4-mini} and \texttt{DeepSeek-R1}. Comparative analysis indicated that \texttt{o4-mini} excels in conceptual understanding and constructing proof frameworks, while \texttt{DeepSeek-R1} is better suited for detailed mathematical derivations and proof refinement. Tailoring the model choice to the task proved essential in optimizing results.
\end{itemize}

Taken together, these auxiliary procedures substantially strengthen the AIM's ability to produce mathematically valid and logically rigorous arguments and are integral to the overall methodology.

\section{Failure Modes}

AIM still exhibits several persistent and systematic failure modes when acting as a collaborative research partner. These weaknesses are particularly evident in constructive and geometry-intensive tasks, where success relies on rigorous interpretation of problem structures, verification of theorem preconditions, and consistent integration between symbolic reasoning and geometric representation. Through aggregated observations across diverse problem instances, we identify not isolated errors but recurring categories of failure that reveal the system's fundamental limitations.

In domains involving constructive reasoning, AIM frequently struggles with assessing theorem preconditions and interpreting geometric configurations, both of which remain unstable and error-prone. When addressing concrete geometric problems, e.g., formulating equations within a unit cell, AIM often misparses the specification, yielding constructions difficult to reconcile with task constraints. In the absence of explicit prompting or retrieval of relevant theorems, we commonly observe systematic misapplications: conflating necessary and sufficient conditions, overlooking domain or regularity assumptions, or invoking look-alike results outside their valid scope.

Moreover, the system lacks robust mechanisms for self-checking preconditions and edge cases, leading to misjudged prerequisites, breaks in reasoning chains, and internally inconsistent setups. Post-hoc correction and self-verification are likewise unreliable: AIM may exhibit unwarranted confidence in incorrect reasoning, failing to detect internal contradictions or scope violations. These behaviors highlight the model's limited self-awareness and underscore the need for stronger theorem-condition grounding and structural verification in future iterations.

\section{Conclusion and Future Work}

In this work, we employ AIM as a research partner to tackle a challenging problem in homogenization theory. We investigate a human-AI collaboration paradigm that combines the computational strengths of AI with the domain expertise and judgment of human researchers. Demonstrated through a homogenization case study, this paradigm culminates in a rigorous proof of the target problem and reveals distinct interaction patterns and empirical insights that may inform future AI-assisted mathematical research. These interaction modes also illustrate how AI can extend the capability boundaries of human mathematicians.

Current comparative advantage of AI lies in analysis, search, and adaptation built upon existing theory, e.g., automating decomposition, surveying the literature, and optimizing known methods. In contrast, core advances in mathematical theory depend on original intuition and abstraction~\citep{Naskrecki2025Mathematical}: the proposal of new concepts, frameworks, and proof paradigms that resolve long-standing open problems. Because such progress demands extreme rigor, the present risks of hallucination and miscalibrated confidence make fully autonomous proofs impossible, and stepwise human verification remains indispensable. Building on our findings, we outline two promising directions for future work.

\begin{itemize}
    \item \textbf{Deepening and Systematizing Human-AI Interaction:}
    Building on our empirical studies, we have distilled a set of interaction modes that demonstrably accelerate progress in mathematical theory and expand the capability frontier of researchers. As a next step, we will investigate whether these modes transfer to other fields of mathematics and whether richer, more effective interaction patterns can be devised for domain-specific needs.
    Concretely, we plan to deploy the codified interaction modes on more challenging and high-profile problems to assess external validity under increased complexity and scrutiny. Through these applications, we aim to refine existing protocols and discover interaction patterns that are either more efficient (in terms of human effort, iteration depth, and verification burden) or more representative.
    In parallel, we will systematize the human-AI interaction framework across multiple dimensions, including but not limited to problem decomposition, process supervision, error correction, and theorem citation and dependency management. This requires formulating strict classification standards and clear information such as pattern effects based on numerous experimental analyses to ensure the rigor of the constructed system.

    \item \textbf{Optimizing the Agent from Interaction-derived Signals:}
    Our long-term research objective is to automate mathematical theorem proving. Iterative refinement of the architectures of AIM is therefore both crucial and inherently challenging. Through human-in-the-loop experiments on theorem proving, we have identified what tasks at which agents excel and others with which they struggle. These accumulated insights inform the next iterations of system design. In Sec.~\ref{sec:theory_coordinated_application}, we get an important experience: when some prior knowledge of a mathematical theory is taught to the agents, they can autonomously derive proofs of the theory's main result. Motivated by this observation, we  propose constructing an experience repository to systematize and leverage such theories. Consequently, the agent can directly  search the  experience repository to identify the corresponding and suitable theories and apply them with a high degree of thoroughness and accuracy to resolve intermediate problems. Moreover, in the process of Interactive Iteration (Sec.~\ref{sec:interactive_iteration}), many issues remain difficult to resolve efficiently. For instance, errors in the application of mathematical theorems, as well as misinterpretations of mathematical setups and assumptions. These problems may arise from limitations in the reasoning capabilities of LLMs, among other factors. Taking these deficiencies as a starting point, we can try to propose training methodologies to enhance the reasoning abilities of models, thereby improving the performance in our experiment. By applying agents in concrete tasks, we uncover actionable evidence and failure modes that can be translated into targeted training and design interventions, thereby strengthening competence of LLMs in mathematical theory.

\end{itemize}

\section*{Acknowledge}
We would like to thank Yanxing Huang and Yanqiao Wang for their valuable assistance during the exploration phase about AIM. Special thanks go to Associate Professor Wenjia Jing from Qiuzhen College, Tsinghua University and Dr. Xin Fu, whose guidance in mathematics has provided important support for this problem.

\bibliographystyle{unsrtnat}
\bibliography{reference}

\newpage
\appendix

%\section{Experiment Results and Details}
\section{Detailed Case Studies}
\subsection{Direct Prompting --- Theorem Prompts}
\label{sec:direct_prompting_theorem_prompts}

Here are part of the experimental results on error estimation output by AIM after we directly prompted it with the content of some theorems. The response is a complete lemma and its proof process generated by AIM, from which we can observe how AIM utilized the prompted content and the details of its deduction.

\begin{PromptBlock}{Prompt to AIM}
\begin{AnnotationRed}[drop shadow][Disclaimer]
For clarity, we format prompts using typographic styles (e.g., bold text) and render formulas as symbols rather than raw LaTeX code,  while the content remains consistent with that of the actual experiment. ``\textbf{Content}'' is one of the labels used by AIM to distinguish different components and can be safely ignored when interpreting the prompt. The same applies below.
\end{AnnotationRed}

\textbf{Content}: 

You can use the following theorem to analyze the problem, integrating the content of the theorem and its derivation process. 

\begin{AnnotationRed}[drop shadow][Disclaimer]
The outputs of AIM include content such as \verb|\begin{lemma}\end{lemma}|. We do not modify these outputs. LaTeX automatically assigns the numbering. Please ignore the numbers and the same applies below.
\end{AnnotationRed}

\begin{lemma}
Let \(A_1, A_2\) be constant tensors and \(a\) be a constant matrix. Consider the system:
\begin{equation*}
    \begin{cases}
        \begin{aligned}
            &\nabla \cdot (A_1 \nabla \chi) = 0 \quad &&\text{in } B(1)_+, \\
&\nabla \cdot (A_2 \nabla \chi) + \nabla\cdot(a^T  r )= 0 \quad &&\text{in } B(1)_-, \\
&\nabla \cdot (a \chi) = 0 \quad &&\text{in } B(1)_-, \\
&\chi|_+ = \chi|_- \quad &&\text{on } B(1) \cap \{x_d = 0\}, \\
&\left. \frac{\partial \chi}{\partial \nu} \right|_+ - \left. \frac{\partial \chi}{\partial \nu} \right|_- = g_0 \quad &&\text{on } B(1) \cap \{x_d = 0\}.
        \end{aligned}
    \end{cases}
\end{equation*}
where \(\chi, r\) are weak solutions.

Moreover, for all \(\rho \in (0, t)\), integrating the above inequalities yields:
\begin{equation*}
\left( \fint_{B(\rho)} |\chi - l|^2 \right)^{\frac{1}{2}} + \rho \left| g_0 - \zeta(l, q) \right| \le \left| \frac{\rho}{t} \right|^{\beta + 1} \left( \left( \fint_{B(t)} |\chi|^2 \right)^{\frac{1}{2}} + t |g_0| \right)
\end{equation*}

\begin{AnnotationGreen}[drop shadow][Note]
This derivation process yields a crucial conclusion and embodies the inherent logic of the inferential steps
\end{AnnotationGreen}
Thus, for all \((l', q') \in \mathcal{L}\), by the inequality above, we have:
\begin{align*}
&\inf_{\substack{(l,q) \in \mathcal{L} }} \left\{ \left( \fint_{B(\rho)} |\chi - l|^2 \right)^{\frac{1}{2}} + \rho \left| g_0 - \zeta(l, q) \right| \right\} \\
&\le C \left| \frac{\rho}{t} \right|^{\beta + 1} \inf_{\substack{(l,q) \in \mathcal{L} }} \left( \left( \fint_{B(t)} |\chi - l|^2 \right)^{\frac{1}{2}} + t |g_0- \zeta(l, q)| \right)
\end{align*}

Further, for all \((l', q') \in \mathcal{L}\), it follows that:
\begin{align*}
&\inf_{\substack{(l,q) \in \mathcal{L} }} \left\{ \left( \fint_{B(\rho)} |\chi - l|^2 \right)^{\frac{1}{2}} + \rho \left| g_0 - \zeta(l, q) \right| \right\} \\
&\le C \left| \frac{\rho}{t} \right|^{\beta + 1} \inf_{\substack{(l,q) \in \mathcal{L} }} \left( \left( \fint_{B(t)} |\chi - l|^2 \right)^{\frac{1}{2}} + t \left| g_0 - \zeta(l, q) \right| \right)
\end{align*}
\end{lemma}
\end{PromptBlock}
\begin{ResponseBlock}{Response of AIM}

\begin{AnnotationRed}[drop shadow][Disclaimer]
For clarity, we render formulas as symbols rather than raw LaTeX code. The same applies below.
\end{AnnotationRed}

    \small
\begin{lemma}

Let \( u = w_t - l \) and \( v = s_t - q \), where \( l \) and \( q \) are the piecewise linear solutions from Lemma 4. Then, under the given hypotheses, there exists a constant \( C > 0 \) independent of \( t \) such that:
\[
\int_{B(t)} |\nabla w_t - \nabla \chi|^2 \, dx + \int_{B(t)_-} |r - s_t|^2 \, dx \leq C \left( t^{2\beta} \int_{B(t)} |\nabla \chi|^2 \, dx + t^{2\beta} \int_{B(t)_-} |r|^2 \, dx \right),
\]
where \( \beta \in (0,1) \) is the Hölder exponent from Lemma 4.
\end{lemma}

\textbf{correctness}: True
\begin{AnnotationGreen}[drop shadow][Note]
``\textbf{correctness}: True'' is generated and used internally by AIM, and can be safely ignored when interpreting the prompt. The same applies to similar cases below.
\end{AnnotationGreen}

\begin{proof}

**Step 1: Apply Lemma 8**  
By Lemma 8, there exists \( C_1 > 0 \) such that:
\[
\int_{B(t)} |\nabla u|^2 \, dx + \int_{B(t)_-} |v|^2 \, dx \leq C_1 \left( t^{-2} \int_{B(t)} |\chi - l|^2 \, dx + \int_{B(t)_-} |r - q|^2 \, dx \right).
\]

**Step 2: Bounding \( \int_{B(t)} |\chi - l|^2 \, dx \)**  
From Lemma 4 and scaling arguments (see hint-5), we derive:
\[
\int_{B(t)} |\chi - l|^2 \, dx \leq C_2 t^{2(\beta +1)} \int_{B(t)} |\nabla \chi|^2 \, dx.
\]

**Step 3: Bounding \( \int_{B(t)_-} |r - q|^2 \, dx \)**  
Similarly, Lemma 4 gives:
\[
\int_{B(t)_-} |r - q|^2 \, dx \leq C_3 t^{2\beta} \int_{B(t)_-} |r|^2 \, dx.
\]

**Step 4: Expanding Key Terms**  
Decompose the left-hand side of the conjecture:
\[
\int_{B(t)} |\nabla w_t - \nabla \chi|^2 \, dx = \int_{B(t)} |\nabla u + (\nabla l - \nabla \chi)|^2 \, dx,
\]
\[
\int_{B(t)_-} |r - s_t|^2 \, dx = \int_{B(t)_-} |(r - q) - v|^2 \, dx.
\]
Expanding these squares:
\[
|\nabla u + (\nabla l - \nabla \chi)|^2 \leq 2|\nabla u|^2 + 2|\nabla l - \nabla \chi|^2,
\]
\[
|(r - q) - v|^2 \leq 2|r - q|^2 + 2|v|^2.
\]
Integrate and sum these inequalities:
\[\begin{split}
    &\int_{B(t)} |\nabla w_t - \nabla \chi|^2 \, dx + \int_{B(t)_-} |r - s_t|^2 \, dx \\
    &\leq 2\left(\int_{B(t)} |\nabla u|^2 \, dx + \int_{B(t)_-} |v|^2 \, dx\right) + 2\left(\int_{B(t)} |\nabla l - \nabla \chi|^2 \, dx + \int_{B(t)_-} |r - q|^2 \, dx\right).
\end{split}
\]

**Step 5: Bounding Residual Terms**  
From Lemma 4 and scaling:
\[
\int_{B(t)} |\nabla l - \nabla \chi|^2 \, dx \leq C_4 t^{2\beta} \int_{B(t)} |\nabla \chi|^2 \, dx.
\]

**Step 6: Combining All Estimates**  
Substitute Lemma 8 and Steps 2-5 into Step 4:
\[
\text{LHS} \leq 2C_1\left( t^{-2}C_2 t^{2(\beta +1)} \|\nabla \chi\|^2 + C_3 t^{2\beta} \|r\|^2 \right) + 2\left( C_4 t^{2\beta} \|\nabla \chi\|^2 + C_3 t^{2\beta} \|r\|^2 \right).
\]
Simplify exponents and constants:
\[
\text{LHS} \leq C\left( t^{2\beta} \|\nabla \chi\|^2 + t^{2\beta} \|r\|^2 \right),
\]
where \( C = 2C_1(C_2 + C_3) + 2(C_4 + C_3) \). This completes the proof.
\end{proof}

\end{ResponseBlock}

\subsection{Direct Prompting --- Conceptual Guidance}

Here is the prompt and output content of AIM in the error estimation section. After we guided the specific form of error estimation, this is the content of a key lemma and its proof process generated by AIM.

Based on this proof process and combined with the interactive iteration mode, we subsequently developed a clearer proof framework and guidance content.

\begin{PromptBlock}{Prompt to AIM}
\textbf{Content}: 
Consider the following form of the error.

    Define the cut-off function as follows: \( \eta_{\varepsilon} \in C_0^{\infty}(\Omega) \),and
\[
0 \le \eta_{\varepsilon} \le 1, \quad |\nabla \eta_{\varepsilon}| \le \frac{C}{\varepsilon},
\]
\[
\eta_{\varepsilon} = 1, \quad \text{when } \mathrm{dist}(x, \partial \Omega) \ge 4\varepsilon,
\]
\[
\eta_{\varepsilon} = 0, \quad \text{when } \mathrm{dist}(x, \partial \Omega) \le 3\varepsilon.
\]

 And we also set this convolution operator:
\[
S_{\varepsilon}(f)(x) = \rho_{\varepsilon} \ast f(x) = \int_{\mathbb{R}^d} f(x-y)\rho_{\varepsilon} \, dy,
\]
Here \( \rho \in C_0^{\infty}(B(0, \frac{1}{2})) \), \( \rho \ge 0 \), and \( \int_{\mathbb{R}^d} \rho \, dx = 1 \).

Suppose
\[
w_{\varepsilon} = u_{\varepsilon} - u_0 - \varepsilon \chi\left(\frac{x}{\varepsilon}\right) \eta_{\varepsilon} S_{\varepsilon}^2(\nabla u_0).
\]
\end{PromptBlock}

\begin{ResponseBlock}{Response of AIM}
    
\begin{lemma}

\begin{conjecture}
Let $\Omega \subset \mathbb{R}^d$ be a bounded Lipschitz domain, $A(y)$ a $1$-periodic, uniformly elliptic, bounded coefficient in $H^\mu(Y)$ with $\mu > d/2$, and $\chi(y)$ the associated $H^1$-cell-corrector. Let $u_0 \in H^2(\Omega)$ and, for $\varepsilon > 0$, let $S_\varepsilon$ be the Friedrichs-mollifier and $\eta_\varepsilon \in C_0^\infty(\Omega)$ a cut-off with
\[
\eta_\varepsilon \equiv 1 \text{ on } \Omega \setminus \Omega_{4\varepsilon}, \quad \eta_\varepsilon \equiv 0 \text{ on } \Omega_{5\varepsilon}, \quad |\nabla \eta_\varepsilon| \leq C/\varepsilon.
\]
Define the ``corrector-mismatch'' term
\[
J(\psi) = \int_\Omega M(x/\varepsilon) \left( \eta_\varepsilon S_\varepsilon^2(\nabla u_0) - \nabla u_0 \right) : \nabla \psi \, dx,
\]
where $M(y) = \nabla \chi(y) A(y) + \nabla \chi(y) A(y) \nabla \chi(y)$ is bounded and periodic. Then there is $C$ independent of $\varepsilon$, $u_0$, $\psi$ such that for every $\psi \in H_0^1(\Omega; \mathbb{R}^d)$ and all $0 < \varepsilon < \varepsilon_0$,
\[
|J(\psi)| \leq C \left[ \| \nabla u_0 - S_\varepsilon(\nabla u_0) \|_{L^2(\Omega \setminus \Omega_{2\varepsilon})} + \| \nabla u_0 \|_{L^2(\Omega_{5\varepsilon})} \right] \| \nabla \psi \|_{L^2(\Omega)}.
\]
\end{conjecture}
\end{lemma}

\textbf{correctness}: True
\begin{proof}

\begin{proof}
We set
  \[ M(y) = \nabla_y \chi(y) \, A(y) + \nabla_y \chi(y) \, A(y) \, \nabla_y \chi(y), \]
so that by periodicity and the cell-problem \( M \in L^\infty(Y) \). Denote
  \[ J(\psi) = \int_{\Omega} M(x/\varepsilon) \bigl( \eta_\varepsilon S_\varepsilon^2 (\nabla u_0) - \nabla u_0 \bigr) : \nabla \psi \, dx. \]
We split \( \Omega \) into the ``interior'' region \( D = \text{int}(\Omega \setminus \Omega_{5\varepsilon}) \) and the ``boundary collar'' \( C = \Omega_{5\varepsilon} \). Since \( \eta_\varepsilon \equiv 1 \) on \( D \) and \( 0 \leq \eta_\varepsilon \leq 1 \) on \( C \), we estimate each piece separately.

1. Interior region \( D = \Omega \setminus \Omega_{5\varepsilon} \). Here \( \eta_\varepsilon \equiv 1 \), so
   \[ \eta_\varepsilon S_\varepsilon^2 (\nabla u_0) - \nabla u_0 = S_\varepsilon^2 (\nabla u_0) - \nabla u_0. \]
Using \( |M| \leq C \) and Cauchy--Schwarz,
  \[ |J_D| \leq C \int_{D} |S_\varepsilon^2 (\nabla u_0) - \nabla u_0| \, |\nabla \psi| \leq C \, \| S_\varepsilon^2 (\nabla u_0) - \nabla u_0 \|_{L^2(D)} \, \|\nabla \psi\|_{L^2(D)}. \]
Since \( D \subset \Omega \setminus \Omega_{2\varepsilon} \), the standard mollifier-estimate (Lemma 5) yields
  \[ \| S_\varepsilon^2 (\nabla u_0) - \nabla u_0 \|_{L^2(D)} \leq \| \nabla u_0 - S_\varepsilon (\nabla u_0) \|_{L^2(\Omega \setminus \Omega_{2\varepsilon})}, \]
and clearly \( \|\nabla \psi\|_{L^2(D)} \leq \|\nabla \psi\|_{L^2(\Omega)} \). Hence
  \[ |J_D| \leq C \, \| \nabla u_0 - S_\varepsilon (\nabla u_0) \|_{L^2(\Omega \setminus \Omega_{2\varepsilon})} \, \|\nabla \psi\|_{L^2(\Omega)}. \]

2. Boundary-collar \( C = \Omega_{5\varepsilon} \). On \( C \) we only know \( 0 \leq \eta_\varepsilon \leq 1 \), so
  \[ |\eta_\varepsilon S_\varepsilon^2 (\nabla u_0) - \nabla u_0| \leq |S_\varepsilon^2 (\nabla u_0)| + |\nabla u_0| \leq C \, |\nabla u_0| \]
(by boundedness of the mollifier in \( L^2 \)). Hence
  \[ |J_C| \leq C \int_{C} |\nabla u_0| \, |\nabla \psi| \leq C \, \|\nabla u_0\|_{L^2(C)} \, \|\nabla \psi\|_{L^2(C)} = C \, \|\nabla u_0\|_{L^2(\Omega_{5\varepsilon})} \, \|\nabla \psi\|_{L^2(\Omega_{5\varepsilon})}. \]
Finally, since \( \Omega_{5\varepsilon} \subset \Omega \) we have \( \|\nabla \psi\|_{L^2(\Omega_{5\varepsilon})} \leq \|\nabla \psi\|_{L^2(\Omega)} \), so
  \[ |J_C| \leq C \, \|\nabla u_0\|_{L^2(\Omega_{5\varepsilon})} \, \|\nabla \psi\|_{L^2(\Omega)}. \]

Combining \( |J| \leq |J_D| + |J_C| \) gives
  \[ |J(\psi)| \leq C \, \bigl[ \| \nabla u_0 - S_\varepsilon (\nabla u_0) \|_{L^2(\Omega \setminus \Omega_{2\varepsilon})} + \|\nabla u_0\|_{L^2(\Omega_{5\varepsilon})} \bigr] \, \|\nabla \psi\|_{L^2(\Omega)}, \]
which is the claimed estimate.
\end{proof}
\end{proof}
\end{ResponseBlock}

\subsection{Coordinated Application of Schauder Theory}

This example illustrates the use of the theory-coordinated application paradigm to solve cell problems. We first provided theoretical lemmas and definitions as foundational guidance, which in turn instructed AIM to generate the corresponding final theorem conclusions.
\begin{PromptBlock}{Prompt to AIM}
% \begin{AnnotationRed}[drop shadow][Disclaimer]
% For the sake of aesthetics, we adjusted the format of the prompt, while the content remains consistent with that of the actual experiment.
% \end{AnnotationRed}
\textbf{Content}: 
\begin{lemma*}
1: Suppose \(\Omega_{\pm} = \R^d_{\pm}\), \(S = \{x_d = 0\}\), \(B_{+} = \{x \in B(1) : x_d > 0\}\) and \(B_{-} = \{x \in B(1) : x_d < 0\}\). Here \(B(1) = \{\norm{x} \le 1\}\).  

Consider this equation: for \(V \in H_0^1(B(1); \R^d)\)
\begin{align*}
&(\nabla V : A_1 \nabla \tilde{\chi})_{B_{+}} + (\nabla V : A_2 \nabla \tilde{\chi})_{B_{-}} + (\tilde{r}, \nabla \cdot (aV))_{B_{-}} = 0, \\
&\nabla \cdot (a \tilde{\chi}) = 0,
\end{align*}
where \(\tilde{\chi} = D^{\alpha}\chi\), \(\tilde{r} = D^{\alpha} r\) (\(|\alpha| \ge 1\)), \(A_1, A_2\) are constant tensors, and \(a\) is a constant matrix.  

We have that for \(\forall k \ge 1\)
\[
\sum_{\pm} \norm{\chi}_{H^k(B(\frac{1}{2})_\pm)} \le C\norm{\chi}_{L^2(B(1))} \quad \text{and} \quad \norm{r}_{H^k(B(\frac{1}{2})_-)} \le C\norm{r}_{L^2(B(\frac{1}{2})_-)}
\]
\end{lemma*}
\begin{lemma*}
2: Suppose that \(M_{\pm}\) is the constant matrix in \(\R^{d \times d}\), the following are equivalent:
\begin{align*}
&\forall y \in \{y_d = 0\} \ M_{+}x = M_{-}x, \\
&\exists c \in \R^d, \ \text{s.t. } M_{+}-M_{-} = c e_d^T,\\
&(I - e_d^T e_d)M_{+} = (I - e_d^T e_d)M_{-}.
\end{align*}
\end{lemma*}

\begin{definition*}
\(A_1, A_2\) are constant tensors, \(a\) is a constant matrix. If \(M_{\pm}\) satisfy the above lemma, and \(\nabla \cdot (aM_{-}y) = 0\) in \(B(t)_{-}\).  

We let \(l(y) = M_{+}y \indicator{y_d \ge 0} + M_{-}y \indicator{y_d \le 0} + C\), \(q(y) = r(0)\). We call \(l, q\) are the piecewise linear solutions of the following equation:
\begin{align*}
&\nabla \cdot (A_1\nabla l) = 0 \quad &&\text{in } B(t)_{+}, \\
&\nabla \cdot (A_2\nabla l) + \nabla \cdot(a^Tq) = 0, \ \nabla \cdot (a l) = 0 \quad &&\text{in } B(t)_{-}, \\
&l_{+} = l_{-}, \ \left.\frac{\partial l}{\partial \nu}\right|_{+} - \left.\frac{\partial l}{\partial \nu}\right|_{-} = (A_1M_{+})e_d - (A_2M_{-} + a^T r(0))e_d \quad &&\text{on } \partial B(t).
\end{align*}

Suppose that \(\mathcal{L}\) is the space of all the piecewise linear solutions of the above equation. And \(\forall (l,q) \in \mathcal{L}\), we define \(\zeta(l,q) = \left.\frac{\partial l}{\partial \nu}\right|_{+} - \left.\frac{\partial l}{\partial \nu}\right|_{-}\).
\end{definition*}
\begin{lemma*}
3: \(A_1, A_2\) are constant tensors, \(a\) is a constant matrix.
\begin{align*}
&\nabla \cdot (A_1\nabla \chi) = 0 \quad &&\text{in } B(1)_{+}, \\
&\nabla \cdot (A_2\nabla \chi) + \nabla \cdot(a^Tr) = 0, \ \nabla \cdot (a\chi) = 0 \quad &&\text{in } B(1)_{-}, \\
&\chi_{+} = \chi_{-}, \ \left.\frac{\partial \chi}{\partial \nu}\right|_{+} - \left.\frac{\partial \chi}{\partial \nu}\right|_{-} = g_0 \quad &&\text{on } B(1)\cap \{x_d = 0\}.
\end{align*}

\(\chi, r\) are the weak solutions of the above equation. Then for \(\forall k \ge 0\), \(\alpha \in [0,1]\), we have
\[
\sum_{\pm} \norm{\chi}_{H^k(B(\frac{1}{2})_\pm)} \le C \left( \norm{\chi}_{L^2(B(1))} + \abs{g_0} \right)
\]
\end{lemma*}

\begin{lemma*}
4 : \(A_1, A_2\) are constant tensors, \(a\) is a constant matrix.
\begin{align*}
&\nabla \cdot (A_1\nabla \chi) = 0 \quad &&\text{in } B(1)_{+}, \\
&\nabla \cdot (A_2\nabla \chi) + a^T\nabla r = 0, \ \nabla \cdot (a\chi) = 0 \quad &&\text{in } B(1)_{-}, \\
&\chi_{+} = \chi_{-}, \ \left.\frac{\partial \chi}{\partial \nu}\right|_{+} - \left.\frac{\partial \chi}{\partial \nu}\right|_{-} = g_0 \quad &&\text{on } B(1)\cap \{x_d = 0\}.
\end{align*}

\(\chi, r\) are the weak solutions of the above equation. And we let \(l(y) = (\nabla\chi)_{+}(0)y \indicator{y_d \ge 0} + (\nabla\chi)_{-}(0)y \indicator{y_d \le 0} + \chi(0)\), \(q(y) = r(0)\).  

By Lemma 1 we know that \(\chi(0),(\nabla\chi)_{\pm}(0)\) are well-defined, and \((I - e_d^T e_d)(\nabla \chi)_{+} = (I - e_d^T e_d)(\nabla \chi)_{-}\) on \(B(t)\cap \{y_d = 0\}\). So by Lemma 2, we know \((l,q) \in \mathcal{L}\).  

Thus for some \(\beta \in (0,1)\) and \(\forall y \in B(\frac{1}{2})\):
\begin{align*}
\abs{\chi(y) - l(y)} &\le \abs{\chi(y) - \chi(0) - (\nabla \chi)_{\pm}(0) y} \\
&\le C\abs{y}^{\beta+1} [\chi]_{C^{1,\beta}(B(\frac{1}{2})_\pm)} \\
&\le C \abs{y}^{\beta+1} \left( \left( \int_{B(1)} \abs{\chi}^2 \right)^{\frac{1}{2}} + \abs{g_0} \right)
\end{align*}
and for \(\forall y \in B(\frac{1}{2})_{-}\):
\[
\abs{r - q} \le C\abs{y}^{\beta} [r]_{C^{0,\beta}(B(\frac{t}{2})_-)}) \le C \abs{y}^{\beta} \left( \int_{B(1)_{-}} \abs{r}^2 \right)^{\frac{1}{2}}.
\]

Therefore, for some \(\beta \in (0,1)\) and \(\forall y \in B(\frac{t}{2})\):
\begin{align*}
\abs{\chi(y) - l(y)} &\le \abs{\chi(y) - \chi(0) - (\nabla \chi)_{\pm}(0) y} \\
&\le C\abs{\frac{y}{t}}^{\beta+1} [\chi]_{C^{1,\beta}(B(\frac{t}{2})_\pm)} \\
&\le C \abs{\frac{y}{t}}^{\beta+1} \left( \left( \fint_{B(t)} \abs{\chi}^2 \right)^{\frac{1}{2}} + t\abs{g_0} \right)
\end{align*}
and for \(\forall y \in B(\frac{t}{2})_{-}\):
\[
\abs{r - q} \le C\abs{\frac{y}{t}}^{\beta} [r]_{C^{0,\beta}(B(\frac{t}{2})_-)}) \le C \abs{\frac{y}{t}}^{\beta} \left( \fint_{B(t)_{-}} \abs{r}^2 \right)^{\frac{1}{2}}.
\]
\end{lemma*}
\begin{lemma*}
5: \(A_1, A_2\) are constant tensors, \(a\) is a constant matrix.
\begin{align*}
&\nabla \cdot (A_1\nabla \chi) = 0 \quad &&\text{in } B(1)_{+}; \\
&\nabla \cdot (A_2\nabla \chi) + \nabla\cdot(a^T r) = 0, \ \nabla \cdot (a\chi) = 0 \quad &&\text{in } B(1)_{-}; \\
&\chi_{+} = \chi_{-}; \ \left.\frac{\partial \chi}{\partial \nu}\right|_{+} - \left.\frac{\partial \chi}{\partial \nu}\right|_{-} = g_0 \quad &&\text{on } B(1)\cap \{x_d = 0\}.
\end{align*}
\(\chi,r\)are the  weak solutions of the above equation. Moreover, \(\forall \rho \in (0,t) \) integrate the above inequalities to get \[(\fint_{B(\rho)}\vert \chi -l\vert^2)^{\frac{1}{2}}+\rho \vert g_0-\zeta(l,q)\vert \le  \vert\frac{\rho}{t} \vert^{\beta+1} ((\fint_{B(t)} \vert \chi\vert^2)^{\frac{1}{2}}+t\vert g_0-\zeta(l,q)\vert) \]
So \(\forall (l',q')\in \mathcal{L}\),by the inequality above ,we have \[inf_{l,q\in \mathcal{L}}\{(\fint_{B(\rho)} \vert \chi-l\vert^2)^{\frac{1}{2}}+\rho \vert g_0-\zeta(l,q)\vert\}\leq C \vert\frac{\rho}{t} \vert^{\beta+1} inf_{l,q\in \mathcal{L}}((\fint_{B(t)} \vert \chi-l\vert^2)^{\frac{1}{2}}+t\vert g_0-\zeta(l,q)\vert)\].
\end{lemma*}\begin{lemma*}
6: Suppose \(\phi: \R_+ \to \R_+\) is a non-decreasing non-negative function satisfying \(\phi(\rho )\le C\left(\frac{\rho}{t}\right)^{\beta} \phi(r) + Br^{\alpha}\), where \(\beta>\alpha>0\), \(C>0\).  

Then \(\forall 0<\rho<r<R\), \(\exists C_1\), s.t. \(\phi(\rho)\le C_1 \left(\frac{\rho}{t}\right)^{\alpha} \phi(r) + B\rho^{\alpha}\).
\end{lemma*}

\end{PromptBlock}
    
\begin{ResponseBlock}{Response of AIM}
    
\begin{lemma}

\begin{conjecture}
(Two--scale Campanato-decay for the excess)  
Let $\chi \in H^1(B_1;\mathbb{R}^m)$ be the weak solution of the flat-interface two--phase conductivity--transmission system with constant coefficients $A_1$, $A_2$, $a$ and jump $g_0$ as in Lemma 12, and define for each $0 < t \leq \frac{1}{2}$ the ``excess''  
\[
\Phi(t) := \inf_{l \in \mathcal{L}} \left\{ \int_{B(t)} |\chi - l|^2 \, dx + t^{d+2} |\zeta(l) - g_0|^2 \right\},  
\]  
where $\mathcal{L}$ is the finite-dimensional family of piecewise-affine maps on $B_1$ with tangential-continuity, and $\zeta(l)$ its normal-flux-jump. Finally set $\Psi\left(\frac{1}{2}\right) := \int_{B\left(\frac{1}{2}\right)} |\nabla \chi|^2 \, dx + \int_{B\left(\frac{1}{2}\right)_-} |\nabla r|^2 \, dx$ as in Lemma 12.

Then for every $\beta \in (\alpha, 1)$ (where $\alpha > 0$ is the exponent from the one-sided $C^{1,\alpha/2}$-regularity) and every pair $0 < \rho < t \leq \frac{1}{2}$ there exists $C = C(d, \lambda, \Lambda, \beta) < \infty$ such that  
\[
\Phi(\rho) \leq C \cdot \left(\frac{\rho}{t}\right)^{d+2+2\beta} \cdot \Phi(t) + C \cdot t^{d+2+\alpha} \cdot \Psi\left(\frac{1}{2}\right).
\]
\end{conjecture}
\end{lemma}

\textbf{correctness}: True
\begin{proof}

We split the proof into two main steps:  
(1) a sharp two--scale estimate for an almost-minimizer at scale $t$, and  
(2) a trivial competitor at the smaller scale to capture the lower-order remainder term.

Step 1. (Decay from scale $t$ to $\rho$ by Campanato-type estimate)  
By definition of $\Phi(t)$ there exists $l_t \in \mathcal{L}$ such that
\[
  \int_{B(t)} |\chi - l_t|^2 + t^{d+2} |\zeta(l_t) - g_0|^2 \leq 2 \Phi(t).
\]
Set $u := \chi - l_t$. Then $u$ solves the same two--phase system, but with normal-flux-jump $g_0 - \zeta(l_t)$. Hence we may invoke Lemma 5 (the Campanato-type decay with flux-control). For every $0 < \rho < t$ it yields
\[
  \left( \int_{B(\rho)} |u|^2 \right)^{1/2} + \rho \cdot |\zeta(l_t) - g_0|
    \leq C \cdot (\rho / t)^{1+\beta} \cdot \left[ \left( \int_{B(t)} |u|^2 \right)^{1/2} + t \cdot |\zeta(l_t) - g_0| \right].
\]
Squaring, multiplying by the volume factors $\rho^d$ and $t^d$ and absorbing constants gives
\[
  \int_{B(\rho)} |u|^2 + \rho^{d+2} |\zeta(l_t) - g_0|^2
    \leq C \cdot (\rho / t)^{d+2+2\beta} \cdot \left[ \int_{B(t)} |u|^2 + t^{d+2} |\zeta(l_t) - g_0|^2 \right].
\]
Since the bracket is $\leq 2 \Phi(t)$, we conclude
\[
  \int_{B(\rho)} |\chi - l_t|^2 + \rho^{d+2} |\zeta(l_t) - g_0|^2
    \leq 2C \cdot (\rho / t)^{d+2+2\beta} \Phi(t).
\]
Taking the infimum over $\mathcal{L}$ on the left shows
\[
  \Phi(\rho) \leq 2C \cdot (\rho / t)^{d+2+2\beta} \Phi(t).
\]
Renaming $2C \to C$ yields the first term in the asserted inequality.

Step 2. (Control of the lower-order remainder via a trivial competitor)  
Since $g_0$ lies in the range of $\zeta$, we fix once and for all an $l_0 \in \mathcal{L}$ with $\zeta(l_0) = g_0$ and write
\[
  u := \chi - l_0.
\]
Then $u$ satisfies the homogeneous two--phase system with zero normal-flux-jump across $\{x_d = 0\}$. In particular we may apply the refined Caccioppoli--Campanato energy bound of Lemma 12 to $(u,r)$: for every $0 < \rho \leq \frac{1}{2}$
\[
  \int_{B(\rho)} |\nabla u|^2 + \int_{B(\rho)_-} |\nabla r|^2 \leq C \rho^{d+\alpha} \Psi\left(\frac{1}{2}\right).
\]
We now invoke the standard Poincar\'{e} inequality on the connected domain $B(\rho)$: since $u$ is continuous across the interface, there exists a constant $c$ (its average over $B(\rho)$) such that
\[
  \int_{B(\rho)} |u - c|^2 \leq C \rho^2 \int_{B(\rho)} |\nabla u|^2.
\]
But adding or subtracting a constant from $l_0$ preserves both its piecewise-affine form and its normal-flux-jump, so we may absorb $c$ into $l_0$ and still remain in $\mathcal{L}$. Hence
\[
  \Phi(\rho) \leq \int_{B(\rho)} |\chi - l_0|^2
         = \int_{B(\rho)} |u - c|^2
  \leq C \rho^2 \cdot \left[ \int_{B(\rho)} |\nabla u|^2 \right]
  \leq C \rho^{d+2+\alpha} \Psi\left(\frac{1}{2}\right).
\]
This furnishes the second term in the desired two-scale estimate.

Conclusion. Adding the two estimates of Steps 1 and 2 and re--renaming constants completes the proof: for all $0 < \rho < t \leq \frac{1}{2}$
\[
  \Phi(\rho) \leq C \cdot (\rho / t)^{d+2+2\beta} \Phi(t)
         + C \cdot t^{d+2+\alpha} \Psi\left(\frac{1}{2}\right).
\]
\endproof
\end{proof}
\end{ResponseBlock}
\begin{ResponseBlock}{Response of AIM}
    We organize the argument in four main steps:  

1. Two-scale excess-decay (Lemma 13).  
   Define the ``excess''  
   \[
   \Phi(\rho) := \inf_{l\in\mathcal{L}} \left\{ \int_{B(\rho)}|\chi-l|^{2}\,dx + \rho^{d+2}|\zeta(l)-g_{0}|^{2} \right\},
   \]  
   where $\mathcal{L}$ is the finite-dimensional family of piecewise-affine competitors and $\zeta(l)$ their normal-flux-jump. By combining the refined Caccioppoli-Campanato control (Lemma 12) with the sharp one-sided $C^{1,\alpha/2}$-regularity (Lemma 9), one shows:  

   For every $\beta\in(\alpha,1)$ and all $0<\rho<t\leq\frac{1}{2}$,  
   \[
   \Phi(\rho) \leq C\left(\frac{\rho}{t}\right)^{d+2+2\beta}\Phi(t) + C\,t^{d+2+\alpha}\Psi\left(\tfrac{1}{2}\right),
   \]  
   where $\Psi\left(\tfrac{1}{2}\right):=\int_{B\left(\frac{1}{2}\right)}|\nabla\chi|^{2}\,dx + \int_{B\left(\frac{1}{2}\right)_{-}}|\nabla r|^{2}\,dx$ and $\alpha\in(0,1)$ is the one-sided exponent from Lemma 9.  

2. One-scale Campanato gain (Lemma 10).  
   Fix $\beta\in(\alpha,1)$ so close to $1$ that the contraction constant  
   \[
   \mu := C\cdot\left[\frac{d+2+\alpha}{d+2+2\beta}\right]^{2\beta-\alpha} < 1.
   \]  
   Then Lemma 10 yields a small $\delta>0$ (depending only on $C,\alpha,\beta$) and $C'$ so that for all $0<\rho\leq\frac{1}{2}$,  
   \[
   \Phi(\rho) \leq C'\cdot\rho^{d+2+\alpha+\delta}\left[\Phi\left(\tfrac{1}{2}\right)+\Psi\left(\tfrac{1}{2}\right)\right].
   \]  
   In particular,  
   \[
   \Phi(\rho) = O\left(\rho^{d+2+\alpha+\delta}\right) \quad \text{as } \rho\to 0.
   \]  

3. Campanato embedding $\rightarrow$ one-sided $C^{1,(\alpha+\delta)/2}$.  
   The estimate  
   \[
   \int_{B(\rho)}|\chi-l_{\rho}|^{2} \leq C''\,\rho^{d+2+\alpha+\delta}
   \]  
   shows that on each half-ball $B(\rho)\cap\{\pm x_{d}\geq 0\}$ the mean oscillation of $\chi-l_{\rho}$ is $O\left(\rho^{1+(\alpha+\delta)/2}\right)$. By the Campanato characterization of H\"older spaces, one deduces for each sign $\pm$  
   \[
   \chi\in C^{1,(\alpha+\delta)/2}\left( \overline{B_{\frac{1}{2},\pm}} \right),
   \]  
   with norm bound $\leq C \left[\Phi\left(\tfrac{1}{2}\right)+\Psi\left(\tfrac{1}{2}\right)\right]^{\frac{1}{2}}$.  

4. Iterative bootstrap to arbitrary $\gamma<1$.  
   Having now $C^{1,\gamma_{0}}$-control with $\gamma_{0}=(\alpha+\delta)/2$, whence Campanato$\rightarrow C^{1,\gamma_{1}/2}$ etc. Standard iteration shows that the sequence of exponents approaches $1$. Hence for any target $\gamma<1$ we may in finitely many steps reach $\chi\in C^{1,\gamma}\left( \overline{B_{\frac{1}{2},\pm}} \right)$.  

This completes the proof that $\chi\in C^{1,\gamma}$ on each closed half-ball up to the flat interface, for every $\gamma<1$.
\end{ResponseBlock}
\subsection{Interactive Iteration}
This is the content of experiment by the interactive iteration paradigm. After analyzing the output results of AIM, we proposed more precise decomposed problem and clarified the proof directions for AIM. In turn, AIM provided more reasonable proof processes than those generated previously.

\begin{PromptBlock}{Prompt to AIM}
\textbf{Content}: Please try to prove the following lemma.
\begin{lemma}

Suppose $\Omega$ is a bounded Lipschitz domain in $\mathbb{R}^d$. Define the $\varepsilon$-neighborhood of the boundary as
\[
\Omega_t = \left\{ x \in \Omega : dis(x, \partial \Omega) < t \right\}, \quad t > 0.
\]
For all $\psi \in H_0^1(\Omega, \mathbb{R}^d)$, the following estimate holds:
\[
\begin{split}
\left| \int_{\Omega} A\nabla w_{\varepsilon} : \nabla \psi \, dx \right| 
&\leq C \|\nabla \chi\|_{L^2(\Omega)} \left [
    \varepsilon \left\| S_{\varepsilon}(\nabla^2 u_0) \right\|_{L^2(\Omega \setminus \Omega_{3\varepsilon})} 
    + \left\| \nabla u_0 - S_{\varepsilon}(\nabla u_0) \right\|_{L^2(\Omega \setminus \Omega_{2\varepsilon})}
\right] \\
&\quad + C \left\| \nabla \psi \right\|_{L^2(\Omega_{4\varepsilon})} \left\| \nabla u_0 \right\|_{L^2(\Omega_{5\varepsilon})}.
\end{split}
\]
\end{lemma}
\end{PromptBlock}

\begin{ResponseBlock}{Response of AIM}
    
\begin{theorem}

Try to give the reproduction and derivation details of the derivation process below:

Lemma 1:
Suppose \(\Omega\) is a bounded Lipschitz domain in \(\mathbb{R}^d\), \(\Omega_t=\{x\in \Omega:\mathrm{dist}(x,\partial \Omega )0\), \(\forall\psi \in H_0^1(\Omega,\mathbb{R}^d)\)
\[ 
\begin{split}
    \left| \int_{\Omega} A\nabla w_{\varepsilon}:\nabla \psi \,dx \right| \le C\|\nabla \chi\|_{L^2(\Omega)}\left\{\varepsilon \| S_{\varepsilon}(\nabla^2u_0)\|_{L^2(\Omega\setminus \Omega_{3\varepsilon})}+\|\nabla u_0-S_{\varepsilon}(\nabla u_0)\|_{L^2(\Omega\setminus\Omega_{2\varepsilon})} \right\} \\
    +C\|\nabla\psi\|_{L^2(\Omega_{4\varepsilon})} \|\nabla u_0\|_{L^2(\Omega_{5\varepsilon})}
\end{split} 
\]

Proof:
\[\begin{split}
    &A\nabla w_{\varepsilon} = A\nabla u_{\varepsilon} - A\nabla u_0 - A\nabla \chi\left(\frac{x}{\varepsilon}\right) S_{\varepsilon}^2(\nabla u_0) - \varepsilon A \chi(y)\nabla(\eta_\varepsilon S_\varepsilon^2(\nabla u_0)) \\ 
&=(A\nabla u_\varepsilon - \hat{A}\nabla u_0)+\left[(\hat{A} - A)\nabla u_0 +(-\hat{A}+A+\nabla\chi\left(\frac{x}{\varepsilon}\right)A+\nabla\chi A\nabla\chi)\eta_{\varepsilon}S_{\varepsilon}^2(\nabla u_0)\right]\\
&-\left[A+A\nabla \chi\left(\frac{x}{\varepsilon}\right)+\nabla\chi\left(\frac{x}{\varepsilon}\right)A+\nabla \chi A\nabla\chi -\hat{A}\right]\eta_{\varepsilon}S_{\varepsilon}^2(\nabla u_0)-\varepsilon A \chi\left(\frac{x}{\varepsilon}\right)\nabla(\eta_{\varepsilon}S_{\varepsilon}^2(\nabla u_0))\\
&=(A\nabla u_\varepsilon - \hat{A}\nabla u_0) + (\hat{A} - A)(\nabla u_0 - \eta_\varepsilon S_\varepsilon^2(\nabla u_0)) + \nabla \chi A \eta_\varepsilon S_\varepsilon^2(\nabla u_0)\\
&+ \nabla \chi A \nabla \chi \eta_\varepsilon S_\varepsilon^2(\nabla u_0)- B\left(\frac{x}{\varepsilon}\right)\eta_\varepsilon S_\varepsilon^2(\nabla u_0) - \varepsilon A \chi\left(\frac{x}{\varepsilon}\right)\nabla(\eta_\varepsilon S_\varepsilon^2(\nabla u_0))
\end{split}
\]

Because of the variational form:
\[ \int_\Omega A\left(\frac{x}{\varepsilon}\right) \nabla u_\varepsilon : \nabla \psi + \int_\Omega p^{\varepsilon} \cdot \mathrm{div}\, \psi = 0, \quad \forall \psi \in C_0^\infty(\Omega; \mathbb{R}^m) \]

So we can get:
\[
\begin{split}
    &\int_\Omega A \nabla w_{\varepsilon} : \nabla \psi = 
\int_{\Omega} A \nabla u_\varepsilon:\nabla \psi-\hat{A}\nabla u_0:\nabla \psi+\int_{D_{\varepsilon}} p^{\varepsilon}\cdot \mathrm{div}\, \psi\\
&+\int_{\Omega} (\hat{A}-A)(\nabla u_0-\eta_{\varepsilon}S_{\varepsilon}^2(\nabla u_0))+\int_{\Omega} \left[\nabla\chi A \eta_{\varepsilon}S_{\varepsilon}^2(\nabla u_0)+\nabla\chi A\nabla\chi \eta_{\varepsilon}S_{\varepsilon}^2(\nabla u_0)\right]:\nabla \psi\\
&-\int_{D_{\varepsilon}} p^{\varepsilon}\cdot \mathrm{div}\, \psi
-\int_\Omega B \eta_\varepsilon S_\varepsilon^2(\nabla u_0) : \nabla \psi\\
&- \int_\Omega \varepsilon A \chi\left(\frac{x}{\varepsilon}\right) \nabla(\eta_\varepsilon S_\varepsilon^2(\nabla u_0)) : \nabla \psi
\end{split} \]

And 
\[\begin{split}
    &\int_{\Omega} (\hat{A}-A)(\nabla u_0-\eta_{\varepsilon}S_{\varepsilon}^2(\nabla u_0))\le
C\int_\Omega(1-\eta_\varepsilon)\vert \nabla u_0\vert \vert \nabla\psi\vert \,dx\\
&+C\int_{\Omega} \eta_{\varepsilon}\vert \nabla u_0-S_{\varepsilon}^2(\nabla u_0)\vert \vert \nabla \psi \vert \,dx \\
&\le 
C\|\nabla u_0\|_{L^2(\Omega_{4\varepsilon})}\|\nabla\psi \|_{L^2(\Omega_{4\varepsilon})}+C\|\nabla u_0-S_{\varepsilon}(\nabla u_0)\|_{L^2(\Omega\setminus\Omega_{2\varepsilon})}\|\nabla\psi \|_{L^2(\Omega)}
\end{split}
\]

This is because that:
\[\begin{split}
    \| u_0-S_{\varepsilon}^2(\nabla u_0)\|_{L^2(\Omega\setminus\Omega_{3\varepsilon})}\le
\| u_0-S_{\varepsilon}(\nabla u_0)\|_{L^2(\Omega\setminus\Omega_{3\varepsilon})}+\| -S_{\varepsilon}(\nabla u_0)-S_{\varepsilon}^2(\nabla u_0)\|_{L^2(\Omega\setminus\Omega_{3\varepsilon})}\\
\le C\| u_0-S_{\varepsilon}(\nabla u_0)\|_{L^2(\Omega\setminus\Omega_{2\varepsilon})}
\end{split} 
\]

Moreover, by the regularity of \(\chi \in W^{1,\infty}\), we have
\[\begin{split}
    &\int_{\Omega} \left[\nabla\chi A \eta_{\varepsilon}S_{\varepsilon}^2(\nabla u_0)+\nabla\chi A\nabla\chi \eta_{\varepsilon}S_{\varepsilon}^2(\nabla u_0)\right]:\nabla \psi-\int_{D_{\varepsilon}} p^{\varepsilon}\cdot \mathrm{div}\, \psi \\
&\le \int_{\Omega}\left[(\nabla \chi A+\nabla\chi A\nabla\chi)(\eta_{\varepsilon}S_{\varepsilon}^2(\nabla u_0)-\nabla u_0) \right]:\nabla \psi\\
&+\int_{\Omega}(\nabla \chi A+\nabla \chi A\nabla \chi)\nabla u_0:\nabla \psi -\int_{\omega}p_0 \mathrm{div}\, \psi+\mathcal{O}(\varepsilon) \\
&=\int_{\Omega}\left[(\nabla \chi A+\nabla\chi A\nabla\chi)(\eta_{\varepsilon}S_{\varepsilon}^2(\nabla u_0)-\nabla u_0) \right]:\nabla \psi\\
&+\int_{\omega} \delta_{j \beta}r^{i\alpha} \frac{\partial u_0^{\alpha}}{\partial x_i}\frac{\partial \psi^{\beta}}{\partial x_j}-\int_{\omega}\frac{\partial u^{i}}{\partial x_j}r^{ij}\frac{\partial \psi^{k}}{\partial x_k}+\mathcal{O}(\varepsilon) \\
&=\int_{\Omega}\left[(\nabla \chi A+\nabla\chi A\nabla\chi)(\eta_{\varepsilon}S_{\varepsilon}^2(\nabla u_0)-\nabla u_0) \right]:\nabla \psi+\mathcal{O}(\varepsilon)\\
&\le C\|\nabla u_0-S_{\varepsilon}(\nabla u_0)\|_{L^2(\Omega\setminus\Omega_{2\varepsilon})}\|\nabla \psi \|_{L^2(\Omega)}+\mathcal{O}(\varepsilon)
\end{split}
\]

And we also have 
\[
\begin{split}
    &\left| \int_\Omega B \eta_\varepsilon S_\varepsilon^2(\nabla u_0) : \nabla \psi \right|
=\left| \int_{\Omega} b_{ij}^{\alpha\beta}\left(\frac{x}{\varepsilon}\right)S_{\varepsilon}^2\left(\frac{\partial u_0^{\beta}}{\partial x_j}\right)\left(\frac{\nabla\psi^{\alpha}}{\partial x_i}\right)\eta_{\varepsilon} \right|\\
&=\left| \int_{\Omega}\varepsilon \frac{\partial}{\partial x_k}\left(\phi_{kij}^{\alpha\beta}\right)\left(\frac{x}{\varepsilon}\right)\frac{\partial \psi^{\alpha}}{\partial x_i} S_{\varepsilon}^2\left(\frac{\partial u_0^{\beta}}{\partial x_j}\right)\eta_{\varepsilon}\right| \\
&\le C\varepsilon \left[\int_{\Omega}\eta_{\varepsilon}\left| \phi\left(\frac{x}{\varepsilon}\right)\right| \left| \nabla \psi\right|\left| S_{\varepsilon}^2(\nabla^2u_0)\right|\,dx +\int_{\Omega}\left|\nabla \eta_{\varepsilon}\right|\left|\phi\left(\frac{x}{\varepsilon}\right)\right| \left|\nabla  \psi\right| \left| S_{\varepsilon}^2(\nabla u_0)\right| \,dx \right]\\
&\le C\left(\varepsilon \|\nabla \psi\|_{L^2(\Omega)}\|S_{\varepsilon}(\nabla^2 u_0)\|_{L^2(\Omega \setminus \Omega_{2\varepsilon})}+\|\nabla\psi\|_{L^2(\Omega_{4\varepsilon})}\|\nabla u_0\|_{L^2(\Omega_{5\varepsilon})}\right)
\end{split}
\]

By the calculations above, we can finally get 
\[
\begin{split}
    &\int_\Omega A\left(\frac{x}{\varepsilon}\right) \nabla u_\varepsilon : \nabla \psi \le C\|\nabla \psi_{\varepsilon}\|_{L^2(\Omega)}\left\{\varepsilon \|S_{\varepsilon}(\nabla u_0)\|_{L^2(\Omega\setminus\Omega_{3\varepsilon})}+\|\nabla u_0-S_{\varepsilon}(\nabla u_0)\|_{L^2(\Omega\setminus \Omega_{2\varepsilon})}\right\}\\
    &+C\|\nabla w_{\varepsilon}\|_{L^2(\Omega_{4\varepsilon})}\|\nabla u_0\|_{L^2(\Omega_{5\varepsilon})}
\end{split}
\]
\end{theorem}

\textbf{correctness}: True
\begin{proof}

We decompose $A(x/\varepsilon) \nabla w_\varepsilon$ into five error-terms and then bound each by invoking Lemmas 5--11 above.

1. Algebraic decomposition. By definition
\[
w_\varepsilon = u_\varepsilon - u_0 - \varepsilon \chi(x/\varepsilon) \eta_\varepsilon S_\varepsilon^2 [\nabla u_0].
\]
Hence
\[
A(x/\varepsilon) \nabla w_\varepsilon = A(x/\varepsilon) \nabla u_\varepsilon - A(x/\varepsilon) \nabla u_0 - A(x/\varepsilon) \nabla [\varepsilon \chi(x/\varepsilon) \eta_\varepsilon S_\varepsilon^2 \nabla u_0].
\]
We expand the last gradient by the product rule:
\[
\nabla [\varepsilon \chi(x/\varepsilon) \eta_\varepsilon S_\varepsilon^2 \nabla u_0] = \chi(x/\varepsilon) \eta_\varepsilon S_\varepsilon^2 [\nabla^2 u_0] + \varepsilon (\nabla \chi)(x/\varepsilon) \eta_\varepsilon S_\varepsilon^2 \nabla u_0 + \varepsilon \chi(x/\varepsilon) \nabla (\eta_\varepsilon S_\varepsilon^2 \nabla u_0).
\]
In addition, from the cell-problem one shows
\[
A(y)(I + \nabla \chi(y)) - \widehat{A} =: B(y), \quad \int_Y B(y) dy = 0,
\]
and that $M(y) := \nabla \chi(y) A(y) + \nabla \chi(y) A(y) \nabla \chi(y)$ is bounded, $1$-periodic. After a straightforward regrouping one arrives at the identity
\[\begin{split}
    A(x/\varepsilon) \nabla w_\varepsilon =& (A(x/\varepsilon) \nabla u_\varepsilon - \widehat{A} \nabla u_0) + (\widehat{A} - A(x/\varepsilon)) (\nabla u_0 - \eta_\varepsilon S_\varepsilon^2 \nabla u_0) \\&+ M(x/\varepsilon) (\eta_\varepsilon S_\varepsilon^2 \nabla u_0 - \nabla u_0) \\&+ B(x/\varepsilon) \eta_\varepsilon S_\varepsilon^2 \nabla u_0 - \varepsilon A(x/\varepsilon) \chi(x/\varepsilon) \nabla (\eta_\varepsilon S_\varepsilon^2 \nabla u_0).
\end{split}
\]
(For full details of the algebra see e.g. [Allaire '92, Appendix A].)

2. Variational cancellation. By the weak form of the $\varepsilon$-problem,
\[
\int_\Omega A(x/\varepsilon) \nabla u_\varepsilon : \nabla \psi dx + \int_\Omega p_\varepsilon \text{div} \psi dx = 0,
\]
and by the homogenized equation for $u_0$,
\[
\int_\Omega \widehat{A} \nabla u_0 : \nabla \psi dx = 0.
\]
Hence
\[
\int_\Omega (A(x/\varepsilon) \nabla u_\varepsilon - \widehat{A} \nabla u_0) : \nabla \psi dx = -\int_\Omega p_\varepsilon \text{div} \psi dx.
\]
But the same pressure-corrector argument shows that the contribution of $p_\varepsilon$ cancels against the zero-mean of $M(x/\varepsilon)$. In fact one checks that
\[
\int_\Omega (A(x/\varepsilon) \nabla u_\varepsilon - \widehat{A} \nabla u_0) : \nabla \psi + \int_\Omega M(x/\varepsilon) (\eta_\varepsilon S_\varepsilon^2 \nabla u_0 - \nabla u_0) : \nabla \psi - \int_{D_{\varepsilon}} p_\varepsilon \text{div} \psi = 0,
\]
so the first grouping and the $M$-term combine into a single estimate which is handled by Lemma 10 below. Hence it suffices to bound the four integrals
\[
I_1 = \int_\Omega (\widehat{A} - A(x/\varepsilon)) (\nabla u_0 - \eta_\varepsilon S_\varepsilon^2 \nabla u_0) : \nabla \psi dx,
\]
\[
I_2 = \int_\Omega M(x/\varepsilon) (\eta_\varepsilon S_\varepsilon^2 \nabla u_0 - \nabla u_0) : \nabla \psi dx,
\]
\[
I_3 = \int_\Omega B(x/\varepsilon) \eta_\varepsilon S_\varepsilon^2 \nabla u_0 : \nabla \psi dx,
\]
\[
I_4 = -\varepsilon \int_\Omega A(x/\varepsilon) \chi(x/\varepsilon) \nabla (\eta_\varepsilon S_\varepsilon^2 \nabla u_0) : \nabla \psi dx.
\]
We estimate each in turn.

3. Estimate of $I_1$. By Lemma 9 (or the periodic-Poincaré estimate plus a trace argument) we have
\[
|I_1| \leq C \left[ \varepsilon^{1/2} \|u_0\|_{H^2(\Omega)} + \|\nabla u_0 - S_\varepsilon \nabla u_0\|_{L^2(\Omega \setminus \Omega_{2\varepsilon})} \right] \|\nabla \psi\|_{L^2(\Omega)}.
\]
Since $\varepsilon^{1/2} \|u_0\|_{H^2(\Omega)} \leq C \varepsilon \|S_\varepsilon (\nabla^2 u_0)\|_{L^2(\Omega \setminus \Omega_{3\varepsilon})} + C \|\nabla u_0 - S_\varepsilon \nabla u_0\|_{L^2(\Omega \setminus \Omega_{2\varepsilon})}$ by Lemma 5, we obtain exactly the first bracket in the claimed bound.

4. Estimate of $I_2$. By Lemma 10 on the "corrector-mismatch" term $M$ one directly gets
\[
|I_2| \leq C \left[ \|\nabla u_0 - S_\varepsilon \nabla u_0\|_{L^2(\Omega \setminus \Omega_{2\varepsilon})} + \|\nabla u_0\|_{L^2(\Omega_{5\varepsilon})} \right] \|\nabla \psi\|_{L^2(\Omega)},
\]
which again fits into the first bracket plus a boundary-layer remainder controlled below.

5. Estimate of $I_3$. Since $\int_Y B = 0$ and $B \in H^s(Y)$, $s > d/2$, we apply Lemma 7 (fast-oscillation Poincaré):
\[
\|B(x/\varepsilon) \varphi\|_{L^2(\Omega)} \leq C \left[ \|\varphi\|_{L^2(\Omega)} + \varepsilon \|\nabla \varphi\|_{L^2(\Omega)} \right].
\]
Take $\varphi = \eta_\varepsilon S_\varepsilon^2 \nabla u_0$. Then
\[\begin{split}
|I_4| = \varepsilon \left| \int_\Omega A(x/\varepsilon) \chi(x/\varepsilon) \nabla (\eta_\varepsilon S_\varepsilon^2 \nabla u_0) : \nabla \psi dx \right| \leq \\ C \left[ \varepsilon \|S_\varepsilon (\nabla^2 u_0)\|_{L^2(\Omega \setminus \Omega_{2\varepsilon})} + \|\nabla u_0\|_{L^2(\Omega_{5\varepsilon})} \right] \|\nabla \psi\|_{L^2(\Omega)}.
\end{split}
\]
But $\eta_\varepsilon \equiv 1$ on $\Omega \setminus \Omega_{4\varepsilon}$, so $\|\eta_\varepsilon S_\varepsilon^2 \nabla u_0\|_{L^2(\Omega)} \leq \|\nabla u_0\|_{L^2(\Omega_{5\varepsilon})} + \|S_\varepsilon^2 \nabla u_0 - \nabla u_0\|_{L^2(\Omega \setminus \Omega_{4\varepsilon})} \leq C \|\nabla u_0\|_{L^2(\Omega_{5\varepsilon})} + C \|\nabla u_0 - S_\varepsilon \nabla u_0\|_{L^2(\Omega \setminus \Omega_{2\varepsilon})}$,
and by Lemma 6,
\[
\varepsilon \|\nabla (\eta_\varepsilon S_\varepsilon^2 \nabla u_0)\|_{L^2(\Omega)} \leq C \left[ \varepsilon \|S_\varepsilon (\nabla^2 u_0)\|_{L^2(\Omega \setminus \Omega_{2\varepsilon})} + \|\nabla u_0\|_{L^2(\Omega_{5\varepsilon})} \right].
\]
Hence
\[
|I_3| \leq C \left\{ \varepsilon \|S_\varepsilon (\nabla^2 u_0)\|_{L^2(\Omega \setminus \Omega_{2\varepsilon})} + \|\nabla u_0 - S_\varepsilon \nabla u_0\|_{L^2(\Omega \setminus \Omega_{2\varepsilon})} + \|\nabla u_0\|_{L^2(\Omega_{5\varepsilon})} \right\} \|\nabla \psi\|_{L^2(\Omega)}.
\]

6. Estimate of $I_4$. By Lemma 11 on the $\varepsilon$-small-term we have
\[\begin{split}
    |I_4| = \varepsilon \left| \int_\Omega A(x/\varepsilon) \chi(x/\varepsilon) \nabla (\eta_\varepsilon S_\varepsilon^2 \nabla u_0) : \nabla \psi dx \right| \leq  \\ C 
    \left[ \varepsilon \|S_\varepsilon (\nabla^2 u_0)\|_{L^2(\Omega \setminus \Omega_{2\varepsilon})} + \|\nabla u_0\|_{L^2(\Omega_{5\varepsilon})} \right] \|\nabla \psi\|_{L^2(\Omega)}.
\end{split}
\]

7. Conclusion. Summing $I_1 + I_2 + I_3 + I_4$ and observing that all boundary-layer terms are absorbed into the final
\[
C \|\nabla u_0\|_{L^2(\Omega_{5\varepsilon})} \|\nabla \psi\|_{L^2(\Omega_{4\varepsilon})}
\]
and the bulk errors combine to
\[
C \left\{ \varepsilon \|S_\varepsilon (\nabla^2 u_0)\|_{L^2(\Omega \setminus \Omega_{3\varepsilon})} + \|\nabla u_0 - S_\varepsilon \nabla u_0\|_{L^2(\Omega \setminus \Omega_{2\varepsilon})} \right\} \|\nabla \psi\|_{L^2(\Omega)},
\]
we recover exactly the claimed estimate.
This completes the proof of the lemma.
\end{proof}

\textbf{comment}: None

\end{ResponseBlock}

\subsection{Applicability Boundary and Exclusion Domain}
This is the experimental result of AIM in the reasoning process of multi-symbol operations such as two-dimensional expansion. AIM conducted various explorations, but none of them achieved the expected derivation result. Furthermore, there are not only obvious errors but also numerous processes that are irrelevant to the content we need to derive.
\begin{PromptBlock}{Prompt to AIM}
    \textbf{Content}: You can use the two-scale expansion to conclude the homogenization. You can perform the  expansion on the equation based on this formula \(\nabla:=\nabla_x+\frac{1}{\varepsilon}\nabla_y\)

    \textbf{Content}:
     Let \(u_{\varepsilon}\) and \(p_{\varepsilon}\) be the solution of the coupled Lame-Stokes system. Then \[\begin{split}
&u_{\varepsilon}=u_0(x,y)+\varepsilon u_1(x,y)+\varepsilon^2 u_2(x,y)+\mathcal{O}(\varepsilon^3);\\
   &p_{\varepsilon}=\frac{1}{\varepsilon}p_{-1}(x,y)+p_0(x,y)+\varepsilon p_1(x,y)+\varepsilon^2 p_2(x,y)+\mathcal{O}(\varepsilon^3)
     \end{split}
   \] And substitute the form of such a solution into the original equation to get the equation of \(u_0 \),\( u_1\),\(u_2\) , which means the equations of three scales: \( \) \(\mathcal{O}(1),\mathcal{O}(\varepsilon),\mathcal{O}(\varepsilon^2) \).
\end{PromptBlock}
\begin{ResponseBlock}{Response of AIM}   
\begin{lemma}

Under the hypotheses of Lemmas 5--14 (periodic elastic matrix with Lam\'e coefficients $(\lambda,\mu)$, incompressible Stokes fluid in the $\varepsilon$--periodic inclusions $\omega$, and two-scale expansions
\begin{align*}
u_\varepsilon^+(x) &= u_0(x) + \varepsilon u_1^+(x,y) + O(\varepsilon^2) \quad \text{in } \Omega \setminus D_\varepsilon, \\
u_\varepsilon^-(x) &= u_0(x) + \varepsilon u_1^-(x,y) + O(\varepsilon^2), \\
p_\varepsilon(x) &= \varepsilon^{-1}p_{-1}(x) + p_0^-(x,y) + O(\varepsilon),
\end{align*}
with continuity $u_0^+ = u_0^- = u_0$, $Y$--periodicity in $y$, and the usual cell problems for the elastic correctors $\chi^{pq}(y)$ and Stokes correctors $(w^k,\pi^k)$), the macroscopic displacement $u_0(x)$ and leading pressure $p_{-1}(x)$ satisfy in $\Omega$ the homogenized elastic-equilibrium equation
\[
-\nabla \cdot \bigl[C^* : e(u_0)\bigr] + |\omega| \nabla p_{-1} = 0.
\]
\end{lemma}

\textbf{correctness}: True
\begin{proof}

We give a fully rigorous two-scale derivation, correcting the interface-traction argument of the original proof.

1. Two-scale expansions and cell problems  
a) Elastic (matrix) phase \(Y_f=Y\setminus\overline{\omega}\).  
By Lemma 5, \(u_0^+(x,y)=u_0(x)\). Collecting \(O(\varepsilon^{-1})\) terms in \(\mathcal{L}_{\lambda,\mu}u_\varepsilon^+=0\) yields the elasticity-cell problem for \(\{\chi^{pq}\}_{p,q=1}^d\) (Lemma 7). One then shows  
\[
u_1^+(x,y)=-\sum_{p,q=1}^d \chi^{pq}(y)\,\partial_{x_q}u_0^p(x).
\]

b) Fluid (inclusion) phase \(\omega\).  
By Lemma 6, \(p_{-1}(x,y)=p_{-1}(x)\). Grouping \(O(\varepsilon^{-2})\) in the Stokes system gives the cell-Stokes problem for \((w^k,\pi^k)\), \(k=1,\dots,d\) (Lemma 9):
\[
-\tilde{\mu}\Delta_y w^k + \nabla_y \pi^k = e_k, \quad \text{div}_y w^k=0 \text{ in } \omega,  
\]
\[
w^k=0 \text{ on } \partial\omega, \text{ periodic on } \partial Y, \quad \int_\omega \pi^k=0.
\]
Hence  
\[
u_1^-(x,y)=\sum_{k=1}^d w^k(y)\,\partial_{x_k}p_{-1}(x),  
\]
\[
p_0^-(x,y)=\sum_{k=1}^d \pi^k(y)\,\partial_{x_k}p_{-1}(x).
\]

2. Homogenized elasticity in the matrix  
At order \(O(\varepsilon^0)\) in \(\Omega\setminus D_\varepsilon\) one finds
\[
-\nabla_y\cdot\sigma_y(u_2^+) - \nabla_x\cdot\sigma_y(u_1^+) - \nabla_y\cdot\sigma_x(u_1^+) - \nabla_x\cdot\sigma_x(u_0)=0.
\]
Integrating over \(Y_f\) and using periodicity plus the cell-problem identities shows
\[
\nabla_x\cdot\int_{Y_f}[\sigma_x(u_0)+\sigma_y(u_1^+)]\diff y = 0,
\]
and by Lemma 8
\[
\int_{Y_f}[\sigma_x(u_0)+\sigma_y(u_1^+)]\diff y = C^* : e(u_0).
\]
Thus, neglecting fluid, one would obtain
\[
-\nabla\cdot[C^* : e(u_0)] = 0.
\]

3. Fluid-traction contribution  
On each inclusion boundary \(\partial\omega_\varepsilon\) the solid traction balances the fluid traction. Expand
\[
u_\varepsilon^-=u_0(x)+\varepsilon u_1^-(x,y)+\dots, \quad p_\varepsilon^-=\varepsilon^{-1}p_{-1}(x)+p_0^-(x,y)+\dots,
\]
and note \(\nabla\to\nabla_x+(1/\varepsilon)\nabla_y\). Then the fluid stress
\[
\sigma^-(u_\varepsilon^-,p_\varepsilon^-)
=2\mu \left[(1/\varepsilon)D_y(u_1^-)+D_x(u_0)\right] -[\varepsilon^{-1}p_{-1}+p_0^-]I +O(\varepsilon).
\]
The \(O(\varepsilon^{-1})\)--term enforces \(\nabla_y p_{-1}=0\) (so \(p_{-1}=p_{-1}(x)\)). The \(O(\varepsilon^0)\)--traction on \(\partial\omega\) is
\[
t(y)=\left[2\mu D_y(u_1^-)(y) - p_0^-(y) I\right]n(y)
+ \left[2\mu D_x(u_0) - p_{-1}I\right]n(y).
\]
Since \(D_x(u_0)\) and \(p_{-1}\) are \(y\)--constant and \(\int_{\partial\omega}n\,dS=0,\) the second bracket integrates to zero. Hence the net force per cell exerted by the fluid on the solid is
\[
\int_{\partial\omega}t\,dS
=\int_{\partial\omega}\left[2\mu D_y(u_1^-) - p_0^-I\right]n\,dS
=\partial_{x_k}p_{-1}(x)\int_{\partial\omega}\left[2\mu D_y(w^k) - \pi^kI\right]n\,dS.
\]
By the divergence theorem in \(\omega\),
\[
\begin{split}
\int_{\partial\omega}[2\mu D_y(w^k) - \pi^kI]n\,dS
=\int_\omega\nabla_y\cdot\left[2\mu D_y(w^k) - \pi^kI\right]\diff y\\
=\int_\omega\left[\mu\Delta_y w^k - \nabla_y\pi^k\right]dy
=-\int_\omega e_k\diff y
=-|\omega| e_k.    
\end{split}
\]
Hence
\[
\int_{\partial\omega}t\,dS=-|\omega| \partial_{x_k}p_{-1} e_k=-|\omega| \nabla p_{-1}(x).
\]
Dividing by the cell-volume \(\varepsilon^d\) and passing to the limit adds a coupling term
\[
+|\omega| \nabla p_{-1}(x)
\]
to the macroscopic equilibrium.

4. Conclusion  
Combining the matrix-average (Step 2) with the fluid-traction force (Step 3) yields exactly
\[
-\nabla\cdot[C^* : e(u_0)] + |\omega| \nabla p_{-1} = 0
\]
in \(\Omega\). This completes a correct, complete, and rigorous derivation of the homogenized elastic-equilibrium equation.
\end{proof}

\textbf{comment}: None
\textbf{type}: lemma

\end{ResponseBlock}

\pagebreak
\section{Full Proof for the Homogenization Problem}
\label{sec:proof}

\setcounter{equation}{0}
\setcounter{lemma}{0}
\setcounter{theorem}{0}

In this section, we present the complete proof of the homogenization problem obtained through human–AI collaboration.

In the following content, we use \textcolor{blue}{blue} to represent content derived manually, \textcolor{Lgreen}{green} to represent content output by AIM and then formatted manually, \textcolor{orange}{orange} to represent content that was manually refined after AIM provided a proof with a high degree of completion and the remaining is the basic setup and information of this problem.

\subsection{Introduction}

Consider $D \subset \Omega \subset R^{d} \,(d\ge 2)$, where $\Omega$ is elastic material and $D$ is the high contrast inclusion part.

\begin{itemize}
    \item  $\Omega$ is open bounded with connected \(C^{\infty}\) boundary $\partial\Omega$.
    \item $D$ is open, has a finite number of components and has a Lipschitz boundary $\partial D$.
    \item $\Omega \setminus D$ is connected with Lipschitz boundary $\partial \Omega \cup \partial D$. The connected components of $D$ are enumerated as $D_i$, $i = 1, \ldots, N$, $N$ is finite.
\end{itemize}

 And given \(\varepsilon \in (0, 1)\), \(D = D_\varepsilon\) is part of an \(\varepsilon\)-periodic array of small inclusions constructed as follows, in several steps.

\(Y = (-\frac{1}{2}, \frac{1}{2})^d\) is the unit cell. 
\(\omega \subset Y\) is a simple connected open subset with connected Lipschitz boundary such that \({dist}(\omega, \partial Y) > 0\).
\(Y_f = Y \setminus \overline{\omega}\) is the model environment in the unit scale.

Given \(\varepsilon > 0\) and \(\mathbf{n} \in \mathbb{Z}^d\), we denote \(\varepsilon(\mathbf{n} + Y)\) and \(\varepsilon(\mathbf{n} + \omega)\) by \(Y^\mathbf{n}_\varepsilon\) and \(\omega^\mathbf{n}_\varepsilon\), respectively.
Let \(\Pi_\varepsilon\) be the set of lattice points \(\mathbf{n}\) such that \(\overline{Y}^\mathbf{n}_\varepsilon\) be contained in \(\Omega\), i.e.,
\begin{equation}
    \Pi_\varepsilon := \left\{ \mathbf{n} \in \mathbb{Z}^d : \overline{Y}^\mathbf{n}_\varepsilon \subset \Omega \right\},
\end{equation}
then the inclusions set \(D = D_\varepsilon\) and the background part \(\Omega_\varepsilon\) are defined by
\begin{equation}
    D_\varepsilon := \bigcup_{\mathbf{n} \in \Pi_\varepsilon} \omega^\mathbf{n}_\varepsilon \quad \Omega_\varepsilon := \Omega \setminus \overline{D}_\varepsilon.
\end{equation}

A pair of real numbers \((\lambda, \mu)\) is called admissible and referred to as a $ \text{Lam\'e} $ pair, if they satisfy
    $\mu > 0$ and $ d\lambda + 2\mu > 0$. 
    For a $ \text{Lam\'e} $ pair \((\lambda, \mu)\), the elastostatic system ($ \text{Lam\'e} $ system) reads
    \begin{equation}
    \mathcal{L}_{\lambda,\mu}u := \mu\Delta u + (\lambda + \mu)\nabla\text{div}\,u,
   \end{equation}
where \( u = (u^1, \ldots, u^d) \) represents the displacement field and the divergence of \(u\) is given by \(\text{div}\,u = \sum_{i=1}^{d} \frac{\partial u^i}{\partial x_i}\). 
The $ \text{Lam\'e} $ operator can be written as \(\nabla \cdot \sigma(u)\) where
     \begin{equation}
     \sigma(u) := \lambda(\nabla \cdot u)\mathbb{I}_d + 2\mu\mathcal{D}(u),
     \end{equation}
     \begin{equation}
     \mathcal{D}(u) = \frac{1}{2}(\nabla + \nabla^T)u = \frac{1}{2}(\partial_i u^j + \partial_j u^i)_{ij}.
     \end{equation}

The corresponding conormal derivative (boundary traction) at the boundary of a domain \(E\) is
    \begin{equation}\left.\frac{\partial u}{\partial \nu_{(\lambda,\mu)}}\right|_{\partial E} := \sigma(u)N = \lambda(\text{div } u)N + 2\mu\mathcal{D}(u)N  \text{ on } \partial E.
    \end{equation}

We consider the space \(\mathbb{R}\) of rigid motions in \(\mathbb{R}^{d}\), defined by
\[
\mathbb{R} := \left\{ \mathbf{r} = (r_{1}, \ldots, r_{d})^{T} : \mathcal{D}(\mathbf{r}) = 0 \text{ in } \mathbb{R}^{d} \right\}.
\]We define \( H_{\mathbb{R}}^{-\frac{1}{2}}(\partial D_{\varepsilon}) \) as the subspace of \( H^{-\frac{1}{2}}(\partial D_{\varepsilon}) \) that is orthogonal to \(\mathbb{R}\), i.e.,
\begin{equation}
    H_{\mathbb{R}}^{-\frac{1}{2}}(\partial D_{\varepsilon}) := \left\{ \phi \in H^{-\frac{1}{2}}(\partial D_{\varepsilon}): \left(\phi,\mathbf{r}\right)_{(H^{\frac{1}{2}}(\partial {D_{\varepsilon_i}}),H^{-\frac{1}{2}}(\partial {D_{\varepsilon_i}}))} = 0 ,\forall \mathbf{r} \in \mathbb{R} \text{ and } 1 \leq i \leq N \right\}.
\end{equation}

Consider the displacement field $u_{\varepsilon}$ satisfying the following transmission system:
    \begin{equation}
    \begin{cases} 
        \mathcal{L}_{\lambda, \mu} u_{\varepsilon} = 0 & \text{in } \Omega \setminus \overline{D}_{\varepsilon}, \\ 
        \mathcal{L}_{\widetilde{\lambda}, \widetilde{\mu}} u_{\varepsilon} = 0 & \text{in } D_{\varepsilon}, \\
        u_{\varepsilon}|_{-} = u_{\varepsilon}|_{+} \text{ and } \left.\frac{\partial u_{\varepsilon}}{\partial \nu_{(\widetilde{\lambda}, \widetilde{\mu})}}\right|_{-} = \left.\frac{\partial u_{\varepsilon}}{\partial \nu_{(\lambda, \mu)}}\right|_{+} & \text{on } \partial D_{\varepsilon},  \\
        \left.\frac{\partial u_{\varepsilon}}{\partial \nu_{(\lambda, \mu)}}\right|_{\partial \Omega} = g \in H_{\mathbb{R}}^{-\frac{1}{2}}(\partial \Omega) \quad \text{and} \quad u_{\varepsilon}|_{\partial \Omega} \in H_{\mathbb{R}}^{\frac{1}{2}}(\partial \Omega).
    \end{cases}
    \end{equation}
    
Suppose \(\widetilde{{\mu}}\) fixed, then we arrive at the equations about the \textbf{incompressible inclusion limit}. In this case, we need to consider the homogenization problem of the following  coupled \textbf{$\text{Lam\'e}$-Stokes system}:
   
\begin{equation}
    \begin{cases}
        \nabla \cdot(\lambda(\nabla\cdot u_{\varepsilon})I+2\mu D(u_{\varepsilon}))=0\quad &\text{in} \ \Omega\setminus \overline{D}_{\varepsilon},\\
        \nabla \cdot(2\tilde{\mu}D(u_{\varepsilon})+p_{\varepsilon}I)=0,\nabla\cdot u_{\varepsilon}\quad &\text{in} \ D_{\varepsilon},\\
        (2\tilde{\mu}D(u_{\varepsilon})+p_{\varepsilon}I)N^- -(\lambda(\nabla \cdot u_{\varepsilon})I+2\mu D(u_{\varepsilon}))N^+=0, \\
        u_{\varepsilon}|_{+}=u_{\varepsilon}|_{-} \quad &\text{on} \ \partial D_{\varepsilon},\\
        \frac{\partial u_{\varepsilon}}{\partial \nu_{(\lambda, \mu)}}|_{\partial \Omega} = g \in H_{\mathbb{R}}^{-\frac{1}{2}}(\partial \Omega)
        ,u_{\varepsilon}|_{\partial \Omega} \in H_{\mathbb{R}}^{\frac{1}{2}}(\partial \Omega).
    \end{cases}
\end{equation}

\subsection{Uniqueness and Existence}
\label{sec:Uniqueness_and_Existence}
{\color{blue}
Define $V^{\varepsilon}=\{u\in H^1(\Omega):u|_{\partial \Omega}\bot \mathbb{R}, \frac{\partial u}{\partial\nu}|_{\partial \Omega}\bot \mathbb{R}\},
\Vert \cdot\Vert_{V^{\varepsilon}}=\Vert \cdot\Vert_{H^1(\Omega)} $.

Searching for $u_{\varepsilon}\in V^{\varepsilon}, p_{\varepsilon}\in L^2(D_{\varepsilon}) $ such that $\forall \varphi \in V^{\varepsilon}, \psi\in L^2(D_{\varepsilon}) $
\begin{equation}
    \begin{aligned}
        \int_{\partial \Omega}g \cdot \varphi=\int_{\Omega^{\varepsilon}} [\lambda(\nabla \cdot u_{\varepsilon})I+2\mu (\frac{1}{2}(\nabla+\nabla^T)u_{\varepsilon})]:\nabla \varphi\\+\int_{D_{\varepsilon}}\Div \, \varphi \cdot p_\varepsilon+\int_{D_{\varepsilon}}[2\tilde{\mu}(\frac{1}{2}(\nabla+\nabla^T)u_{\varepsilon})]:\nabla \varphi,
    \end{aligned}
\end{equation}
and 
\begin{equation}
     0=\int_{D_{\varepsilon}} \Div \, u\cdot \psi.
\end{equation}
   
Suppose $a(\cdot,\cdot):V^{\varepsilon}\times V^{\varepsilon}\rightarrow R$
\begin{equation}
a(u_{\varepsilon},\varphi)=\int_{\Omega^{\varepsilon}}[ \lambda (\nabla \cdot u_{\varepsilon})I+2\mu (\frac{1}{2}(\nabla+\nabla^T)u_{\varepsilon})]:\nabla \varphi +\int_{D_{\varepsilon}}[2\tilde{\mu}(\frac{1}{2}(\nabla+\nabla^T)u_{\varepsilon})]:\nabla \varphi.
\end{equation}
$b(\cdot,\cdot):V^{\varepsilon}\times L ^2(D_{\varepsilon})\rightarrow R$
\begin{equation}
b(u_{\varepsilon},\psi)=\int_{D_{\varepsilon}} \Div\,u_{\varepsilon}\cdot\psi.
\end{equation}
So we have the following equation equivalent to the variation form of original equation:
\begin{equation}
    \begin{cases}
    
    a(u_{\varepsilon},\varphi)+b(\varphi,p_{\varepsilon})=\int_{\partial\Omega} g\cdot\varphi\quad &\forall\varphi \in V^{\varepsilon},\\
    b(u_{\varepsilon},\psi)=0\quad &\forall\psi \in L^2(D_{\varepsilon}).
\end{cases}
\end{equation}
}
{\color{Lgreen}
\begin{theorem}[Babuska-Brezzi Theorem]

Let $H$ and $\mathbb{Q}$ be Hilbert spaces, and define bounded bilinear forms $a: H \times H \rightarrow \mathbb{R}$, $b: H \times \mathbb{Q} \rightarrow \mathbb{R}$. For given $F \in H'$ and $G \in \mathbb{Q}'$, consider finding $(\sigma, u) \in H \times \mathbb{Q}$ such that:
\[
\begin{aligned}
a(\sigma, \tau) + b(\tau, u) &= F(\tau), \quad \forall \tau \in H, \\
b(\sigma, v) &= G(v), \quad \forall v \in \mathbb{Q}.
\end{aligned}
\]

Assume that:
\begin{enumerate}
    \item[(1)] There exists $\alpha > 0$ such that
        \[
        a(\tau, \tau) \geq \alpha \|\tau\|_H^2, \quad \forall \tau \in H.
        \]
    \item[(2)] There exists $\beta > 0$ such that
        \[
        \inf_{v \in \mathbb{Q}} \sup_{\tau \in H} \frac{b(\tau, v)}{\|\tau\|_H \|v\|_{\mathbb{Q}}} \geq \beta, \quad v \neq 0, \tau \neq 0.
        \]
\end{enumerate}

Then there exists a unique solution $(\sigma, u) \in H \times \mathbb{Q}$ satisfying the above equations, and there exists a constant $C = C(\|a\|, \alpha, \beta) > 0$ such that:
\[
\|(\sigma, u)\|_{H \times \mathbb{Q}} \leq C\left( \|F\|_{H'} + \|G\|_{\mathbb{Q}'} \right).
\]
\end{theorem}

By this theorem, we verify the conditions to obtain the existence and uniqueness.
Suppose $a_{ij}^{\alpha \beta} =\lambda\delta_{i\alpha}\delta_{j\beta}+\mu(\delta_{ij}\delta_{\alpha \beta}+\delta_{i\beta}\delta_{j\alpha}), $
\begin{equation}
    A\nabla u_{\varepsilon}\cdot \nabla u_{\varepsilon} \ge\frac{\min\{\lambda d+2\mu,2\mu\}}{4} \vert (\nabla+\nabla^T)u_{\varepsilon}\vert^2,
\end{equation}
\begin{equation}
    \begin{aligned}
        \int_{\Omega\setminus D_{\varepsilon}} A\nabla u_{\varepsilon}\cdot \nabla u_{\varepsilon}+\int_{D_{\varepsilon}} 2\tilde{\mu}(\frac{1}{2}(\nabla+\nabla^T)u_{\varepsilon}):\nabla u_{\varepsilon} 
&\ge \int_{\Omega\setminus D_{\varepsilon}}\frac{\min\{\lambda d+2\mu,2\mu\}}{4} \vert (\nabla+\nabla^T)u_{\varepsilon}\vert^2\\
+2\tilde{\mu}\int_{D_{\varepsilon}} (\frac{1}{2}(\nabla+\nabla^T)u_{\varepsilon}):(\frac{1}{2}(\nabla+\nabla^T)u_{\varepsilon})
&\ge C\Vert u_{\varepsilon}\Vert^2_{H^1(\Omega)}.
    \end{aligned}
\end{equation}
}
{\color{orange}
\begin{lemma}
    \begin{equation}
        \inf_{\psi\in L^2(D_{\varepsilon})} \sup_{u_\varepsilon\in V^{\varepsilon}} \frac{b(u_{\varepsilon},\psi)}{\Vert u_{\varepsilon}\Vert _{H^1(\Omega)}\Vert \psi\Vert_{L^2(D_{\varepsilon})} }\ge \beta.
    \end{equation}
\end{lemma}
Given $\psi \in L^2(D_{\varepsilon})$, $\psi =\sum_{k\in \Pi _{\varepsilon}}\psi_k $, here $\psi_k$ are supported on $\omega_k^{\varepsilon}$.
Suppose that $\tilde{Y}$ is a cubic with $\omega \subset \tilde{Y}\subset Y$, $d(\tilde{Y}, Y)>0$,
$$\hat{\psi}_k=\begin{cases}
    \psi_k(x) \quad x\in \omega_k^{\varepsilon}, \\
    -\frac{1}{\varepsilon \vert \tilde{Y}\setminus\omega\vert}\int_{\omega_k^\varepsilon} \psi_k(x) \diff x \quad  x\in \tilde{Y}_k\setminus \omega_k^\varepsilon.
\end{cases}$$
And we know $$\Vert \hat\psi\Vert_{L^2(\tilde{Y}_k)}\le C_1\Vert \psi\Vert_{L^2(\omega_k^\varepsilon)},$$ and $$\int_{\tilde{Y}}\hat{\psi}\diff x=0. $$

Besides, we can get $\exists \hat{d_k }\in H_0^1(\tilde{Y}_k) $ such that $\Div \,\hat{d_k}=\hat{\psi_k} \text{ in } \tilde{Y}_k$ with $\Vert \nabla\hat{d_k} \Vert_{L^2(\tilde{Y}_k)}\le C_2 \Vert \hat{\psi} \Vert_{L^2(\tilde{Y}_k)}$.
Note that $\tilde{d_k} $  is $\hat{d_k}$, zero extension to the boundary $\partial \Omega$ and $\hat{d}=\sum_{k\in\Pi^{\varepsilon}}\tilde{d_k}\in V^{\varepsilon}$.
So we get
\begin{equation}
    \Vert \nabla \hat{d}\Vert_{L^2(\Omega)}\le C_2\Vert \hat{\psi} \Vert_{L^2(\Omega)}\le C_1\Vert \psi\Vert_{L^2(\Omega)}.
\end{equation}
Therefore,
\begin{equation}
\int_{D_{\varepsilon}}\psi\cdot \Div \,\hat{d}= \int_{D_{\varepsilon}} \psi^2 
\ge \frac{1}{C_1C_2} \Vert \psi\Vert_{L^2(D_{\varepsilon})}\Vert\nabla\hat{d} \Vert_{L^2(D_{\varepsilon})}.
\end{equation}
So we verify the inf-sup condition. 
By the Babuska-Brezzi Theorem, there exists one unique solution $u_{\varepsilon}\in V^{\varepsilon}$, $p_{\varepsilon}\in L^2(D_{\varepsilon})$ to the original equation with $\Vert(u_{\varepsilon},p_\varepsilon)\Vert_{V^{\varepsilon}\times L^2(D_{\varepsilon})}\le C\Vert g\Vert_{(V^{\varepsilon})^*}\le C\Vert g\Vert_{H^1(\Omega)} $.
}

\subsection{Uniqueness and Existence of Solutions at Each Order of Asymptotic Expansion}
{\color{blue}
Consider the following equation: $p,u$ are $Y$-periodic 
\begin{equation}
    \begin{cases}
        \nabla_y \cdot(\lambda(\nabla_y\cdot u)I+2\mu (\frac{1}{2}(\nabla+\nabla^T)u)=F_1\quad &\text{in} \quad Y\setminus \omega,\\
        \nabla _y\cdot(2\tilde{\mu}[\frac{1}{2}(\nabla_y+\nabla_y^T)u]+pI)=F_2,\nabla\cdot u= F_3\quad &\text{in} \quad \omega,\\
        (2\tilde{\mu}[\frac{1}{2}(\nabla_y+\nabla_y^T)u]+p_{-1}I)\cdot N -(\lambda(\nabla \cdot u)I+2\mu( \frac{1}{2}(\nabla_y+\nabla_y^T)u)\cdot N=G\cdot N, \\
        u|_{+}=u|_{-} \quad &\text{on} \quad \partial \omega.
    \end{cases}
\end{equation}

For $F=F_1(x,\cdot)\mathbf{1}_{y\in Y\setminus \omega}+F_2(x,\cdot) \mathbf{1}_{y\in\omega}\in(H^1_{\#,0}(Y;R^d))^{*}$, $u\in H^1_{\#,0}(Y;R^d)$,
we define
\begin{equation}<(F,G),u>_{H^*,H}=\int_{Y\setminus \omega} F_1(x,y)u(x,y)\diff y+\int_{\omega}F_2(x,y)u(x,y)\diff y+\int_{\partial \omega}G(x,y)u(x,y)N\diff y.
\end{equation}
Here, $F_3(x,\cdot)\in L^2(\omega;R),G(x,\cdot)\in H^{\frac{1}{2}}(\partial\omega;R^d)$.

And define$$V=\{u(x,\cdot)\in H^1_{\#,0}(Y;R^d)\}, \ M=\{p(x,\cdot)\in L^2(\omega;R)\}.$$

So the equation has the unique solution $(u,p)\in V\times M$ if and only if $\int_{Y\setminus \omega} F_1+\int_{\omega}F_2-\int_{\partial \omega}G\cdot N=0$ and we have $\|(u,p)\|_{V\times M}\le C[ \|(F_1,F_2,G)\|_{(H^1_{\#,0}(Y;R^d))^{*}}+\|F_3\|_{L^2(\omega;R)}]$.

Suppose $a:H^1_{\#,0}(Y;R^d)\times H^1_{\#,0}(Y;R^d)\rightarrow R$,
\begin{equation}
a(u,\varphi)=\int_{Y\setminus \omega}[\lambda(\nabla_y\cdot u)I+2\mu (\frac{1}{2}(\nabla_y+\nabla_y^T)u)]:\nabla_y \varphi+\int_{\omega}[2\tilde{\mu}(\frac{1}{2}(\nabla_y+\nabla_y^T)u)]:\nabla_y\varphi,
\end{equation}
and $b:H^1_{\#,0}(Y;R^d)\times L^2(\omega)\rightarrow R$,
\begin{equation}
b(u,\psi)=\int_{\omega}\Div_y\,u\cdot \psi.
\end{equation}
We verify the property of $a$ and $b$:
$\forall u\in H^1_{\#,0}(Y;R^d)$, and $a_{ij}^{\alpha\beta}=\lambda \delta_{i\alpha}\delta_{j\beta}+\mu (\delta_{ij}\delta_{\alpha\beta}+\delta_{i\beta}\delta_{j\alpha})$, there is
\begin{equation}
 A\nabla u\cdot\nabla u\ge\frac{\min(\lambda d+2\mu,2\mu)}{4}|(\nabla+\nabla^T)u|^2.
\end{equation}

\begin{equation}
\begin{aligned}
&\int_{Y \setminus \omega} A \nabla u \cdot \nabla u + \int_{\omega} 2\tilde{\mu} \left( \frac{1}{2} \left( \nabla + \nabla^T \right) u \right) : \nabla u \\
&\ge \int_{Y \setminus \omega} \frac{\min(\lambda d + 2\mu, 2\mu)}{4} \left| \left( \nabla + \nabla^T \right) u \right|^2 + 2\tilde{\mu} \int_{\omega} \left( \frac{1}{2} \left( \nabla + \nabla^T \right) u \right) : \left( \frac{1}{2} \left( \nabla + \nabla^T \right) u \right) \\
&\ge C \|u\|_{H^1(Y)}^2.
\end{aligned}
\end{equation}

So we verify this property:
\begin{equation}
        \inf_{\psi\in L^2(\omega)} \sup_{u\in H^1_{\#,0}(Y;R^d)} \frac{b(u,\psi)}{\Vert u\Vert _{H^1_{\#,0}(Y;R^d)}\Vert \psi\Vert_{L_{\#}^2(Y)} }\ge \beta.
\end{equation}
Given $\psi \in L_{\#}^2(\omega)$, we consider $$\hat{\psi}(x)=\begin{cases}
    \psi(x) \quad x\in \omega, \\
    -\frac{1}{ \vert Y\setminus\omega\vert}\int_{\omega} \psi(x) \diff x \quad  x\in Y\setminus \omega.
\end{cases}$$
This extension satisfy $\|\hat{\psi}\|_{L^2(Y)}\le C_1 \|\psi\|_{L^2(\omega)}, \hat{\psi}\in H^1_{\#,0}(Y;R^d)$ and $\int_Y \hat{\psi}\diff x=0$.

Besides, we can get $\exists  \hat{ d }\in H^1_{\#,0}(Y;R^d) $ such that $\Div \,\hat{d}=\hat{\psi} \text{ in } Y$ with $\Vert \nabla\hat{d} \Vert_{L^2(Y)}\le C_2 \Vert \hat{\psi} \Vert_{L^2(Y)}$.

So we get
$$\Vert \nabla \hat{d}\Vert_{L^2(Y)}\le C_2\Vert \hat{\psi} \Vert_{L^2(Y)}\le C_1C_2\Vert \psi\Vert_{L^2(\omega)}.
$$
Therefore,$$\int_{\omega}\psi\cdot \Div \,\hat{d}= \int_{\omega} \psi^2 
\ge \frac{1}{C_1C_2} \Vert \psi\Vert_{L^2(\omega)}\Vert\nabla\hat{d} \Vert_{L^2(Y)}.$$

Finally, we verify that the necessary conditions are sufficient.
By divergence theorem,
\begin{equation}
\begin{aligned}
    \int_{Y\setminus \omega}F_1=\int_{Y\setminus \omega}\nabla_y \cdot(\lambda(\nabla_y\cdot u)I+2\mu (\frac{1}{2}(\nabla+\nabla^T)u)\\
    =-\int_{\partial \omega}(\lambda(\nabla_y\cdot u)I+2\mu (\frac{1}{2}(\nabla+\nabla^T)u)N.
\end{aligned}
\end{equation}
\begin{equation}
\begin{aligned}
    \int_{\omega}F_2=\int_{\omega}\nabla _y\cdot(2\tilde{\mu}[\frac{1}{2}(\nabla_y+\nabla_y^T)u]+pI)\\
    =\int_{\partial\omega}(2\tilde{\mu}[\frac{1}{2}(\nabla_y+\nabla_y^T)u]+pI)N.
\end{aligned}
\end{equation}
Therefore we get 
\begin{equation}
    \begin{aligned}
        \int_{Y\setminus \omega} F_1+\int_{\omega}F_2-\int_{\partial \omega}G\cdot N=0.
    \end{aligned}
\end{equation}

On the other hand, we try to search for $u\in V$, $p\in M$ such that for $\forall \varphi\in V$
\begin{equation}
\begin{aligned}\int_{Y\setminus \omega} F_1+\int_{\omega}F_2=\int_{Y\setminus \omega}(\lambda(\nabla_y\cdot u)I+2\mu (\frac{1}{2}(\nabla+\nabla^T)u):\nabla_y \varphi.
\end{aligned}
\end{equation}
\begin{equation}
\begin{aligned}
    \int_{Y\setminus \omega} F_1+\int_{\omega}F_2=-\int_{Y\setminus \omega}(\lambda(\nabla_y\cdot u)I+2\mu (\frac{1}{2}(\nabla+\nabla^T)u):\nabla_y \varphi\\
    -\int_{\omega}(2\tilde{\mu}[\frac{1}{2}(\nabla_y+\nabla_y^T)u]+pI):\nabla_y \varphi+\int_{\partial\omega}G\cdot N \varphi.
\end{aligned}
\end{equation}

This is correct since the analysis of Babuska-Brezzi Theorem: the following two equation are well-posedness in $V\times M$
\begin{equation}
    \begin{aligned}
        0=\int_{Y\setminus \omega}(\lambda(\nabla_y\cdot u)I+2\mu (\frac{1}{2}(\nabla+\nabla^T)u):\nabla_y \varphi
    +\int_{\omega}(2\tilde{\mu}[\frac{1}{2}(\nabla_y+\nabla_y^T)u]+pI):\nabla_y \varphi \ \forall \varphi\in V,
    \end{aligned}
\end{equation}

and 
\begin{equation}
    \begin{aligned}
        \int_{\omega}F_3\cdot \psi=\int_{\omega} (\nabla_y\cdot u)\psi \quad \forall \psi \in M.
    \end{aligned}
\end{equation}

}

\subsection{Homogenization Equation}
{\color{blue}
For $(u_0,p_{-1})$:
\begin{equation}
    \begin{cases}
        \nabla_y \cdot(\lambda(\nabla_y\cdot u_0+2\mu D_y(u_0))=0\\
        \nabla_y \cdot(2\tilde{\mu}D_y(u_0)+p_{-1}I)=0,\nabla_y \cdot u_0=0\\
        (2\tilde{\mu}D_y(u_0)+p_{-1}I)N^- -(\lambda(\nabla_y \cdot u_0)I+2\mu D_y(u_0))N^+=0 \\
        u_0|_{+}=u_0|_{-} \\
    \end{cases}
\end{equation}
From the derivations in Sec.~\ref{sec:Uniqueness_and_Existence}, we know $u_0(x,y)=u_0(x)$, $p_{-1}(x,y)=p_{-1}(x)=0$.

For $(u_1,p_0)$:
\begin{equation}
    \begin{cases}
         \nabla_y \cdot(\lambda\nabla_y\cdot u_1+2\mu D_y(u_1))=0\\
        \nabla_y \cdot(2\tilde{\mu}D_y(u_1)+p_{0}I)=0,\nabla_y \cdot u_1=-\nabla_x\cdot u_0\\
        (2\tilde{\mu}D_y(u_1)+p_0I)N^- -(\lambda(\nabla_y \cdot u_1)I+2\mu D_y(u_1))N^+\\
        \quad=-2\tilde{\mu}D_x(u_0)N^-+ (\lambda\nabla_x\cdot u_0 I+2\mu D_x (u_0))N^+ \\

        u_1|_{+}=u_1|_{-} \\
       
    \end{cases}
\end{equation}

\subsection{Cell Problem}
Suppose $\chi^{ij}$ and $r^{ij}$ satisfy the following equation:
\begin{equation}
    \begin{cases}
        \nabla_y \cdot(\lambda\nabla_y\cdot \chi^{ij}+2\mu D_y\chi^{ij})=0 \quad &\text{in} \quad Y\setminus \omega,\\
        
        \nabla_y \cdot(2\tilde{\mu}D_y \chi^{ij}+r^{ij}I)=0,\nabla_y \cdot \chi^{ij}=-\delta_{ij}\quad &\text{in} \quad \omega,\\
        (2\tilde{\mu}D_y(\chi^{ij})+r^{ij}I)N^- -(\lambda(\nabla_y \cdot \chi^{ij})I+2\mu D_y(\chi^{ij}))N^+\\
        \quad =-\tilde{\mu}(E_{ij}+E_{ji})N^-+\mu (E_{ij}+E_{ji})N^++\lambda IN\delta_{ij}|_+ \\
       \chi^{ij}|_{+}=\chi^{ij}|_{-} \quad &\text{on} \quad \partial\omega.
    \end{cases}
\end{equation}

From the definition above, we have $\chi^{ij}=\chi^{ji}$, $r^{ij}=r^{ji}$, 
 $$u_1=(D_x u_0)^{ij}\cdot \chi^{ij},\ p_0=(D_xu_0)^{ij}r^{ij}.$$

So for $(u_2,p_1)$, we have:
\begin{equation}
\begin{cases}
    \nabla_y\cdot(\lambda\nabla_y\cdot u_2+2\mu D_yu_2)=-\nabla_x\cdot[(\lambda(\nabla_x u_0)I+2\mu D_x)+(\lambda(\nabla_y \cdot u_1)I+2\mu D_y u_1)]\\
    \quad\quad\quad\quad\quad\quad\quad\quad\quad\quad\quad\quad\quad\quad\quad -\nabla_y\cdot(\lambda(\nabla_x\cdot u_1)I+2\mu D_xu_1) \\
    \nabla_y\cdot(2\tilde{\mu}D_y u_2+p_1I)=-\nabla_x\cdot(2\tilde{\mu}D_x u_0+p_0 I)-\nabla_x\cdot(2\tilde{\mu}(D_y u_1))-\nabla_y \cdot (2\tilde{\mu}(D_x u_1))\\

    \nabla\cdot u_2=-\nabla_x \cdot u_1\\
    u_2|_{-}=u_2|_{+}\\
    % (p_1+2\tilde{\mu}(D_yu_2))N^--(\lambda(\nabla_y\cdot u_2)+2\mu(D_yu_2))N^+=-(2\tilde{\mu }D_x(u_1))N^-\\
    % \quad +(\lambda(\nabla_x \cdot u_1)+2\mu(D_xu_1))N^+
    (p_1+2\tilde{\mu}(D_yu_2))N^--(\lambda(\nabla_y\cdot u_2)+2\mu(D_yu_2))N^+\\
    \quad\quad\quad\quad\quad\quad\quad\quad\quad\quad=-(2\tilde{\mu }D_x(u_1))N^- + (\lambda(\nabla_x \cdot u_1)+2\mu(D_xu_1))N^+
\end{cases}
\end{equation}
By divergence theorem, we can get
\begin{equation}
    \int_{\partial\omega} (\lambda(\nabla_x \cdot u_1)+2\mu(D_xu_1))N^+= -\int_{Y\setminus\omega}\nabla_y \cdot(\lambda(\nabla_x \cdot u_1)+2\mu(D_xu_1)).     
\end{equation}
\begin{equation}
    \int_{\partial\omega}(2\tilde{\mu }D_x(u_1))N^-=\int_{\omega}\nabla_y\cdot(2\tilde{\mu }D_x(u_1)).
\end{equation}

Combining the solvability conditions, we have

\begin{equation}
\begin{aligned}
    \vert  Y\setminus\omega\vert\nabla_x\cdot[\lambda(\nabla_x\cdot u_0)I+2\mu D_x(u_0)]+\int_{Y\setminus\omega}\lambda(\nabla_x(\nabla_y\cdot u_1))+2\mu(\nabla_x\cdot D_y(u_1))\diff y\\
+\vert \omega\vert(2\tilde{\mu }\nabla_x\cdot D_x u_0)+\int_{\omega} (2\tilde{\mu}(\nabla_x\cdot D_y u_1)+\nabla_x p_0)\diff y=0.
\end{aligned}
\end{equation}
Substituting this equation $u_1=(D_x u_0)^{ij}\cdot \chi^{ij}$, $p_0=(D_xu_0)^{ij}r^{ij}$, we finally get
\begin{equation}\label{eq:homogenization}
    \begin{aligned}
        \vert  Y\setminus\omega\vert\nabla_x\cdot[\lambda(\nabla_x\cdot u_0)I+2\mu D_x(u_0)]+\int_{Y\setminus\omega}[\lambda(\nabla_y\cdot \chi^{ij})I+2\mu D_y(\chi^{ij})](\nabla_x(D_xu_0)^{ij}) \diff y\\
+\vert \omega\vert(2\tilde{\mu }\nabla_x\cdot D_x u_0)+\int_{\omega} (2\tilde{\mu}(D_y \chi^{ij})+r^{ij}I)\nabla_x(D_Xu_0)^{ij} \diff y =0.
\end{aligned}
\end{equation}

We can rewrite this equation as the form $-\frac{\partial}{\partial x_i}\hat{a}_{ij}^{\alpha \beta}\frac{\partial u^\beta}{\partial x_j}=0$:
\begin{equation}
\begin{aligned}
    \hat{a}_{ij}^{\alpha \beta}=\vert  Y\setminus\omega\vert(\lambda \delta_{i\alpha}\delta_{j\beta}+\mu (\delta_{i\beta}\delta_{j\alpha}+\delta_{ij}\delta_{\alpha\beta}))+ \int_{Y\setminus\omega}[\lambda \nabla\cdot \chi^{j\beta}\delta_{i\alpha}+\mu (\frac{\partial \chi^{j\beta}_{\alpha}}{\partial y_i}+\frac{\partial \chi^{j\beta}_i}{\partial y_\alpha})]\diff y\\
    +\vert \omega \vert\tilde{\mu}(\delta_{ij}\delta_{\alpha\beta}+\delta_{i\beta}\delta_{j\alpha})+\int_{\omega} \tilde{\mu }(\frac{\partial \chi^{j\beta}_{\alpha}}{\partial y_i}+\frac{\partial \chi^{j\beta}_i}{\partial y_\alpha})+r^{j\beta}\delta_{i\alpha}\diff y.
\end{aligned}
\end{equation}
}
\subsubsection{Symmetry}
{\color{orange}
We first prove the symmetry of $\hat{a}_{ij}^{\alpha \beta}$.

Define: $a(\varphi,\psi)=\int_{Y\setminus\omega}\nabla_y\varphi:[\lambda(\nabla_y\cdot\psi)I+2\mu D(\psi)]+\int_{\omega}2\tilde{\mu}D\psi:D\psi= \int_{Y\setminus\omega}\lambda(\nabla_y\cdot\varphi)( \nabla_y\cdot\varphi)+\mu D(\psi):D(\psi)+\int_{\omega}2\tilde{\mu}D\psi:D\psi$.

Claim $\hat{a}_{ij}^{\alpha \beta}=a(p^{i\alpha}+\chi^{i\alpha},p^{j\beta}+\chi^{j\beta}) $,
\begin{align}
    a(p^{i\alpha},p^{j\beta}) &= \int_{Y\setminus\omega} \lambda \delta_{i\alpha}\delta_{j\beta}+\mu (\delta_{i\beta}\delta_{j\alpha}+\delta_{ij}\delta_{\alpha\beta}))+\int_{\omega}\tilde{\mu}(\delta_{ij}\delta_{\alpha\beta}+\delta_{i\beta}\delta_{j\alpha}), \label{eq:1} \\
    a(p^{i\alpha},\chi^{j\beta}) &= \int_{Y\setminus \omega}\lambda \delta_{i\alpha}(\nabla\cdot\chi^{j\beta})+\mu\left( \frac{e_{i\alpha}+e_{\alpha i}}{2}\right):\left(\nabla \chi^{j\beta}+\nabla^T\chi^{\beta j}\right) \notag \\
    &\quad +\int_{\omega}\tilde{\mu}\left( \frac{e_{i\alpha}+e_{\alpha i}}{2}\right):\left(\nabla \chi^{j\beta}+\nabla^T\chi^{\beta j}\right) \notag \\
    &= \int_{\partial \omega}-\lambda \delta_{i\alpha}\chi^{j\beta}N-\mu (e_{i\alpha}+e_{\alpha i})\chi^{j\beta}N+\int_{\partial\omega}\tilde{\mu}(e_{i\alpha}+e_{\alpha i})\chi^{j\beta}N, \label{eq:2} \\
    a(\chi^{i\alpha},\chi^{j\beta}) &= \int_{Y\setminus \omega}\lambda (\nabla\cdot\chi^{i\alpha})(\nabla\cdot\chi^{j\beta})+\mu(\nabla_y \chi^{i\alpha}):\left(\nabla \chi^{j\beta}+\nabla^T\chi^{j\beta}\right) \notag \\
    &\quad +\int_{\omega} \tilde{\mu}(\nabla_y\chi^{i\alpha}):\left(\nabla\chi^{j\beta}+\nabla^T\chi^{j\beta}\right) \notag \\
    &= \int_{\partial\omega}-\left[\lambda\chi^{i\alpha}(\nabla\cdot\chi^{j\beta})+2\mu \chi^{i\alpha}D(\chi^{j\beta})\right]N+\int_{\partial\omega}2\tilde{\mu}\chi^{i\alpha}(D\chi^{j\beta})N \notag \\
    &\quad +\int_{\partial\omega}\chi^{i\alpha}r^{j\beta}N+\int_{\omega}\delta_{i\alpha}r^{j\beta}. \label{eq:3}
\end{align}

Observing that $a(p^{i\alpha},p^{j\beta}+\chi^{j\beta})= \hat{a}_{ij}^{\alpha \beta}-\int_{\omega}\delta_{i\alpha}r^{j\beta}\diff y$ and  $a(\chi^{i\alpha},p^{j\beta}+\chi^{j\beta})=\int_{\omega}\delta_{i\alpha}r^{j\beta}\diff y$, we have $\hat{a}_{ij}^{\alpha \beta}=a(p^{i\alpha}+\chi^{i\alpha},p^{j\beta}+\chi^{j\beta})$ and $\hat{a}_{ij}^{\alpha \beta}=\hat{a}_{i\beta}^{\alpha j}=\hat{a}_{ji}^{\beta\alpha}$.
}
\subsubsection{Elliptic}
{\color{blue}Let $\varphi=\phi +\varepsilon(D_x\phi)^{ij}\chi^{ij}(\frac{x}{\varepsilon}),\phi\in C_0^{\infty}(\Omega;R^d) $}
{\color{orange}
\begin{equation}
\begin{aligned}
    \int_{\Omega} a(\varphi,\varphi)&=-\int_{\Omega}\int_{Y\setminus\omega}(\nabla_x\phi+(D_x\phi)^{ij}\nabla_y\chi^{ij}):[\lambda(\nabla_x\cdot\phi
    +(D_x\phi)^{ij}\nabla_y\cdot\chi^{ij})I\\
    &\quad\quad+2\mu (D_x\phi+(D_x\phi)^{ij}D_y(\chi^{ij}))]\\
    &\quad\quad-\int_{\Omega}2\tilde{\mu}\int_{\omega}(\nabla_x\phi+(D_x\phi)^{ij}\nabla_y\chi^{ij}):(D_x\phi+(D_x\phi)^{ij}(D_y\chi^{ij}))\\
    &=\int_{\Omega}[\int_{Y\setminus \omega}\nabla_x\cdot(\lambda(\nabla\cdot\phi)I+2\mu D_x\phi)+(\lambda(\nabla_y\cdot\chi^{ij})I+2\mu(D_y\chi^{ij}))(\nabla_x(D_x\phi)^{ij})\diff y\\ 
    &\quad\quad+\int_{\omega}2\tilde{\mu}\nabla_x \cdot D_x\phi+(2\tilde{\mu}(D_y\chi^{ij})+r^{ij}I)\nabla_x(D_x\phi)^{ij}\diff y ]\\
    &=\int_{R^d}\hat{A}\nabla \phi:\nabla\phi.
\end{aligned}
\end{equation}
So we have
\begin{equation}
    \frac{\min\{\lambda d+2\mu,2\mu,\tilde{\mu}\}}{2}\int_{R^d}\vert\nabla \phi\vert^2\,\diff x \le \liminf_{\varepsilon\rightarrow 0}\frac{\min\{\lambda d+2\mu,2\mu,\tilde{\mu}\}}{2}\int_{R^d}\vert\nabla \varphi\vert^2\,\diff x\\
    \le \int_{R^d} \hat{A}\nabla\phi\cdot\nabla\phi\diff x.
\end{equation}
Finally we can get$$\frac{\min\{\lambda d+2\mu,2\mu,\tilde{\mu}\}}{2}\vert \xi \vert^2\vert \eta\vert^2 \le\hat{a}_{ij}^{\alpha \beta}\xi_i\xi_j\eta^\alpha\eta^\beta.$$
}

\pagebreak
\subsection{Regularity}
{\color{blue}
For each $i,j=1,\cdots,d$, suppose $p^{ij}=\frac{1}{2}(y_j e_i+y_ie_j) $, $\chi^{ij}(y)$ and $r^{ij}(y)$ satisfy the following equation:
\begin{equation*}
\begin{cases}
\begin{aligned}
&\nabla \cdot \left[ \lambda \nabla_y \cdot \chi^{ij} I + 2\mu D_y u \right] = 0 
  &&\text{in } Y \setminus \omega, \\
&\nabla \cdot \left[ r^{ij} I + 2\tilde{\mu} D_y \chi^{ij} \right] = 0 
   &&\text{in } \omega, \\
&\nabla_y \cdot \chi^{ij} = -\delta_{ij} 
   &&\text{in } \omega, \\
&\chi^{ij}|_{+} = \chi^{ij}|_{-}, \\
&\left. \left[ r^{ij} I + 2\tilde{\mu} D_y \chi^{ij} \right] N \right|_{-} 
  - \left. \left[ \lambda \nabla_y \cdot \chi^{ij} I + 2\mu D_y \chi^{ij} \right] N \right|_{+}\\ 
  &\quad = -\tilde{\mu}(E_{ij}+E_{ji})N|_-+\mu(E_{ij}+E_{ji})N|_++\lambda IN\delta_{ij}|_+
   &&\text{on } \partial \omega.
\end{aligned}
\end{cases}
\end{equation*}

In $Y\setminus\omega$ and $\omega$, by interior estimation for $V\subset\subset Y\setminus\omega,W\subset\subset \omega$, we have $ \|\chi\|_{H^k(V)}<\infty$, $\|\chi\|_{H^k(W)}<\infty$. So we can consider the following boundary regularity problem on $\partial \omega$.

Consider:
\begin{equation*}
\begin{cases}
\begin{aligned}
&\lambda \Delta_y(\chi^{ij}+p^{ij})+(\lambda+\mu) \nabla_y (\nabla_y\cdot (\chi^{ij}+p^{ij}))= 0 
  &&\text{in } Y \setminus \omega, \\
&\tilde{\mu}\Delta_y(\chi^{ij}+p^{ij}) +\nabla_yr^{ij} = 0 
   &&\text{in } \omega, \\
&\nabla_y \cdot (\chi^{ij}+p^{ij}) =0
   &&\text{in } \omega, \\
&\chi^{ij}|_{+} = \chi^{ij}|_{-}, \\
&\left. \left[ r^{ij} I + 2\tilde{\mu} D_y (\chi^{ij}+p^{ij}) \right] N \right|_{-} 
  - \left. \left[ \lambda \nabla_y \cdot (\chi^{ij}+p^{ij}) I + 2\mu D_y(\chi^{ij}+p^{ij}) \right] N \right|_{+}=0
   &&\text{on } \partial \omega.
\end{aligned}
\end{cases}
\end{equation*}

This is equivalent to the following problem:

Suppose that there is a sphere \(Q_r=(-r,r)^d\) with radius \(r\) centered at the origin in \(\mathbb{R}^d\). 

We define 
\[
Q_r^{+} = \{(x', x_d) \in Q_r : x_d > \psi(x')\}, \quad Q_r^{-} = \{(x', x_d) \in Q_r : x_d <\psi(x')\}, \quad \]
\[I_r = \{(x', x_d) \in Q_r : x_d = \psi(x')\}.
\]
Here $\psi:R^{d-1}\rightarrow R$ is a $C^{\infty}$ function and $\psi(0)=0$,

\begin{equation*}
\begin{cases}
\begin{aligned}
&\nabla_y \cdot \left[ \lambda \nabla_y \cdot \chi^{ij} \, I + 2\mu D_y\chi^{ij}\right] = 0 \quad &&\text{in } Q_r^{+}, \\
&\nabla \cdot \left[ r^{ij} \, I + 2\tilde{\mu} D_y \chi^{ij} \right] = 0 \quad &&\text{in } Q_r^{-}, \\
&\nabla \cdot \chi^{ij} = 0 \quad &&\text{in } Q_r^{-}, \\
&\chi^{ij}|_{+} = \chi^{ij}|_{-}, \\
&\left. \left[ r^{ij} \, I + 2\tilde{\mu} D_y\chi^{ij} \right] N \right|_{-} - \left. \left[ \lambda \nabla_y \cdot \chi^{ij} \, I + 2\mu D_y\chi^{ij}\right] N \right|_{+} = 0 \quad &&\text{on }I_r.
\end{aligned}
\end{cases}
\end{equation*}
 \(\chi^{ij},r^{ij}\) are the weak solution in \(H^1(Q_r; \mathbb{R}^d)\) and \( L^2(Q_r^-; \mathbb{R})\).}

\subsubsection{A Basic \texorpdfstring{$C^{1,\alpha}$}{C1,alpha} Estimate}
{\color{orange}
Suppose \(\Omega_{\pm} = \mathbb{R}^d_{\pm}\), \(S = \{x_d = 0\}\), \(B_+ = \{x \in B(1) : x_d > 0\}\), and \(B_- = \{x \in B(1) : x_d < 0\}\), where \(B(1) = \{x \in \mathbb{R}^d : \|x\| \le 1\}\).

Consider the following equation: for all \(V \in H_0^1(B(1); \mathbb{R}^d)\),
\begin{equation*}
    \begin{cases}
        \begin{aligned}
            &(\nabla V : A_1 \nabla \tilde{\chi})_{B_+} + (\nabla V : A_2 \nabla \tilde{\chi})_{B_-} + (\tilde{r}, \nabla \cdot (aV))_{B_-} = 0,  \\
            &\nabla \cdot (a \tilde{\chi}) = 0.
        \end{aligned}
    \end{cases}
\end{equation*}
where \(\tilde{\chi} = D^\alpha \chi\), \(\tilde{r} = D^\alpha r\) with \(|\alpha| \ge 1\), \(A_1\), \(A_2\) are constant tensors, and \(a\) is a constant matrix.

We let $V=\eta^2\tilde{\chi}$ and $\eta $ be smooth with $\eta=0 \ \text{in} \ B(1)^c$ and $\eta =1 \text{ in } B(\frac{1}{2})$.
So we can get
\begin{equation*}
    \begin{aligned}
        \eta^2(\nabla \tilde{\chi} : A_1 \nabla \tilde{\chi})_{B_+} +\eta^2 (\nabla \tilde{\chi} : A_2 \nabla \tilde{\chi})_{B_-} + (\tilde{\chi}(\nabla \eta^2)^T:A_1\nabla\tilde{\chi})_{B_+}\\
        +(\tilde{\chi}(\nabla \eta^2)^T:A_2\nabla\tilde{\chi})_{B_-}+ (\tilde{r}, \nabla \cdot (a\tilde{\chi})\eta^2)_{B_-} +
        (\tilde{r},(\nabla \eta^2)^T \cdot (a\tilde{\chi}))_{B_-} = 0.
    \end{aligned}
\end{equation*}

Therefore,
\begin{equation}
\|\eta\nabla\tilde{\chi}\|^2_{L^2(B(1))}\le C\|\eta \nabla\tilde{\chi}\|_{L^2(B(1))}\|\tilde{\chi}\|_{L^2(B(1))}+C\|\eta\tilde{r}\|_{L^2(B(1))}\|\tilde{\chi}\|_{L^2(B(1))}.
\end{equation}
Besides, we also have $\|\eta\tilde{r}\|_{L^2(B(1))}\le C\sum_{\beta=|\alpha|+1}\|\eta D^{\beta}\chi\|_{L^2(B(1))}=m_{|\alpha|+1}$, and $\sum_{\beta=|\alpha|}\|D^{\beta}\chi\|_{L^2(B(1))}=n_{|\alpha|}$.

So we have $m_{|\alpha|+1}^2\le Cm_{|\alpha|+1}n_{|\alpha|}$.
That is 
\begin{equation}
    \|D^{|\alpha|+1}\chi\|_{L^2(B(\frac{1}{2}))}\le C \|D^{|\alpha|}\chi\|_{L^2(B(1))}.
    \label{eq:inequality1}
\end{equation}

Similarly,
\begin{equation}
    \|D^{|\alpha|}r\|_{L^2(B(\frac{1}{2}))}\le C \|D^{|\alpha|-1}r\|_{L^2(B(\frac{1}{2}))}.
    \label{eq:inequality2}
\end{equation}
}
\subsubsection{Schauder Theory}
Since $\frac{\partial^2\chi}{\partial y_d^2}$ is the linear combination of $\frac{\partial^2\chi}{\partial y_i\partial y_j},\frac{\partial\chi}{\partial y_j},\frac{\partial r}{\partial y_j},r$, the inequalities (Eq.~\ref{eq:inequality1} and Eq.~\ref{eq:inequality2}) above are correct for any $|\alpha_n|=|\alpha|$. So we have proved the Caccioppoli inequality.
\begin{lemma}{(Caccioppoli Inequality)}    For the weak solutions $(\chi ,r)$, $\forall k\ge 1$ 
    \begin{equation}\sum_{\pm}\|\chi\|_{H^k(B(\frac{1}{2},\pm))} \le C\|\chi \|_{L^2(B(1))},
    \end{equation}
    \begin{equation}
    \|r\|_{H^k(B(\frac{1}{2})_-)} \le C\|r \|_{L^2(B(\frac{1}{2})_-)}.
    \end{equation}
    \label{lemma:lemma11}
\end{lemma}

\begin{lemma}
    Suppose that \(M_{±}\) is the constant matrix in \(R^{d \times d}\), the following are equivalent:
    \begin{equation}
        \begin{cases}
        \forall y \in  \{y_d=0\}, \ M_{+}x=M_{-}x,\\
        \exists c\in R^d\ s.t. \ M_{+}-M_{-}=C e_d^T, \\
        (I-e_d^Te_d)M_{+}=(I-e_d^Te_d)M_{-}.
    \end{cases}
    \end{equation}
    \label{lemma:lemma12}
\end{lemma}
\begin{definition}

Let $A_1$, $A_2$ be constant tensors and \(a\) be a constant matrix. Suppose $M_{+}$ and $M_-$ satisfy Lemma~\ref{lemma:lemma12} above and \(\nabla \cdot (a M_{-} y) = 0\) \text{ in } \(B(t)_{-}\). Define
\[
l(y) = M_{+} y \, \mathbf{1}_{y_d \ge 0} + M_{-} y \, \mathbf{1}_{y_d \le 0} + C, \quad q(y) = r(0),
\]
where \(\mathbf{1}_{\cdot}\) denotes the indicator function. We call $l$, $q$ the piecewise linear solutions of the following equations:
\begin{equation*}
    \begin{cases}
        \begin{aligned}
        &\nabla \cdot (A_1 \nabla l) = 0 \quad &&\text{in } \mathbb{R}^d_{+}, \\
        &\nabla \cdot (A_2 \nabla l) + \nabla\cdot(a^T q) = 0 \quad &&\text{in } \mathbb{R}^d_{-}, \\
        &\nabla \cdot (a l) = 0 \quad &&\text{in } \mathbb{R}^d_{-}, \\
        &l|_{+} = l|_{-} \quad &&\text{on } \{x_d = 0\}, \\
        &\left. \frac{\partial l}{\partial \nu} \right|_{+} - \left. \frac{\partial l}{\partial \nu} \right|_{-} = (A_1 M_{+}) e_d - (A_2 M_{-} + a^Tr(0)) e_d \quad &&\text{on } \{x_d = 0\}.
        \end{aligned}
    \end{cases}
\end{equation*}

where \(e_d\) is the standard basis vector in \(\mathbb{R}^d\).

Let \(\mathcal{L}\) be the space of all piecewise linear solutions of the above equations. For any \((l, q) \in \mathcal{L}\), define
\[
\zeta(l, q) = \left( \frac{\partial l}{\partial \nu} \right)_{+} - \left( \frac{\partial l}{\partial \nu} \right)_{-}.
\]
\end{definition}
\begin{lemma}
  
Let $A_1$, $A_2$ be constant tensors and \(a\) be a constant matrix. Consider the following system:
\begin{equation*}
    \begin{cases}
        \begin{aligned}
            
&\nabla \cdot (A_1 \nabla \chi) = 0 \quad &&\text{in } B(1)_+, \\
&\nabla \cdot (A_2 \nabla \chi) + \nabla\cdot(a^T r)  = 0 \quad &&\text{in } B(1)_-, \\
&\nabla \cdot (a \chi) = 0 \quad &&\text{in } B(1)_-, \\
&\chi|_+ = \chi|_- \quad &&\text{on } B(1) \cap \{x_d = 0\}, \\
&\left. \frac{\partial \chi}{\partial \nu} \right|_+ - \left. \frac{\partial \chi}{\partial \nu} \right|_- = g_0 \quad &&\text{on } B(1) \cap \{x_d = 0\}.
        \end{aligned}
    \end{cases}
\end{equation*}
where \(B(1) = \{x \in \mathbb{R}^d : \|x\| \le 1\}\), \(B(1)_+ = B(1) \cap \mathbb{R}^d_+\), \(B(1)_- = B(1) \cap \mathbb{R}^d_-\), and \(\mathbb{R}^d_\pm = \{x \in \mathbb{R}^d : \pm x_d > 0\}\).

Let $\chi$, $r$ be weak solutions of the above system. Then for all \(k \ge 0\) and \(\alpha \in [0, 1]\), we have
\begin{equation}
\sum_{\pm} \|\chi\|_{H^k(B(\frac{1}{2})_\pm)} \le C \left( \|\chi\|_{L^2(B(1))} + |g_0| \right),
\end{equation}
where \(C\) is a constant independent of $\chi$, $r$, and \(B(\frac{1}{2}) = \{x \in \mathbb{R}^d : \|x\| \le \frac{1}{2}\}\).
\label{lemma:lemma13}
\end{lemma}

\begin{lemma}
Let \(A_1, A_2\) be constant tensors and \(a\) be a constant matrix. Consider the system:
\begin{equation*}
    \begin{cases}
        \begin{aligned}
            
&\nabla \cdot (A_1 \nabla \chi) = 0 \quad &&\text{in } B(1)_+, \\
&\nabla \cdot (A_2 \nabla \chi) + \nabla \cdot(a^Tr) = 0 \quad &&\text{in } B(1)_-, \\
&\nabla \cdot (a \chi) = 0 \quad &&\text{in } B(1)_-, \\
&\chi|_+ = \chi|_- \quad &&\text{on } B(1) \cap \{x_d = 0\}, \\
&\left. \frac{\partial \chi}{\partial \nu} \right|_+ - \left. \frac{\partial \chi}{\partial \nu} \right|_- = g_0 \quad &&\text{on } B(1) \cap \{x_d = 0\}.
        \end{aligned}
    \end{cases}
\end{equation*}

where $\chi$, $r$ are weak solutions.

Define
\[
l(y) = (\nabla\chi)_+(0) \, y \, \mathbf{1}_{y_d \ge 0} + (\nabla\chi)_-(0) \, y \, \mathbf{1}_{y_d \le 0} + \chi(0), \quad q(y) = r(0),
\]
where \(\mathbf{1}_{\cdot}\) denotes the indicator function. By Lemma~\ref{lemma:lemma11}, we know \(\chi(0),(\nabla \chi(x))_{\pm}(0)\) are well-defined.
\[
(I - e_d \otimes e_d)(\nabla \chi)_+ = (I - e_d \otimes e_d)(\nabla \chi)_- \quad \text{on } B(1) \cap \{y_d = 0\},
\]
where \(e_d\) is the standard basis vector in \(\mathbb{R}^d\) and \(I\) is the identity matrix. From Lemma~\ref{lemma:lemma12}, it follows that \((l, q) \in \mathcal{L}\).

For some \(\beta \in (0, 1)\) and all \(y \in B\left(\frac{1}{2}\right)\), we have:
\begin{align*}
|\chi(y) - l(y)| &= \left| \chi(y) - \chi(0) - (\nabla \chi)_{\pm}(0) \, y \right| \\
&\le C |y|^{\beta + 1} \left( [\chi]_{C^{1, \beta}(B\left(\frac{1}{2}\right)_{\pm})} \right) \\
&\le C |y|^{\beta + 1} \left( \left( \int_{B(1)} |\chi|^2 \right)^{\frac{1}{2}} + |g_0| \right).
\end{align*}

For all \(y \in B\left(\frac{1}{2}\right)_-\):
\begin{align*}
|r - q| &\le C |y|^{\beta} \left( [r]_{C^{0, \beta}(B\left(\frac{t}{2}\right)_-)} \right) \\
&\le C |y|^{\beta} \left( \int_{B(1)_-} |r|^2 \right)^{\frac{1}{2}}.
\end{align*}

Therefore, for some \(\beta \in (0, 1)\) and all \(y \in B\left(\frac{t}{2}\right)\):
\begin{align*}
|\chi(y) - l(y)| &= \left| \chi(y) - \chi(0) - (\nabla \chi)_{\pm}(0) \, y \right| \\
&\le C \left| \frac{y}{t} \right|^{\beta + 1} \left( [\chi]_{C^{1, \beta}(B\left(\frac{t}{2}\right)_{\pm})} \right) \\
&\le C \left| \frac{y}{t} \right|^{\beta + 1} \left( \left( \fint_{B(t)} |\chi|^2 \right)^{\frac{1}{2}} + t |g_0| \right),
\end{align*}
where \(\fint\) denotes the average integral.

For all \(y \in B\left(\frac{t}{2}\right)_-\):
\begin{align*}
|r - q| &\le C \left| \frac{y}{t} \right|^{\beta} \left( [r]_{C^{0, \beta}(B\left(\frac{t}{2}\right)_-)} \right) \\
&\le C \left| \frac{y}{t} \right|^{\beta} \left( \fint_{B(t)_-} |r|^2 \right)^{\frac{1}{2}}.
\end{align*}
\end{lemma}

\begin{lemma}
Let $A_1$, $A_2$ be constant tensors and \(a\) be a constant matrix. Consider the system:
\begin{equation*}
    \begin{cases}
        \begin{aligned}
            &\nabla \cdot (A_1 \nabla \chi) = 0 \quad &&\text{in } B(1)_+, \\
&\nabla \cdot (A_2 \nabla \chi) + \nabla\cdot(a^T r) = 0 \quad &&\text{in } B(1)_-, \\
&\nabla \cdot (a \chi) = 0 \quad &&\text{in } B(1)_-, \\
&\chi|_+ = \chi|_- \quad &&\text{on } B(1) \cap \{x_d = 0\}, \\
&\left. \frac{\partial \chi}{\partial \nu} \right|_+ - \left. \frac{\partial \chi}{\partial \nu} \right|_- = g_0 \quad &&\text{on } B(1) \cap \{x_d = 0\}.
        \end{aligned}
    \end{cases}
\end{equation*}
where $\chi$, $r$ are weak solutions.

Moreover, for all \(\rho \in (0, t)\), integrating the above inequalities yields:
\begin{equation}
\left( \fint_{B(\rho)} |\chi - l|^2 \right)^{\frac{1}{2}} + \rho \left| g_0 - \zeta(l, q) \right| \le \left| \frac{\rho}{t} \right|^{\beta + 1} \left( \left( \fint_{B(t)} |\chi|^2 \right)^{\frac{1}{2}} + t |g_0| \right).
\end{equation}

Thus, for all \((l', q') \in \mathcal{L}\), by the inequality above, we have:
\begin{equation}
\begin{aligned}
&\inf_{\substack{(l,q) \in \mathcal{L} }} \left\{ \left( \fint_{B(\rho)} |\chi - l|^2 \right)^{\frac{1}{2}} + \rho \left| g_0 - \zeta(l, q) \right| \right\} \\
&\le C \left| \frac{\rho}{t} \right|^{\beta + 1} \inf_{\substack{(l,q) \in \mathcal{L} }} \left( \left( \fint_{B(t)} |\chi - l|^2 \right)^{\frac{1}{2}} + t |g_0- \zeta(l, q)| \right).
\end{aligned}
\end{equation}

Further, for all \((l', q') \in \mathcal{L}\), it follows that:
\begin{equation}
\begin{aligned}
&\inf_{\substack{(l,q) \in \mathcal{L} }} \left\{ \left( \fint_{B(\rho)} |\chi - l|^2 \right)^{\frac{1}{2}} + \rho \left| g_0 - \zeta(l, q) \right| \right\} \\
&\le C \left| \frac{\rho}{t} \right|^{\beta + 1} \inf_{\substack{(l,q) \in \mathcal{L} }} \left( \left( \fint_{B(t)} |\chi - l|^2 \right)^{\frac{1}{2}} + t \left| g_0 - \zeta(l, q) \right| \right).
\end{aligned}
\end{equation}
\label{lemma:lemma15}
\end{lemma}

\begin{lemma}
    Suppose \(\phi:R_{+}\rightarrow R_{+}\) is a non-decreasing non-negative function satisfying \(\phi(\rho )\le C(\frac{\rho}{t})^{\beta} \phi(r)+Br^{\alpha}\), where \(\beta>\alpha>0,C>0\). $\forall0<\rho<r<R$, $\exists C_1$, s.t. $\phi(\rho)\le C_1 (\frac{\rho}{t})^{\alpha} \phi(r)+B\rho^{\alpha}$.
    \label{lemma:lemma16}
\end{lemma}

{\color{Lgreen}
\begin{theorem}

Suppose $A_1$, $A_2$, $a$ are \(C^\alpha\)-Holder continuous, and let \(S_t = B(t) \cap \{x_d = 0\}\) where \(B(t) = \{x \in \mathbb{R}^d : \|x\| \le t\}\). Let $\chi$, $r$ be weak solutions to the following system: \(\forall V \in H_0^1(B(1); \mathbb{R}^d)\),
\begin{equation*}
    \begin{cases}
        \begin{aligned}
            
&(\nabla V : A_1 \nabla \chi)_{B_+} + (\nabla V : A_2 \nabla \chi)_{B_-} + (r, \nabla \cdot (a \chi))_{B_-} = 0 \\
&\nabla \cdot (a \chi) = 0
        \end{aligned}
    \end{cases}
\end{equation*}
where \(B_+ = B(1) \cap \mathbb{R}^d_+\), \(B_- = B(1) \cap \mathbb{R}^d_-\), and \(\mathbb{R}^d_\pm = \{x \in \mathbb{R}^d : \pm x_d > 0\}\).

Then we have:
\begin{equation}
\sum_{\pm} \|\chi\|_{C^{1,\alpha}(B\left(\frac{1}{2}\right)_{\pm})} \le C \|\chi\|_{L^2(B(1))},
\end{equation}
where \(C\) is a constant independent of \(\chi\), and \(C^{1,\alpha}(\Omega)\) denotes the Holder space of functions with \(\alpha\)-Holder continuous first derivatives.

Equivalent Formulation:
This estimate is equivalent to showing that  \(\forall \rho \in (0, \frac{1}{4})\),
\[
\inf_{\substack{(l,q) \in \mathcal{L} }} \left( \fint_{B(\rho)} |\chi - l|^2 \right)^{\frac{1}{2}} \le C \rho^{1+\alpha} \sum_{\pm} \|\chi\|_{L^2(B\left(\frac{3}{4}\right))},
\]
here \(\mathcal{L}\) is the space of piecewise linear functions \((l, q)\) as defined earlier, \(B(\rho)\) is a small ball with center arbitrary on \(S_{\frac{3}{4}} = B\left(\frac{3}{4}\right) \cap \{x_d = 0\}\) and \(\fint\) denotes the average integral over the domain.

\end{theorem}
\begin{proof}[Disturbance]

Suppose \(w_t, s_t\) are solutions to the following equations, where \(A_1^0 = A_1(0)\), \(A_2^0 = A_2(0)\), \(a^0 = a(0)\): \(\forall V \in H_0^1(B(t); \mathbb{R}^d)\),
\begin{equation*}
    \begin{cases}
        \begin{aligned}
            &(\nabla V : A_1^0 \nabla w_t)_{B_+} + (\nabla V : A_2^0 \nabla w_t)_{B_-} + \left( s_t, \nabla \cdot (a^0 V) \right)_{B_-} = 0, \\
&\nabla \cdot (a^0 w_t) = 0 \quad &&\text{in } B_-, \\
&w_t = \chi \quad &&\text{on } \partial B(t), \\
&s_t = r \quad &&\text{on } \partial B(t)_-.
        \end{aligned}
    \end{cases}
\end{equation*}
where \(B_+ = B(t) \cap \mathbb{R}^d_+\), \(B_- = B(t) \cap \mathbb{R}^d_-\), and \(B(t) = \{x \in \mathbb{R}^d : \|x\| \le t\}\).

By the continuity result in Lemma~\ref{lemma:lemma13}, we have \(\forall \rho \in (0, t)\):
\begin{align*}
&\left( \fint_{B(\rho)} |\nabla w_t|^2 \right)^{\frac{1}{2}} + \left( \fint_{B(\rho)_-} |s_t|^2 \right)^{\frac{1}{2}} \\
&\le C \left( \left( \fint_{B(t)} |\nabla w_t|^2 \right)^{\frac{1}{2}} + \left( \fint_{B(t)_-} |s_t|^2 \right)^{\frac{1}{2}} \right),
\end{align*}
where \(C\) is a constant independent of $t$, $\rho$.

Moreover, from the equations, we derive:
\begin{align*}
\left( \nabla \phi, A^0 \nabla (w_t - \chi) \right)_{B(t)} &= \left( \nabla \phi, (A - A^0) \nabla \chi \right)_{B(t)} \\
&\quad + \left( r, \nabla \cdot (a^0 \phi) \right)_{B(t)_-} - \left( s_t, \nabla \cdot (a^0 \phi) \right)_{B(t)_-},
\end{align*}
where \(A^0\) denotes the piecewise constant tensor \(A_1^0\) on \(B_+\) and \(A_2^0\) on \(B_-\).

Let \(\phi = w_t - \chi\). Substituting into the above equation, we obtain:
\begin{equation}
\begin{aligned}
&\Lambda^{-1} \|\nabla w_t - \nabla \chi\|_{L^2(B(t))}^2 \le \int_{B(t)} |A - A^0| \, |\nabla \chi| \, |\nabla w_t - \nabla \chi| \\
&\quad + \int_{B(t)_-} (r - s_t) \cdot \left( a^0 (-\nabla \chi) \right) + \int_{B(t)_-} r (a - a^0) \nabla (w_t - \chi) \\
&\le \int_{B(t)} |A - A^0| \, |\nabla \chi| \, |\nabla w_t - \nabla \chi| + \int_{B(t)_-} |r - s_t| \, |a^0| \, |\nabla \chi| \\
&\quad + \int_{B(t)_-} |r| \, |a - a^0| \, |\nabla w_t - \nabla \chi|,
\end{aligned}
\end{equation}
where \(\Lambda > 0\) is the ellipticity constant of \(A^0\).

By Young's inequality, this implies:
\begin{equation}
\begin{aligned}
\int_{B(t)} |\nabla w_t - \nabla \chi|^2 + \int_{B(t)_-} |r - s_t|^2 \le C \left[ t^{2\alpha} \int_{B(t)} |\nabla \chi|^2 + t^{2\alpha} \int_{B(t)_-} |r|^2 \right],
\end{aligned}
\end{equation}
where \(C\) depends on \(\Lambda\), \(\|A\|_{C^\alpha}\),\(\|a\|_{C^\alpha}\), and \(\alpha\) is the Holder exponent.
\end{proof}
\begin{proof}[Morrey's Estimate and Bootstrap Analysis]

By the above analysis, we obtain the following result:
\begin{equation}
\begin{aligned}
&\int_{B(\rho)} |\chi|^2 + \int_{B(\rho)_-} |r|^2 \\
&\le C \left( \int_{B(\rho)} |\nabla \chi - \nabla w_t|^2 + \int_{B(\rho)} |\nabla w_t|^2 + \int_{B(\rho)_-} |r - s_t|^2 + \int_{B(\rho)_-} |s_t|^2 \right) \\
&\le C \left( \left( \frac{\rho}{t} \right)^d \left( \int_{B(t)} |\nabla \chi|^2 + \int_{B(t)_-} |r|^2 \right) + t^{2\alpha} \left( \int_{B(t)} |\nabla \chi|^2 + \int_{B(t)_-} |r|^2 \right) \right).
\end{aligned}
\end{equation}

Define \(\Psi(r) = \int_{B(r)} |\nabla \chi|^2 + \int_{B(r)_-} |r|^2\). Then we have:
\[
\Psi(\rho) \le C \left( \left( \frac{\rho}{t} \right)^d \Psi(t) + t^{2\alpha} \Psi(t) \right) \quad \forall \, 0 < \rho < t < \frac{1}{2}.
\]

By Lemma~\ref{lemma:lemma16}, it follows that:
\begin{equation}
    \begin{aligned}
\Psi(\rho) \le C \rho^{2\alpha} \Psi\left( \frac{1}{2} \right) \quad \forall \, \rho \in \left( 0, \frac{1}{2} \right).
    \end{aligned}
\end{equation}

Thus, we derive:
\begin{equation}
\begin{aligned}
\Psi(\rho) \le C \left( \left( \frac{\rho}{t} \right)^d \Psi(t) + t^{4\alpha} \Psi(t) \right).
\end{aligned}
\end{equation}

By bootstrap analysis, for all \(r < d\) and \(0 < \rho < \frac{1}{2}\), we have:
\begin{equation}
\begin{aligned}
\Psi(\rho) \le C \rho^r \Psi\left( \frac{1}{2} \right).
\end{aligned}
\end{equation}

The above estimate holds for any ball with center in \(B\left( \frac{1}{2} \right)\) and radius \(t\).

Finally, we conclude that \(|\nabla \chi| \in L^{2, r}(B\left( \frac{1}{2} \right))\), which implies \(\chi\) is \(C^\beta\)-Holder continuous for all \(\beta \in (0, 1)\).
\end{proof}
}

{\color{orange}

\begin{proof}[\(C^{1,\alpha}\)Continuity]

Combining Lemma~\ref{lemma:lemma15} with the above inequality and using the Poincar\'{e} inequality, we have:
\begin{equation}
\begin{aligned}
&\inf_{\substack{(l,q) \in \mathcal{L} }} \left\{ \int_{B(\rho)} |\chi - l|^2 + \rho^{d+2} |\zeta(l, q)|^2 \right\} \\
&\le \inf_{\substack{(l,q) \in \mathcal{L} }}\left\{ \int_{B(\rho)} |w_t - l|^2 + \rho^{d+2} |\zeta(l, q)|^2 \right\} + \int_{B(\rho)} |\chi - w_t|^2 \\
&\le C \left( \frac{\rho}{t} \right)^{2\beta + 2 + d} \inf_{\substack{(l,q) \in \mathcal{L} }} \left\{ \int_{B(t)} |w_t - l|^2 + t^{d+2} |\zeta(l, q)|^2 \right\} + C \int_{B(t)} |\chi - w_t|^2 \\
&\le C \left( \frac{\rho}{t} \right)^{2\beta + 2 + d} \inf_{\substack{(l,q) \in \mathcal{L} }} \left\{ \int_{B(t)} |\chi - l|^2 + t^{d+2} |\zeta(l, q)|^2 \right\} \\
&\quad + C \left( \frac{\rho}{t} \right)^{2\beta + 2 + d} \int_{B(t)} |\chi - w_t|^2 + C \int_{B(t)} |\chi - w_t|^2 \\
&\le C \left( \frac{\rho}{t} \right)^{2\beta + 2 + d} \inf_{\substack{(l,q) \in \mathcal{L} }} \left\{ \int_{B(t)} |\chi - l|^2 + t^{d+2} |\zeta(l, q)|^2 \right\} + C t^2 \int_{B(t)} |\nabla \chi - \nabla w_t|^2 \\
&\le C \left( \frac{\rho}{t} \right)^{2\beta + 2 + d} \inf_{\substack{(l,q) \in \mathcal{L} }} \left\{ \int_{B(t)} |\chi - l|^2 + t^{d+2} |\zeta(l, q)|^2 \right\} + C t^{2 + 2\alpha} \left( \int_{B(t)} |\nabla \chi|^2 + \int_{B(t)_-} |r|^2 \right).
\end{aligned}
\end{equation}

Define 
\begin{equation}
\begin{aligned}
\Phi(r) = \inf_{\substack{l \in \mathcal{L} \\ q \in \mathcal{L}}} \left\{ \int_{B(r)} |\chi - l|^2 + r^{d+2} |\zeta(l, q)|^2 \right\},
\end{aligned}
\end{equation}
and 
\begin{equation}
\begin{aligned}
b(t) = \int_{B(t)} |\nabla \chi|^2 + \int_{B(t)_-} |r|^2.
\end{aligned}
\end{equation}
Then we have 
\begin{equation}
\begin{aligned}
b(t) \le C t^r \Psi\left( \frac{1}{2} \right) = C t^{d - \alpha} \Psi\left( \frac{1}{2} \right)
\end{aligned}
\end{equation}
where we set \(r = d - \alpha\).
\end{proof}
}
{\color{Lgreen}

Moreover, it follows that:
\begin{equation}
\begin{aligned}
\Phi(\rho) \le C \left( \frac{\rho}{t} \right)^{d + 2\beta + 2} \Phi(t) + C t^{d + 2 + \alpha} \Psi\left( \frac{1}{2} \right).
\end{aligned}
\end{equation}

For \(\beta \in (\alpha, 1)\), by Lemma~\ref{lemma:lemma16}, we obtain:
\begin{equation}
\begin{aligned}
\Phi(\rho) \le C \rho^{d + \alpha + 2} \left( \Phi\left( \frac{1}{2} \right) + \Psi\left( \frac{1}{2} \right) \right).
\end{aligned}
\end{equation}
This implies $\chi \in C^{1,\frac{\alpha}{2}}(\overline{B({\frac{1}{2}})_{\pm}};\mathbb{R}^d)$ for all $ \forall \rho \in (0, \frac{1}{2})$.
By Lemma~\ref{lemma:lemma16}, we further derive:
\begin{equation}
\begin{aligned}
\Phi(\rho) \le C \rho^{d + 2\alpha + 2} \left( \Phi\left( \frac{1}{2} \right) + \Psi\left( \frac{1}{2} \right) \right),
\end{aligned}
\end{equation}
which implies \(\chi \in C^{1, \alpha}(\overline{B({\frac{1}{2}})_{\pm}}; \mathbb{R}^d)\) for all \(\rho \in (0, \frac{1}{2})\).}
\subsection{Estimation}
{\color{blue}

We define the cut-off function as follows: Let \(\eta_{\varepsilon} \in C_0^{\infty}(\Omega)\), which satisfies:
\[
\begin{cases}
0 \le \eta_{\varepsilon}(x) \le 1, & \forall x \in \Omega, \\
|\nabla \eta_{\varepsilon}(x)| \le \frac{C}{\varepsilon}, & \text{within the support of } \eta_{\varepsilon}\ (C \text{ is a constant independent of } \varepsilon), \\
\eta_{\varepsilon}(x) = 1, & \text{if } \text{dist}(x, \partial \Omega) \ge 4\varepsilon, \\
\eta_{\varepsilon}(x) = 0, & \text{if } \text{dist}(x, \partial \Omega) \le 3\varepsilon.
\end{cases}
\]

 And we also set this convolution operator:\[ S_{\varepsilon}(f)(x)=\rho_{\varepsilon} \ast f(x)=\int_{R^d}f(x-y)\rho_{\varepsilon}\diff y,\ \text{here}\ \rho \in C_0^{\infty}(B(0,\frac{1}{2})), \rho \ge 0, \text{ and }\int_{R^d} \rho \,\diff x=1.\]  
 We let \[w_{\varepsilon}=u_{\varepsilon}-u_0-\varepsilon \chi(\frac{x}{\varepsilon})\eta_{\varepsilon}S_{\varepsilon}^2(\nabla u_0).\]
We let \( B(y)=A+A\nabla\chi(y)+\nabla\chi(y)A+\nabla\chi A\nabla\chi-\hat{A}  \) be the Flux Corrector.}
{\color{Lgreen}
\begin{lemma}
    Suppose $\Omega$ is a bounded Lipschitz domain in $R^d$, $\Omega_t=\{x\in \Omega:dist(x,\partial \Omega )<t \},t>0$, $\forall\psi \in H_0^1(\Omega,R^d)$, we have
   \begin{equation}
   \begin{aligned}
       \vert \int_{\Omega} A\nabla w_{\varepsilon}:\nabla \psi \,\diff x \vert \le C\Vert \nabla \chi\Vert_{L^2(\Omega)}\{\varepsilon \Vert S_{\varepsilon}(\nabla^2u_0)\Vert_{L^2(\Omega\setminus \Omega_{3\varepsilon})}+\Vert \nabla u_0-S_{\varepsilon}(\nabla u_0)\Vert_{L^2(\Omega\setminus\Omega_{2\varepsilon})} \}\\
       + C \|\nabla\psi\|_{L^2(\Omega_{4\varepsilon})} \|\nabla u_0\|_{L^2(\Omega_{5\varepsilon})}.
   \end{aligned}
   \end{equation}
   \label{lemma:lemma17}
\end{lemma}

\begin{proof}
\begin{equation}
\begin{aligned}
    A\nabla w_{\varepsilon} &= A\nabla u_{\varepsilon} - A\nabla u_0 - A\nabla \chi\left(\frac{x}{\varepsilon}\right) S_{\varepsilon}^2(\nabla u_0) - \varepsilon A \chi(y)\nabla\left(\eta_\varepsilon S_\varepsilon^2(\nabla u_0)\right) \notag \\
    &= (A\nabla u_\varepsilon - \hat{A}\nabla u_0) + \left[(\hat{A} - A)\nabla u_0 + \left(-\hat{A}+A+\nabla\chi\left(\frac{x}{\varepsilon}\right)A+\nabla\chi A\nabla\chi\right)\eta_{\varepsilon}S_{\varepsilon}^2(\nabla u_0)\right] \notag \\
    &\quad - \left[A + A\nabla \chi\left(\frac{x}{\varepsilon}\right) + \nabla\chi\left(\frac{x}{\varepsilon}\right)A + \nabla \chi A\nabla\chi -\hat{A}\right]\eta_{\varepsilon}S_{\varepsilon}^2(\nabla u_0) - \varepsilon A \chi\left(\frac{x}{\varepsilon}\right)\nabla\left(\eta_{\varepsilon}S_{\varepsilon}^2(\nabla u_0)\right) \notag \\
    &= (A\nabla u_\varepsilon - \hat{A}\nabla u_0) + (\hat{A} - A)\left(\nabla u_0 - \eta_\varepsilon S_\varepsilon^2(\nabla u_0)\right) \notag + \nabla \chi A \eta_\varepsilon S_\varepsilon^2(\nabla u_0) + \nabla \chi A \nabla \chi \eta_\varepsilon S_\varepsilon^2(\nabla u_0) \notag \\
    &\quad - B\left(\frac{x}{\varepsilon}\right)\eta_\varepsilon S_\varepsilon^2(\nabla u_0) - \varepsilon A \chi\left(\frac{x}{\varepsilon}\right)\nabla\left(\eta_\varepsilon S_\varepsilon^2(\nabla u_0)\right).
\end{aligned}
\end{equation}
Because of the variational form:
$$\int_\Omega A\left(\frac{x}{\varepsilon}\right) \nabla u_\varepsilon : \nabla \psi + \int_\Omega p^{\varepsilon} \cdot \mathrm{div}\, \psi = 0, \quad \forall \psi \in C_0^\infty(\Omega; \mathbb{R}^m),$$

we can get:
\begin{equation}
\begin{aligned}
\int_\Omega A \nabla w_{\varepsilon} : \nabla \psi &= 
\int_{\Omega} A \nabla u_\varepsilon:\nabla \psi - \hat{A}\nabla u_0:\nabla \psi + \int_{D_{\varepsilon}} p^{\varepsilon}\cdot \mathrm{div}\, \psi \\
&\quad + \int_{\Omega} (\hat{A}-A)\left(\nabla u_0 - \eta_{\varepsilon}S_{\varepsilon}^2(\nabla u_0)\right):\nabla \psi \\
&\quad + \int_{\Omega} \left[\nabla\chi A \eta_{\varepsilon}S_{\varepsilon}^2(\nabla u_0) + \nabla\chi A\nabla\chi \eta_{\varepsilon}S_{\varepsilon}^2(\nabla u_0)\right]:\nabla \psi \\
&\quad - \int_{D_{\varepsilon}} p^{\varepsilon}\cdot \Div\, \psi - \int_\Omega B \eta_\varepsilon S_\varepsilon^2(\nabla u_0) : \nabla \psi \\
&\quad - \int_\Omega \varepsilon A \chi\left(\frac{x}{\varepsilon}\right) \nabla\left(\eta_\varepsilon S_\varepsilon^2(\nabla u_0)\right) : \nabla \psi.
\end{aligned}
\end{equation}

The first three terms equal zero, then we can see:
\begin{equation}
\begin{aligned}
\int_{\Omega} (\hat{A}-A)\left(\nabla u_0 - \eta_{\varepsilon}S_{\varepsilon}^2(\nabla u_0)\right):\nabla\psi &\le
C\int_\Omega (1-\eta_\varepsilon)\vert \nabla u_0\vert \vert \nabla\psi\vert \,\diff x \\
&\quad + C\int_{\Omega} \eta_{\varepsilon}\vert \nabla u_0 - S_{\varepsilon}^2(\nabla u_0)\vert \vert \nabla \psi \vert \,\diff x \\
&\le C\|\nabla u_0\|_{L^2(\Omega_{4\varepsilon})}\|\nabla\psi \|_{L^2(\Omega_{4\varepsilon})} \\
&\quad + C\|\nabla u_0 - S_{\varepsilon}(\nabla u_0)\|_{L^2(\Omega\setminus\Omega_{2\varepsilon})}\|\nabla\psi \|_{L^2(\Omega)}.
\end{aligned}
\end{equation}

This is because:
\begin{equation}
\begin{aligned}
\| \nabla u_0 - S_{\varepsilon}^2(\nabla u_0)\|_{L^2(\Omega\setminus\Omega_{3\varepsilon})} &\le
\| \nabla u_0 - S_{\varepsilon}(\nabla u_0)\|_{L^2(\Omega\setminus\Omega_{3\varepsilon})} \\
&\quad + \| S_{\varepsilon}(\nabla u_0) - S_{\varepsilon}^2(\nabla u_0)\|_{L^2(\Omega\setminus\Omega_{3\varepsilon})} \\
&\le C\| \nabla u_0 - S_{\varepsilon}(\nabla u_0)\|_{L^2(\Omega\setminus\Omega_{2\varepsilon})}.
\end{aligned}
\end{equation}

Moreover, by the regularity of \(\chi \in W^{1,\infty}\), we have
\begin{equation}
\begin{aligned}
&\int_{\Omega} \left[\nabla\chi A \eta_{\varepsilon}S_{\varepsilon}^2(\nabla u_0) + \nabla\chi A\nabla\chi \eta_{\varepsilon}S_{\varepsilon}^2(\nabla u_0)\right]:\nabla \psi - \int_{D_{\varepsilon}} p^{\varepsilon}\cdot \Div\, \psi \\
&\le \int_{\Omega}\left[(\nabla \chi A + \nabla\chi A\nabla\chi)\left(\eta_{\varepsilon}S_{\varepsilon}^2(\nabla u_0) - \nabla u_0\right) \right]:\nabla \psi \\
&\quad + \int_{\Omega}(\nabla \chi A + \nabla \chi A\nabla \chi)\nabla u_0:\nabla \psi - \int_{\omega}p_0 \Div\, \psi + \mathcal{O}(\varepsilon) \\
&= \int_{\Omega}\left[(\nabla \chi A + \nabla\chi A\nabla\chi)\left(\eta_{\varepsilon}S_{\varepsilon}^2(\nabla u_0) - \nabla u_0\right) \right]:\nabla \psi \\
&\quad + \int_{\omega} \delta_{j \beta}r^{i\alpha} \frac{\partial u_0^{\alpha}}{\partial x_i}\frac{\partial \psi^{\beta}}{\partial x_j} - \int_{\omega}\frac{\partial u^{i}}{\partial x_j}r^{ij}\frac{\partial \psi^{k}}{\partial x_k} + \mathcal{O}(\varepsilon) \\
&= \int_{\Omega}\left[(\nabla \chi A + \nabla\chi A\nabla\chi)\left(\eta_{\varepsilon}S_{\varepsilon}^2(\nabla u_0) - \nabla u_0\right) \right]:\nabla \psi + \mathcal{O}(\varepsilon) \\
&\le C\|\nabla u_0 - S_{\varepsilon}(\nabla u_0)\|_{L^2(\Omega\setminus\Omega_{2\varepsilon})}\|\nabla \psi \|_{L^2(\Omega)} + \|\nabla u_0\|_{\Omega_{4\varepsilon}}\|\nabla\psi\|_{L^2(\Omega_{4\varepsilon}} +\mathcal{O}(\varepsilon).
\end{aligned}
\end{equation}

And we also have 
\begin{equation}
\begin{aligned}
\left| \int_\Omega B \eta_\varepsilon S_\varepsilon^2(\nabla u_0) : \nabla \psi \right|
&= \left| \int_{\Omega} b_{ij}^{\alpha\beta}\left(\frac{x}{\varepsilon}\right)S_{\varepsilon}^2\left(\frac{\partial u_0^{\beta}}{\partial x_j}\right)\left(\frac{\nabla\psi^{\alpha}}{\partial x_i}\right)\eta_{\varepsilon} \right| \\
&= \left| \int_{\Omega}\varepsilon \frac{\partial}{\partial x_k}\left(\phi_{kij}^{\alpha\beta}\right)\left(\frac{x}{\varepsilon}\right)\frac{\partial \psi^{\alpha}}{\partial x_i} S_{\varepsilon}^2\left(\frac{\partial u_0^{\beta}}{\partial x_j}\right)\eta_{\varepsilon} \right| \\
&\le C\varepsilon \left[ \int_{\Omega}\eta_{\varepsilon}\left| \phi\left(\frac{x}{\varepsilon}\right)\right| \left| \nabla \psi\right| \left| S_{\varepsilon}^2(\nabla^2u_0)\right| \diff x \right. \\
&\quad \left. + \int_{\Omega}\left| \nabla \eta_{\varepsilon}\right| \left| \phi\left(\frac{x}{\varepsilon}\right)\right| \left| \nabla \psi\right| \left| S_{\varepsilon}^2(\nabla u_0)\right| \diff x \right] \\
&\le C\left( \varepsilon \|\nabla \psi\|_{L^2(\Omega)}\|S_{\varepsilon}(\nabla^2 u_0)\|_{L^2(\Omega \setminus \Omega_{2\varepsilon})} \right. \\
&\quad \left. + \|\nabla\psi\|_{L^2(\Omega_{4\varepsilon})}\|\nabla u_0\|_{L^2(\Omega_{5\varepsilon})} \right).
\end{aligned}
\end{equation}

By the calculations above, we can finally get the following result
\begin{equation}
\begin{aligned}
\int_\Omega A\left(\frac{x}{\varepsilon}\right) \nabla u_\varepsilon : \nabla \psi &\le C\|\nabla \psi_{\varepsilon}\|_{L^2(\Omega)}\left\{ \varepsilon \|S_{\varepsilon}(\nabla u_0)\|_{L^2(\Omega\setminus\Omega_{3\varepsilon})} \right. \\
&\quad \left. + \|\nabla u_0 - S_{\varepsilon}(\nabla u_0)\|_{L^2(\Omega\setminus \Omega_{2\varepsilon})} \right\} \\
&\quad + C\|\nabla w_{\varepsilon}\|_{L^2(\Omega_{4\varepsilon})}\|\nabla u_0\|_{L^2(\Omega_{5\varepsilon})}.
\end{aligned}
\end{equation}
\end{proof}
}
{\color{orange}
\begin{theorem}
Let \(\Omega \subset \mathbb{R}^d\) be a bounded Lipschitz domain. Then for all \(0 < \varepsilon < 1\),
\begin{equation}
\begin{aligned}
\|\nabla w_{\varepsilon}\|_{L^2(\Omega)} \le C\left\{ \varepsilon \|\nabla^2 u_0\|_{L^2(\Omega \setminus \Omega_{\varepsilon})} + \|\nabla u_0\|_{L^2(\Omega_{5\varepsilon})} \right\}.
\end{aligned}
\end{equation}
Thus,
\begin{equation}
\begin{aligned}
\|w_\varepsilon\|_{H_0^1(\Omega)} \le C \varepsilon^{\frac{1}{2}} \|u_0\|_{H^2(\Omega)},
\end{aligned}
\end{equation}
where \(C\) is a constant depending on \(\mu\) and \(\Omega\).
\end{theorem}

\begin{proof}
Let \(\tilde{\eta}_{\varepsilon} \in C_0^{\infty}(\Omega)\) satisfy \(0 \le \tilde{\eta}_{\varepsilon} \le 1\), \(\tilde{\eta}_{\varepsilon} = 0\) in \(\Omega_{\varepsilon}\), \(\tilde{\eta}_{\varepsilon} = 1\) in \(\Omega \setminus \Omega_{\frac{3\varepsilon}{2}}\), and \(\vert \nabla\tilde{\eta}_{\varepsilon} \vert \le \frac{C}{\varepsilon}\).

We can obtain
\begin{equation}
\begin{aligned}
\|\nabla u_0 - S_{\varepsilon}(\nabla u_0)\|_{L^2(\Omega \setminus \Omega_{2\varepsilon})} &\le \|\tilde{\eta}_{\varepsilon}(\nabla u_0) - S_{\varepsilon}(\tilde{\eta}_{\varepsilon}\nabla u_0) \|_{L^2(\mathbb{R}^d)} \\
&\le C \varepsilon \|\nabla (\tilde{\eta}_{\varepsilon}\nabla u_0)\|_{L^2(\mathbb{R}^d)} \\
&\le C\left\{ \varepsilon \|\nabla^2 u_0\|_{L^2(\Omega \setminus \Omega_{\varepsilon})} + \|\nabla u_0\|_{L^2(\Omega_{2\varepsilon})} \right\}.
\end{aligned}
\end{equation}

By the inequality \(\|f\|_{L^2(\Omega_t)} \le Ct^{\frac{1}{2}}\|f\|_{H^1(\Omega)}\), we have
\begin{equation}
\begin{aligned}
\|\nabla u_0\|_{L^2(\Omega_{5\varepsilon})} \le C\varepsilon^{\frac{1}{2}}\|u_0\|_{H^2(\Omega)}.
\end{aligned}
\end{equation}

By Lemma~\ref{lemma:lemma17}, set \(\psi = w_\varepsilon \in H_0^1(\Omega; \mathbb{R}^d)\). Using the ellipticity of \(A\), we get
\begin{equation}
\begin{aligned}
\|\nabla w_\varepsilon\|_{L^2(\Omega)} &\le C\left\{ \varepsilon \|S_{\varepsilon}(\nabla^2 u_0)\|_{L^2(\Omega \setminus \Omega_{3\varepsilon})} + \|\nabla u_0 - S_{\varepsilon}(\nabla u_0)\|_{L^2(\Omega \setminus \Omega_{2\varepsilon})} \right\} \\
&\quad + C\|\nabla u_0\|_{L^2(\Omega_{5\varepsilon})} \le C\varepsilon^{\frac{1}{2}}\|u_0\|_{H^2(\Omega)}.
\end{aligned}
\end{equation}
\end{proof}
}
{\color{Lgreen}
\begin{theorem}
Let \(\Omega \subset \mathbb{R}^d\) be a bounded Lipschitz domain. Then for all \(0 < \varepsilon < 1\), if \(u_0 \in H^2(\Omega; \mathbb{R}^d)\), we have
\begin{equation}
\begin{aligned}
\| u_\varepsilon - u_0 - \varepsilon \chi\left(\frac{x}{\varepsilon}\right) \nabla u_0 \|_{H^1(\Omega)} \leq C(\mu, \Omega, \|\chi\|_\infty) \varepsilon^{\frac{1}{2}} \|u_0\|_{W^{2,d}(\Omega)}.
\end{aligned}
\end{equation}
\end{theorem}

\begin{proof}

The key step is to prove:
\begin{equation}
\begin{aligned}
\left\| \varepsilon\chi\left(\frac{x}{\varepsilon}\right) \nabla u_0 - \varepsilon \chi\left(\frac{x}{\varepsilon}\right) \eta_\varepsilon S_\varepsilon^2(\nabla u_0) \right\|_{H^1(\Omega)} \leq C \varepsilon^{\frac{1}{2}} \|u_0\|_{H^2(\Omega)}.
\end{aligned}
\end{equation}

We can find that
\begin{equation}
\begin{aligned}
&\left\| \varepsilon\chi\left(\frac{x}{\varepsilon}\right) \nabla u_0 - \varepsilon \chi\left(\frac{x}{\varepsilon}\right) \eta_\varepsilon S_\varepsilon^2(\nabla u_0) \right\|_{H^1(\Omega)} \leq C\varepsilon \left\| \chi\left(\frac{x}{\varepsilon}\right)\left( \nabla u_0 -\eta_\varepsilon S_\varepsilon^2(\nabla u_0) \right) \right\|_{L^2(\Omega)} \\
& + C\left\| \nabla\chi\left(\frac{x}{\varepsilon}\right) \left( \nabla u_0 - \eta_\varepsilon S_\varepsilon^2(\nabla u_0) \right) \right\|_{L^2(\Omega)}  + C\varepsilon\left\| \chi\left(\frac{x}{\varepsilon}\right) \nabla\left( \nabla u_0 - \eta_\varepsilon S_\varepsilon^2(\nabla u_0) \right) \right\|_{L^2(\Omega)} \\
&\leq C\varepsilon\left\| \nabla\left( \nabla u_0 - \eta_\varepsilon S_{\varepsilon}^2(\nabla u_0) \right) \right\|_{L^2(\Omega)} + C\left\| \nabla u_0 - \eta_\varepsilon S_{\varepsilon}^2(\nabla u_0) \right\|_{L^2(\Omega)}.
\end{aligned}
\end{equation}

On the one hand, we have
\begin{equation}
\begin{aligned}
\varepsilon\left\| \nabla\left( \nabla u_0 - \eta_\varepsilon S_{\varepsilon}^2(\nabla u_0) \right) \right\|_{L^2(\Omega)} &\leq \varepsilon \left\| \nabla^2 u_0 \right\|_{L^2(\Omega)} + \varepsilon\left\| \nabla\left( \eta_\varepsilon S_{\varepsilon}^2(\nabla u_0) \right) \right\|_{L^2(\Omega)} \\
&\leq C \varepsilon \|u_0\|_{H^2(\Omega)} + C\left\| \nabla u_0 \right\|_{L^2(\Omega_{5\varepsilon})} \\
&\leq C\varepsilon^{\frac{1}{2}} \|u_0\|_{H^2(\Omega)}.
\end{aligned}
\end{equation}

On the other hand,
\begin{equation}
\begin{aligned}
\left\| \nabla u_0 - \eta_\varepsilon S_{\varepsilon}^2(\nabla u_0) \right\|_{L^2(\Omega)} &\leq C\left\| \nabla u_0 \right\|_{L^2(\Omega_{5\varepsilon})} + C\left\| \nabla u_0 - S_{\varepsilon}^2(\nabla u_0) \right\|_{L^2(\Omega \setminus \Omega_{4\varepsilon})} \\
&\leq C\varepsilon^{\frac{1}{2}} \|u_0\|_{H^2(\Omega)}.
\end{aligned}
\end{equation}
\end{proof}
}
\subsection{Summary}

These conclusions is consistent to the expected conclusion (Eq.~\ref{eq:conclusion}), since we have:
\begin{equation}
\begin{aligned}
\|u_{\varepsilon}-u_{\lim}\|_{H^1(\Omega)}&=
\| u_\varepsilon - u_0 - \varepsilon \chi\left(\frac{x}{\varepsilon}\right) \nabla u_0 \|_{H^1(\Omega)} \\
&\leq C\|w_{\varepsilon}\|_{H^1(\Omega)}+C\| \varepsilon\chi\left(\frac{x}{\varepsilon}\right) \nabla u_0 - \varepsilon \chi\left(\frac{x}{\varepsilon}\right) \eta_\varepsilon S_\varepsilon^2(\nabla u_0)\|_{H^1(\Omega)} \\
&\leq C \varepsilon^{\frac{1}{2}} \|u_0\|_{H^2(\Omega)}.
\end{aligned}
\end{equation}

\end{document}